\documentclass{article} 
\usepackage[final]{colm2025_conference}

\usepackage{microtype}
\usepackage{hyperref}
\usepackage{url}
\usepackage{booktabs}
\usepackage{multirow}
\usepackage{xcolor}

\usepackage{lineno}

\usepackage{array}
\usepackage{amsmath,amssymb}
\usepackage{mathtools}
\usepackage{pifont}
\usepackage{graphicx}

\usepackage{amsmath,amsfonts,bm}

\def\Figref#1{Figure~\ref{#1}}

\def\eqref#1{equation~\ref{#1}}

\def\1{\bm{1}}

\def\mX{{\bm{X}}}

\DeclareMathAlphabet{\mathsfit}{\encodingdefault}{\sfdefault}{m}{sl}
\SetMathAlphabet{\mathsfit}{bold}{\encodingdefault}{\sfdefault}{bx}{n}

\def\sR{{\mathbb{R}}}

\def\Tref#1{Table~\ref{#1}}

\def\Sref#1{\S\ref{#1}}
\def\sref#1{\S\ref{#1}}

\definecolor{darkblue}{rgb}{0, 0, 0.5}
\hypersetup{colorlinks=true, citecolor=darkblue, linkcolor=darkblue, urlcolor=darkblue}
\definecolor{indianred}{HTML}{CD5C5C}
\definecolor{royalblue}{HTML}{4169E1}

\newcommand{\s}{\textsc{s}}
\newcommand{\wc}{\textsc{wc}}
\newcommand{\p}{\textsc{p}}
\DeclarePairedDelimiter{\norm}{\lVert}{\rVert}

\title{Language models align with brain regions that represent concepts across modalities}

\author{Maria Ryskina$^{1\ddagger}$, Greta Tuckute$^{2}$, Alexander Fung$^{2}$, Ashley Malkin$^{2}$, Evelina Fedorenko$^{2}$ \\
$^{1}$Vector Institute for AI \quad $^{2}$MIT\\
$^{\ddagger}$Work done at MIT\\
\texttt{maria.ryskina@vectorinstitute.ai} \\
}

\begin{document}

\ifcolmsubmission
\linenumbers
\fi

\maketitle

\begin{abstract}

Cognitive science and neuroscience have long faced the challenge of disentangling representations of language from representations of conceptual meaning. As the same problem arises in today's language models (LMs), we investigate the relationship between LM--brain alignment and two neural metrics: (1) the level of brain activation during processing of sentences, targeting linguistic processing, and (2) a novel measure of meaning consistency across input modalities, which quantifies how consistently a brain region responds to the same concept across paradigms (sentence, word cloud, image) using an fMRI dataset \citep{pereira2018toward}.
Our experiments show that both language-only and language-vision models predict the signal better in more meaning-consistent areas of the brain, even when these areas are not strongly sensitive to language processing, suggesting that LMs might internally represent cross-modal conceptual meaning.\footnote{Our code can be found at \url{https://github.com/ryskina/concepts-brain-llms}}
\end{abstract}

\section{Introduction}

Much recent work at the intersection of AI and neuroscience has focused on discovering the similarities and differences between the human brain and increasingly complex and powerful artificial neural models \citep{oota2024deep,sucholutsky2024gettingalignedrepresentationalalignment,tuckute2024language}. 
Often, studies compare how these two systems encode information internally---for example, how sentence representations in a language model (LM) align with the responses to the same sentences in a certain region of the brain. Previous work has found correlations between how sentences or narratives are represented in LMs and in the brain's language network \citep{toneva2019interpreting,schrimpf2021neural,goldstein2022shared,tuckute2024driving}, as well as between image representations in convolutional neural networks and the visual cortex \citep{yamins2014performance, horikawa2017generic,conwell2024large}. 
However, as models become more seamless in integrating different modalities, a new question arises: do these models represent deeper, modality-independent conceptual information in a brain-like way?

\begin{figure}[tb]
\begin{center}
\includegraphics[width=\linewidth]{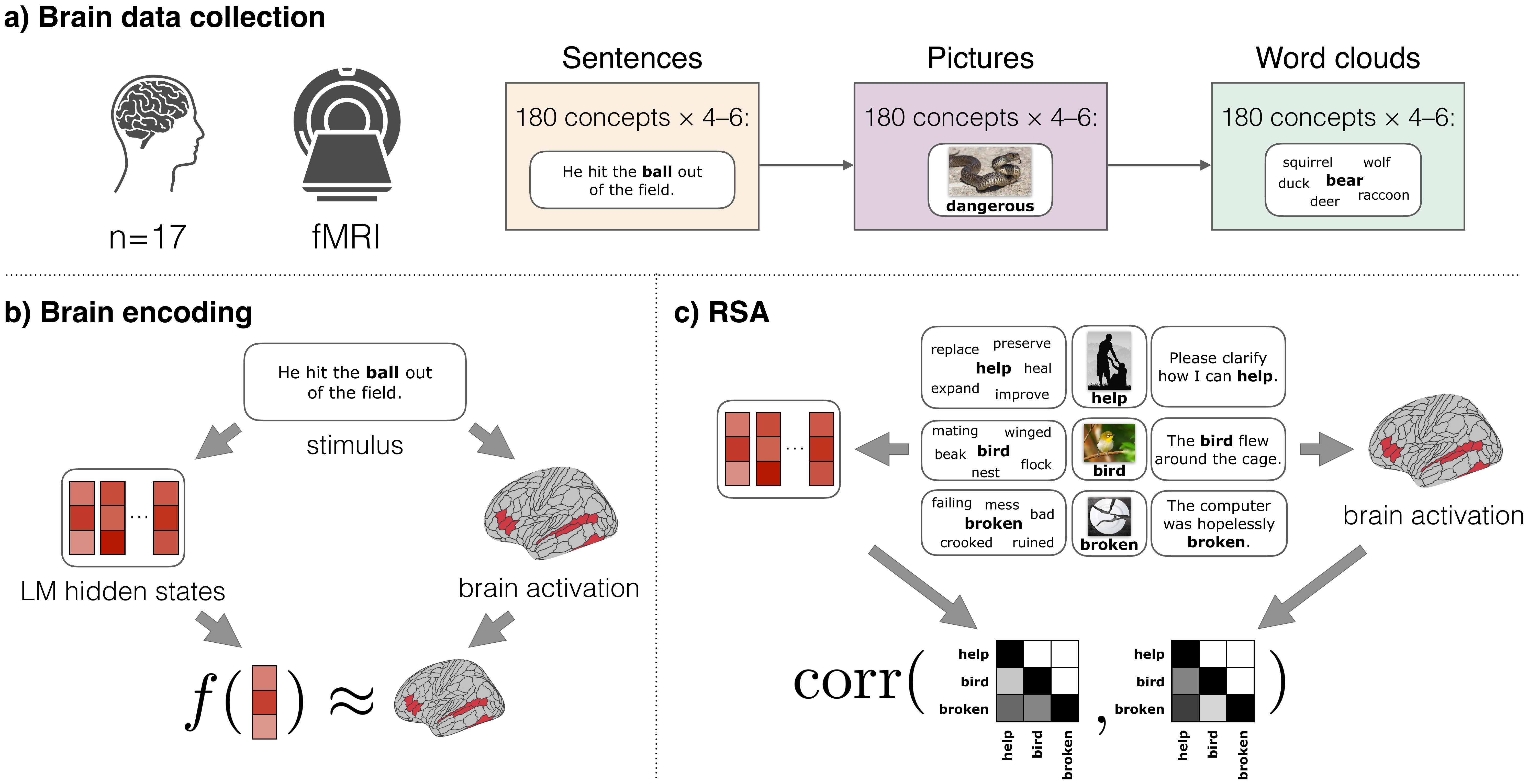}
\end{center}
\caption{\textbf{Brain data collection process for the fMRI dataset 
\citep[Experiment 1]{pereira2018toward} and the schematics of our two LM--brain alignment evaluations.} 
\textbf{(a)} 17 participants underwent three fMRI scan sessions, one per paradigm (sentences, pictures, or word clouds) to record brain activity when thinking of different concepts. Each paradigm presents the 180 concepts in a different format: sentences containing the concept word, pictures presented alongside the concept word, and word clouds with the concept word surrounded by related terms. 
Example stimuli are shown; each concept is represented by 4--6 unique stimuli per session.
\textbf{(b)} Brain encoding (\sref{sec:brain-enc}): we use the LM representation of the stimulus to predict brain activations in a participant viewing the same stimulus.
\textbf{(c)} Representational similarity alignment (RSA) (\sref{sec:rsa}): we combine all stimuli per concept to obtain a single concept representation from the brain and from the LM. We use them to evaluate pairwise concept dissimilarities in the LM and the brain and correlate them between the two.
}
\label{fig:1}
\end{figure}

Recent evidence suggests that such conceptual representations exist in multimodal models \citep{wu2024semantic} and that models learn similar representations from different modalities \citep{merullo2023linearly, maniparambil2024vision, huh2024position}.
However, comparing these representations with the brain is challenging given that the ways in which the brain represents and processes conceptual knowledge remain debated \citep{kiefer2012conceptual} and there are no clearly delineated ``concept-representing regions''. In this paper, we propose a new way of localizing concept-representing areas in the brain by using fMRI data collected in a multimodal experiment targeting conceptual processing \citep{pereira2018toward}. In this study, participants read text or looked at images representing a particular concept, and their brain responses to these stimuli were recorded. Each concept was presented in three \emph{paradigms} spanning two modalities (language and vision): (1) as a highlighted word in a sentence, (2) as a highlighted word in the middle of a relevant word cloud, or (3) as a picture labeled with the concept word (Fig.~\ref{fig:1}a). We introduce a \emph{semantic consistency} metric for how consistently a particular brain unit (voxel) responds to the same concept in all three paradigms (\sref{sec:metric}), and identify three brain areas that show high semantic consistency (\sref{sec:rois}).

Next, we ask if the representations from 15 uni- and multimodal transformer LMs of different sizes are aligned with brain responses in these areas during linguistic and conceptual processing. Our main question is whether LM-based encoding performance correlates with the level of semantic consistency for a given brain region; in addition, we look at the relationship between the encoding quality and the region's selectivity for language.
Methodologically, we use two approaches: (1) using LM features to predict activations in these regions (Fig.~\ref{fig:1}b), and (2) performing a representational similarity analysis (RSA) to probe the geometric structure of concept representations in the brain and in LMs (Fig.~\ref{fig:1}c).
To preface our key results, all models show significant brain alignment in both the prediction and the RSA analyses. Moreover, high semantic consistency correlates with high predictivity---even in  regions with weak language responses, which suggest that these areas indeed represent non-linguistic conceptual information.
Overall, our contributions are the following:
\begin{itemize}
\setlength\itemsep{0.25em}
\item Using an fMRI dataset of brain responses to multimodal stimuli, we define a novel metric for measuring semantic consistency in the brain (\sref{sec:metric}) and use it to find
brain regions that represent concepts most consistently, irrespective of paradigm
(\sref{sec:rois});
\item We evaluate 15 uni- and multimodal transformer models on their ability to predict brain activations in three newly identified semantically consistent regions (\sref{sec:brain-enc}) and compare the models' representational geometry to the brain's (\sref{sec:rsa});
\item We show that models' predictive performance correlates with our metric of semantic consistency in the brain, both across the whole brain and in the high-consistency regions specifically, including brain regions with a low response to language (\sref{sec:exp1});
\item We find significant representational similarity  
between the models and the semantically consistent brain regions and show that  
it further increases when both text and image stimuli are used (\Sref{sec:exp2}).
\end{itemize}

\section{Related work}

\paragraph{LM--brain alignment} 

A growing body of work compares representations in deep neural network language models to brain imaging data
\citep{karamolegkou-etal-2023-mapping,oota2024deep,sucholutsky2024gettingalignedrepresentationalalignment,tuckute2024language}. Many studies adopt a brain encoding approach, predicting brain activations from the model's hidden states \citep{toneva2019interpreting, schrimpf2021neural,merlin-toneva-2024-language}
or attention head outputs \citep{kumar2024shared}. Encoding studies find that best-performing LMs \citep{schrimpf2021neural,caucheteux2022brains} and LMs fine-tuned for certain NLP tasks \citep{oota-etal-2022-neural,aw2023training} tend to be more brain-aligned, and that predictivity increases with scale \citep{antonello2023scaling} and with the addition of instruction tuning \citep{aw2024instruction}. A complementary line of work uses representational similarity alignment \citep[RSA;][]{kriegeskorte2008representational} or direct projection to compare the geometry of the model's and the brain's representational spaces \citep{kaniuth2022feature,yu2024predicting,pmlr-v228-li24a,du2025human}. Such studies can benefit both NLP and neuroscience: there is evidence that increasing brain alignment can improve model performance \citep{toneva2019interpreting} and that models can help scientists elicit targeted levels of neural activity \citep{bashivan2019neural,tuckute2024driving}. 

Recent work has explored if vision--language LMs (VLMs) are more brain-aligned than language-only ones \citep{oota-etal-2022-visio,du2025human,bavaresco-fernandez-2025-experiential}, with two studies in particular using the multimodal, concept-focused Experiment 1 data from \cite{pereira2018toward} as a testbed (used also in this work). \cite{oota-etal-2022-multi} perform brain decoding, predicting LM representations of concept words from the brain responses to stimuli in different modalities. Especially relevant to ours is the work of \cite{bavaresco2024modelling}: in an RSA analysis, they find that VLMs capture multimodal knowledge, leading to higher alignment in both language and visual networks. Unlike these studies, we do not use known brain networks as the alignment target---we identify a novel set of concept-representing brain regions by leveraging the cross-modal nature of the dataset's stimuli.

\paragraph{Concepts in the brain} How the human brain represents conceptual meaning is an open question \citep{kiefer2012conceptual,frisby2023decoding}, but several streams of scientific evidence suggest that language and semantic/conceptual processing are dissociated in the mind and brain \citep[for details and references, see][\emph{Dissent \#1 for event semantics}]{reilly2025we}. Therefore, we propose extracting meaning representations not from the language-selective brain regions commonly used in prior brain--LM work, but from the regions that represent meaning independently of whether it is conveyed through text or image. While there is no established method for localizing such regions, brain imaging studies have used visual and linguistic stimuli in parallel to search for amodal semantic processing \citep[Ch.~5]{wurm2019distinct,popham2021visual,ivanova2022role}. Similarly, we use the multimodal, concept-matched stimuli of \citet[Experiment 1]{pereira2018toward} to identify regions of interest: we propose a novel metric of how consistently a brain area responds to particular concepts---regardless of whether the concept is shown pictorially, in the context of related single words, or in a sentence context---and select areas where it is reliably high (\sref{sec:metric-and-rois}). 

\paragraph{Concepts in LMs} 

While the language models' ability to represent concepts without grounding is subject to debate \citep{bender-koller-2020-climbing,piantadosi2022meaning}, recent work has found that LMs can learn about concepts like color from text input only \citep{abdou-etal-2021-language}. Further studies show evidence for the existence of ``universal representations'', a shared brain-aligned latent space that deep neural models converge on \citep{hosseini2024universality, chen2024universal}. Convergence emerges even between models trained on different modalities \citep{maniparambil2024vision,li2024vision}, and \cite{huh2024position} argue that models are aligning towards a shared representation of reality. \cite{wu2024semantic} connect these findings to a theory of human cognition \citep{patterson2016hub}, showing that LMs  develop ``semantic hubs'' which encode shared meaning across languages and modalities. 

\section{Data}
\label{sec:data}

We use the brain data from Experiment 1 of \cite{pereira2018toward}. They collected fMRI brain recordings of 17 participants who perceived the experimental stimuli (text or images). Each stimulus corresponded to a target \emph{concept}---one of the 180 single-word labels obtained by performing clustering on a static word embedding space \citep{pennington-etal-2014-glove}. The concept words vary in part of speech \texttt{(Seafood, Disturb, Willingly, Great)} and range from concrete and material \texttt{(Table)} to abstract \texttt{(Emotion)}; 
the list of concepts is included in the Appendix (\Tref{tab:concepts}). 
Each stimulus represents a concept in one of the three experimental paradigms: as a sentence containing the concept word (sentence paradigm, or \textsc{s}), a word cloud with the concept word surrounded by relevant terms (word cloud paradigm, or \textsc{wc}), or an image presented alongside the concept word (picture paradigm, or \textsc{p}). 
The full dataset contains six sentences, six images, and six spatial arrangements of the word cloud for each concept (see \Figref{fig:stimuli} in the Appendix).\footnote{Notably, the words in each concept's word cloud remain the same in each of the six \textsc{wc} stimuli.} 
The concept word was always highlighted in bold, and the participants were asked to read the text and think about the target word's meaning in relation to the accompanying image or context.

Each participant underwent three separate 2-hour fMRI scanning sessions, one per paradigm, as shown in \Figref{fig:1}a. In each session, they viewed 4--6 groups of 180 stimuli (one per concept), in random order.
The participant never saw the same exact stimulus more than once: in every new group, the concept was always represented by a new sentence, picture, or spatial configuration of the word cloud, depending on the paradigm. Each stimulus was displayed for 3 seconds, followed by a 2-second break.

An fMRI brain recording captures the changes in blood oxygen levels (Blood Oxygenation Level Dependent (BOLD) signal), an indirect measure of neural activity. Spatially, the brain is discretized into 2mm-sized cubical units (voxels). 
To estimate the activation strength\footnote{We use 
the words `activation' and `response' interchangeably to denote the BOLD percent signal change in response to a stimulus.} 
$\beta$ in each voxel corresponding to each stimulus, we implement a processing pipeline using the GLMsingle toolkit \citep{glmsingle}, with additional upsampling of the BOLD signal time series to align stimuli presentations with the temporal resolution of the scan (2s). Further details about the collection and processing of the fMRI data are provided in Appendix \ref{sec:appendix-data}.

\section{Defining and mapping semantic consistency}
\label{sec:metric-and-rois}

We use the estimated activation values per stimulus to identify which voxels in the brain consistently respond to the same concepts, whether presented as a sentence, a picture, or a word cloud. 
We propose a measure of this conceptual consistency (\sref{sec:metric}) and use it to identify brain regions where significantly consistent voxels are likely to be found (\sref{sec:rois}).

\subsection{Semantic consistency metric}
\label{sec:metric}

We consider a voxel \emph{semantically consistent} if it consistently responds strongly (or weakly) to stimuli representing the same concept, regardless of the paradigm (e.g., if it responds strongly to sentences, pictures, and word clouds for the concept \texttt{Bird} but weakly to those for \texttt{Art}). Suppose that the stimuli associated with the concept $c_i$ 
($1 \leq i \leq 180$) under the paradigm $\Omega \in \{\text{\s, \p, \wc} \}$ elicit an average response 
$\beta_{\Omega}^i \in \mathbb{R}$ in a given voxel. We then obtain vectors $\left\{ \beta^{i}_{\Omega} \right\}_{i=1}^{180} = \Vec{\beta}_{\Omega} \in \mathbb{R}^{180}$ and define the voxel's semantic consistency as follows:

\begin{equation}
\label{eq:semcons}
    C = \frac{1}{3} \left[ r\left(\Vec{\beta}_{\text{\s}} , \Vec{\beta}_{\text{\p}}\right) + r\left(\Vec{\beta}_{\text{\s}} , \Vec{\beta}_{\text{\wc}}\right)  + r\left(\Vec{\beta}_{\text{\wc}} , \Vec{\beta}_{\text{\p}}\right) \right]
\end{equation}

where $r$ denotes the Pearson correlation coefficient (Fig.~\ref{fig:2}a).
To apply this measure to a set of voxels, we average the response values $\Vec{\beta}_{\Omega}$ over voxels before computing the correlations.

\subsection{Brain regions of interest}
\label{sec:rois}

\begin{figure}[t]
\begin{center}
\includegraphics[width=0.9\linewidth]{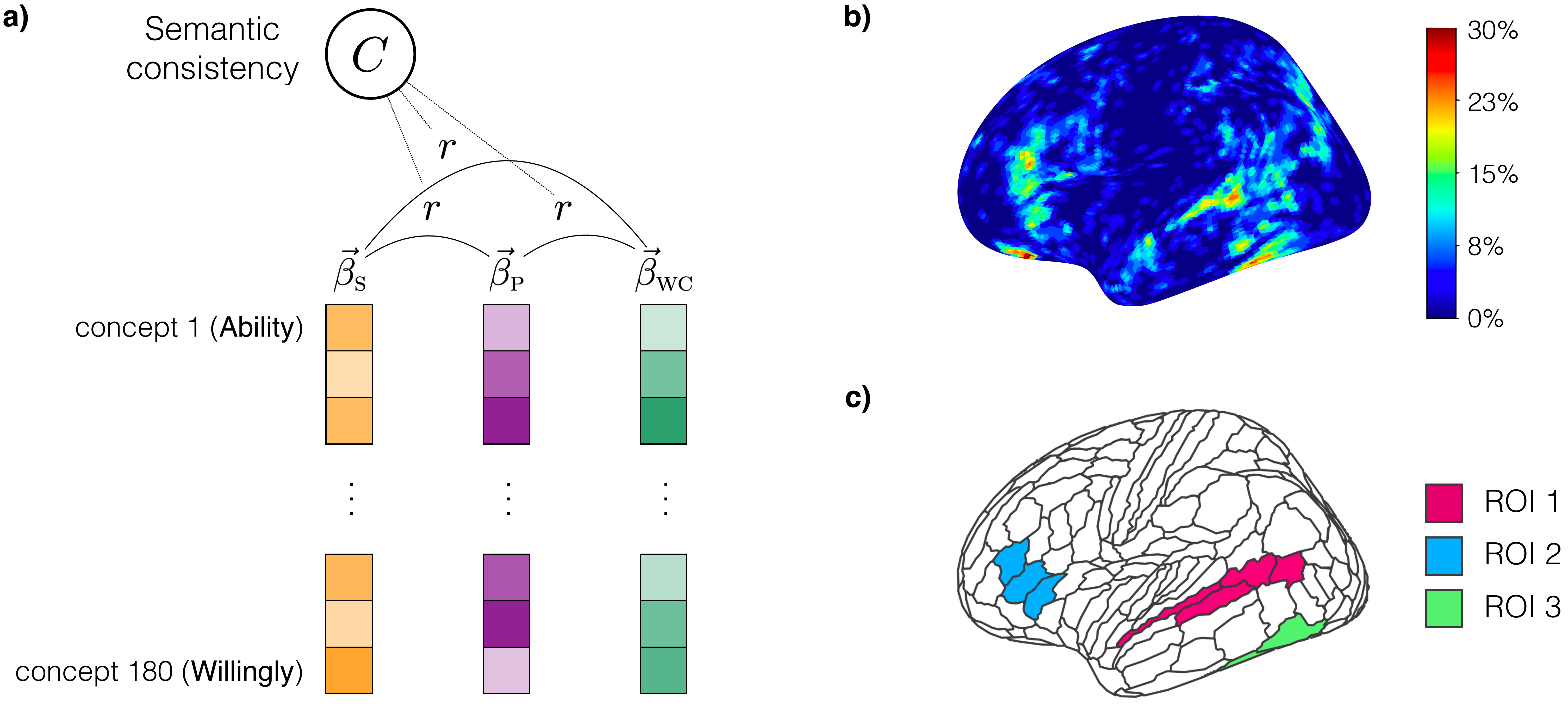}
\end{center}
\caption{\textbf{Semantic consistency and its spatial distribution.} \textbf{(a)} The schematic of the computation of the semantic consistency measure $C$. Given a brain activation vector $\Vec{\beta}$ for each of the experimental paradigms (sentences, pictures, and word clouds) over the 180 concepts, we compute Pearson correlation coefficients between each pair of activation vectors and average them. \textbf{(b)} A probabilistic semantic consistency map of the left hemisphere. Each point shows the \% of participants whose brain displays significant semantic consistency in that voxel, demonstrating where, on average, the semantically consistent brain areas are located. \textbf{(c)} Regions of interest (ROIs) that emerge after overlaying the probabilistic map in (b) with an anatomical segmentation \citep{glasser2016multi}.}
\label{fig:2}
\end{figure}

Brain representations for model--brain alignment are typically extracted from regions engaged by the input modality, e.g., the visual cortex for visual stimuli or the language network for linguistic ones.
Since we aim to explore the effect of representation consistency \emph{across paradigms}, we define our own brain regions of interest (ROIs) in a modality-agnostic way. 

First, in each participant's brain we find all voxels whose semantic consistency $C$ is reliably above chance. To account for noise in fMRI recordings, we select those via two independent permutation tests (shuffling $\Vec{\beta}_{\Omega}$ and recomputing $C$) on two separate halves of the data, and select voxels with $p < 0.05$ in both permutation tests. A probabilistic map of such voxels across all participants is shown in \Figref{fig:2}b: the voxels that show significant $C$ in a larger percentage of participants tend to cluster in certain areas of the left hemisphere. 
For further details on this step, including the whole-brain probabilistic map, see Appendix \ref{sec:appendix-prob-map}.

We define the boundaries of these areas by overlaying this probabilistic map with a popular anatomical segmentation of the brain cortex \citep[the HCP-MMP1.0 atlas;][]{glasser2016multi}, 
which divides each hemisphere into 180 functionally and anatomically distinct areas.
After we threshold contiguous clusters of areas by size and by likelihood of high-consistency voxels (full procedure described in Appendix \ref{sec:appendix-rois}), the three regions of interest (ROIs) are left, marked as ROI 1, 2, and 3 in \Figref{fig:2}c. ROI 2, located in the inferior frontal lobe, and especially ROI 1, which covers parts of the temporal lobe, include areas that are considered to be language-relevant in prior work on brain--LM alignment \citep{oota2023joint, oota-etal-2024-speech}.
ROI 3 contains ventral areas involved in visual processing \citep{10.1093/oso/9780198887911.003.0002}, which have been used for benchmarking representational alignment in computer vision models \citep{kaniuth2022feature}. The full anatomical breakdown of each ROI can be found in \Tref{tab:glasser-parcels} (\sref{sec:appendix-rois}).

\section{Brain--LM alignment}

We now measure how well brain responses to stimuli in the identified ROIs (Fig.~\ref{fig:2}c) align with the LM representations of the same stimuli. This section lists the models used in this study (\Sref{sec:models}), outlines how LM representations are extracted (\Sref{sec:embeddings}), and describes our two methods: brain encoding (predicting brain signal from LM representations; \Sref{sec:brain-enc}, Fig.~\ref{fig:1}b) and RSA (comparing the structure of the representational spaces; \Sref{sec:rsa}, Fig.~\ref{fig:1}c).

\subsection{Models}
\label{sec:models}
\subsubsection{Language-only models}

We experiment with a range of open-weights transformer \citep{Vaswani+2017} LMs from three different series: GPT-2 \citep{gpt, radford2019language}, Qwen-2.5 \citep{qwen,qwen2,qwen2.5}, and Llama-based Vicuna-1.5 \citep{vicuna2023, zheng2024judging}. 

GPT-2 is a series of autoregressive transformer models trained on English text. GPT-2 models are commonly used in brain--model alignment studies and have demonstrated high brain encoding performance \citep{schrimpf2021neural, tuckute2024driving}. We evaluate the small, medium, large, and XL models in this architecture.

Qwen2.5 is a family of large multilingual models pre-trained on a large dataset with a focus on knowledge, coding, and mathematics \citep{qwen2.5}. Both the base pre-trained models and their instruction-tuned version are released; we evaluate the 1.5B-, 3B-, and 7B-parameter models in both versions.

Vicuna-1.5 is a version of the Llama-2 model \citep{touvron2023llama2openfoundation} fine-tuned on user--model conversations from ShareGPT. We use the version of Vicuna-1.5 with 7B parameters.

\subsubsection{Vision-language models}

To incorporate the visual data used in the picture paradigm, we also experiment with FLAVA \citep{singh2022flava}, LLaVA-1.5 \citep{liu2024improved,liu2024visual}, and Qwen2.5-VL \citep{Qwen-VL,Qwen2-VL,Qwen2.5-VL} models.

FLAVA is a multimodal model trained to align the text and image representations from two separate ViT encoders \citep{dosovitskiy2021image} via an extra transformer multimodal encoder. While all other models we consider are autoregressive, FLAVA's encoders are trained to optimize the masked modeling objective.

LLaVA-1.5 is a general-purpose visual and language understanding model. It is based on the Vicuna LM and additionally trained to take in the outputs of a visual encoder \citep[CLIP;][]{pmlr-v139-radford21a},  projected into the shared representation space through an MLP. We use the 7B version of this model in our experiments.

Qwen2.5-VL is a series of large multimodal models (based on Qwen2.5) optimized for visual understanding, including video comprehension, document parsing, and multilingual text recognition in images. We use the Qwen2.5-VL models with 3B and 7B parameters.

\subsection{LM representations}
\label{sec:embeddings}

To get one $d$-dimensional vector per stimulus (inputted into an LM as per \sref{sec:appendix-embedding}), we extract hidden states from all model layers and compare multiple pooling methods over the tokens in each image/sentence. 
For each layer, we take either the last-token hidden state or the mean hidden state over tokens; for FLAVA, we additionally consider the first-token (\texttt{[CLS]}) hidden state. FLAVA also uses independent unimodal encoders, so for multimodal inputs (\textsc{p}) we use their averaged hidden states as well as the multimodal fusion encoder output.

\subsection{Experiment 1: Brain encoding}
\label{sec:brain-enc}

In the brain encoding experiment, we fit a regression model that predicts a scalar activation value from a $d$-dimensional vector representation of a stimulus (Fig.~\ref{fig:1}b). Following \cite{toneva2019interpreting} and \cite{tuckute2024driving}, we add a ridge penalty since the number of predictors ($d$) can be quite large. 
The regression weights are determined as:

\begin{equation}
 \hat{\Vec{w}} = \arg\min_{\Vec{w} \in \mathbb{R}^d} \norm{\Vec{y} - \mX \Vec{w} }_2^2 + \alpha \norm{\Vec{w}}_2^2
\end{equation}

where $\mX \in \sR ^{n \times d}$ is the matrix of LM representations of the $n$ stimuli seen by the participant ($720 \leq n \leq 1080$) and  $\Vec{y} \in \sR^n$ is the vector of the corresponding brain activations.
The quality of fit is evaluated as the Pearson correlation coefficient between the vector of predicted brain responses $\hat{\Vec{y}} = \mX \hat{\Vec{w}}$ and the ground truth activations $\Vec{y}$. To obtain an unbiased estimate of this correlation, we perform five-fold cross-validation, fitting the regression model on 80\% of the stimulus--activation pairs at a time and measuring the correlation on the held-out 20\%; the final estimate is averaged over the five folds.
We report the performance only for the layer and token pooling (\Sref{sec:embeddings}) that yield the best predictivity $r(\hat{\Vec{y}}, \Vec{y})$ for the average participant's response in a given brain region, across all folds (\sref{sec:appendix-layers}).
Following \cite{tuckute2024driving},
we tune the regularization hyperparameter $\alpha \in \{ 10^{-30}, \ldots, 10^{28}, 10^{29} \}$ independently for each fold using leave-one-out cross-validation on the training portion (with \texttt{scikit-learn}; \sref{sec:appendix-sources}). 

\subsection{Experiment 2: Representational similarity alignment}
\label{sec:rsa}

To further explore differences among the models, we conduct an experiment where we measure the Representational Similarity Alignment \citep[RSA;][]{kriegeskorte2008representational} between each model and each of the selected brain regions. 
RSA, which compares the pairwise input representation \textit{distances} in the two spaces (Fig.~\ref{fig:1}c), is frequently used to evaluate brain--model similarity \citep{oota2024deep}. We focus on the concept representations, averaging the vectors for all sentences, pictures, and word clouds to obtain one model (per-layer) vector $\Vec{x}^i \in \mathbb{R}^d$ and one brain activation vector $\Vec{b}^i$ per concept $c_i, 1 \leq i \leq 180$. The elements of $\Vec{b}^i$ are responses to $c_i$ in each voxel in a chosen brain region $A$: $\Vec{b}^i=\{\beta^i_j\}_{j=1}^{|A|}$.
We compute two $180\times180$ matrices of pairwise Pearson correlation distances between these vectors: $1 - r(\Vec{x}^i, \Vec{x}^j)$ and $1 - r(\Vec{b}^i, \Vec{b}^j)$. Finally, we measure the Spearman correlation between the lower triangular portions of these matrices to evaluate how similarly this brain region and this model layer represent the 180 concepts. As before, we repeat this for each model layer and token pooling method and report only the results for the best setting per model.

\subsection{Neural metrics}
\label{sec:factors}

We investigate how LM--brain alignment correlates with two neural measures: (1) semantic consistency of a brain area (as described in \Sref{sec:metric}) and (2) the selectivity of this area for language processing, defined as the response to well-formed sentences compared to a perceptually matched control condition. Specifically, we leveraged data from an independent language localizer task \citep{fedorenko2010new} where
brain activitation is compared between two types of stimuli: English sentences and
unconnected sequences of non-words (e.g., \texttt{REDENTION ZOOD CRE...}). 
The ``sentences $>$ non-words'' contrast has been shown to reliably identify areas of the brain that are engaged in linguistic processing but not other functions (\citealp{fedorenko2011functional,benn2023language,chen2023human}; see \citealp{fedorenko2024language} for a review).
We quantify the language selectivity measure as the difference between a voxel's activations in the two conditions ($\Delta \beta_{\text{Sentences,Non-words}}$)
separately in each participant.

\section{Results}
\subsection{More semantically consistent voxels are better predicted by LMs}
\label{sec:exp1}

\begin{figure}[t]
\begin{center}
\includegraphics[width=\linewidth]{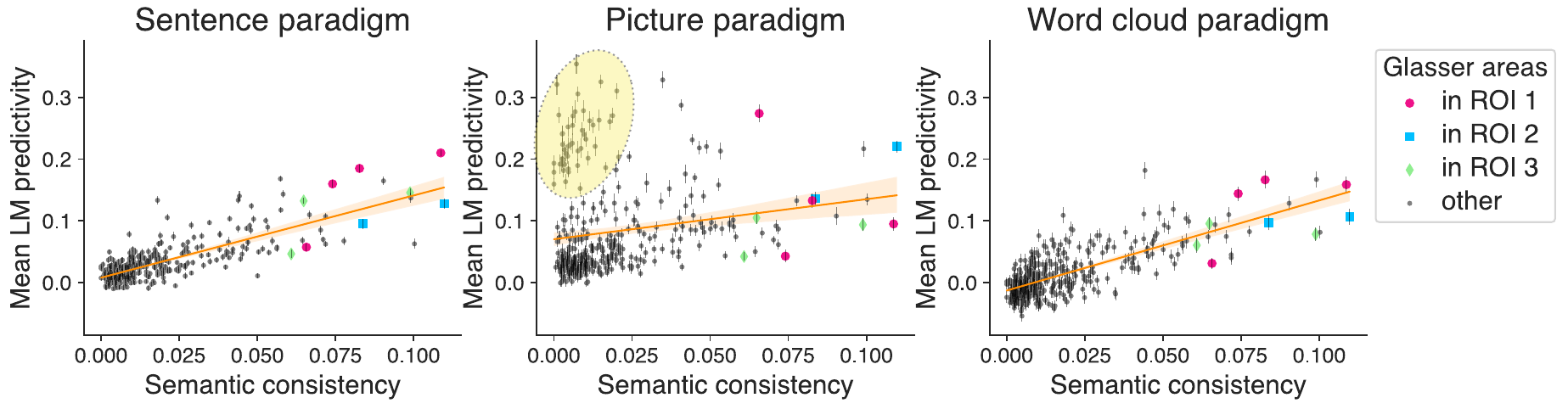}
\end{center}
\caption{\textbf{Predictivity vs. semantic consistency in \cite{glasser2016multi} anatomical areas (both hemispheres).} Each point corresponds to one area, and the areas that fall in the chosen semantically consistent ROIs (\Sref{sec:rois}) are marked by shape and color. Error bars show standard error over participants. All paradigms show a correlation between predictivity and semantic consistency, though for pictures it is skewed by visual cortex areas (circled).}
\label{fig:predict_glasser}
\end{figure}

Brain encoding experiments measure LM predictivity, i.e., the correlation between the voxel activations predicted from the LM representations and the ground truth activations (\sref{sec:brain-enc}). For the sentence and picture paradigms, we predict the brain activations for each stimulus individually ($n$=720--1080 stimuli per participant), but average the brain activations for word clouds since they contain the same words for the same concept ($n$=180).
This section reports all results averaged over the appropriate models (all models for \textsc{s} and \textsc{wc} paradigms, only vision-language models for \textsc{p}) since we did not see strong differences between individual models (individual plots included in Appendix \ref{appendix-indmodels}; see \Sref{sec:discussion} for discussion).

First, we verify that semantic consistency influences predictivity across the whole brain cortex. \Figref{fig:predict_glasser} shows the mean (over LMs and participants) predictivity across the 360 anatomical areas  \citep[180 in each hemisphere;][]{glasser2016multi} for each of the three paradigms. We see a strong positive correlation between the semantic consistency of an area\footnote{Measured as probability of significantly consistent voxels (\sref{sec:appendix-prob-map}) to match \sref{sec:rois}. Overall trends also hold when using the the raw value of $C$ (\sref{sec:appendix-raw-c}) or adjusting for inter-participant noise ceiling (\sref{sec:appendix-noise-ceiling}).}
and how well the activation in it can be predicted by LMs ($r[\text{\textsc{s}}]=0.79$, $r[\text{\textsc{wc}}]=0.74$). The correlation is lower for the picture paradigm ($r[\text{\textsc{p}}] = 0.17$) because of a cluster of visual cortex areas (circled in yellow): they encode images (hence the high VLM alignment) but not necessarily concepts. Taken together, these findings show that a brain region is better predicted if it responds more consistently to concepts, irrespective of modality and paradigm.

\begin{figure}[t]
\begin{center}
\includegraphics[width=\linewidth]{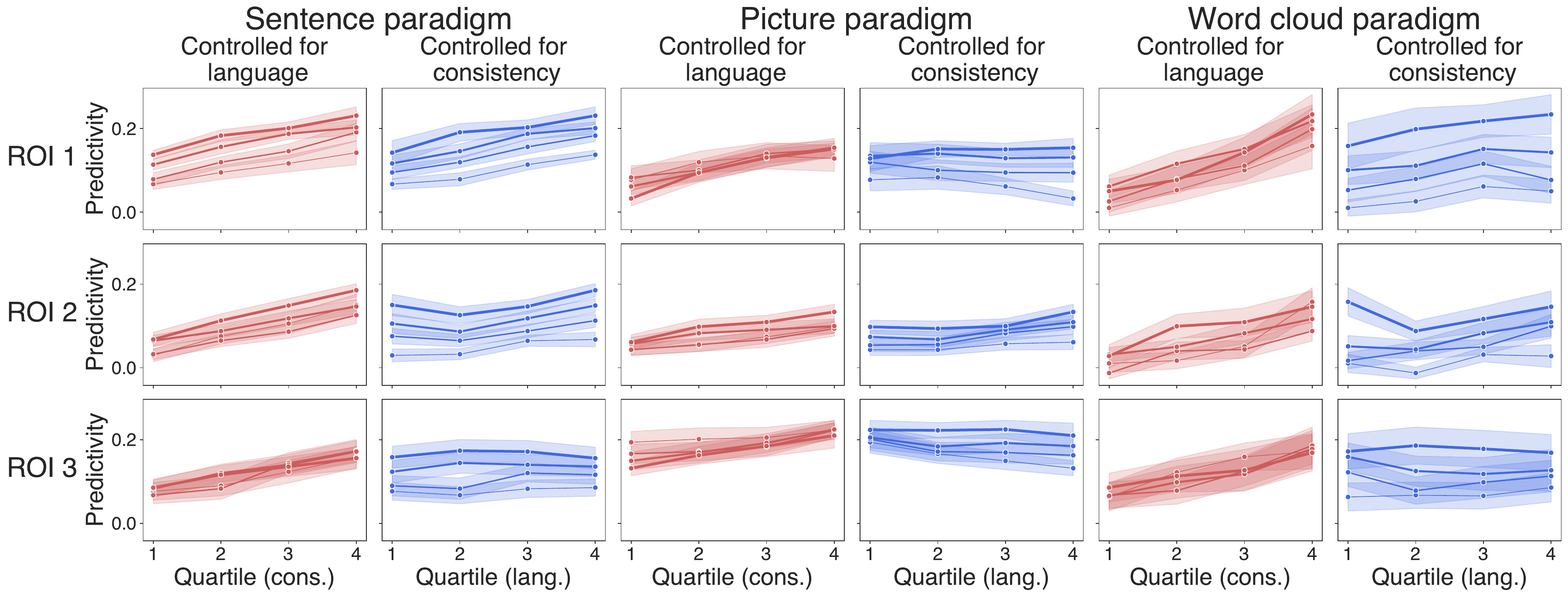}
\end{center}
\caption{\textbf{Mean LM predictivity by quartile for each ROI and paradigm.} Columns 1, 3, and 5 show how predictivity in each ROI changes across voxel quartiles by semantic consistency, with each \textcolor{indianred}{red} line corresponding to one language selectivity quartile. Columns 2, 4, and 6 show how predictivity changes across voxel quartiles by language selectivity, with each \textcolor{royalblue}{blue} line corresponding to one semantic consistency quartile. The thickness of the line corresponds to the quartile (thicker=higher), and the error intervals show standard error across participants. While ROI 1 and ROI 2 (rows 1 and 2) show a positive correlation with both the semantic consistency and the language selectivity (albeit to a lesser extent), the predictivity in the ventral ROI 3 does not correlate with the language selectivity.
}
\label{fig:quartiles}
\end{figure}

Second, we evaluate how the brain encoding performance in our three ROIs correlates with the two brain metrics of interest (\sref{sec:factors}). Each participant's brain voxels in each ROI are divided into bins (quartiles) by either semantic consistency $(1 \leq b_C\leq 4)$ or language selectivity $(1 \leq b_L\leq 4)$, resulting in 16 $(b_C, b_L)$ bins total. Each plot in \Figref{fig:quartiles} keeps one of these metrics fixed while varying the other: for example, in columns 1, 3, and 5 each red line corresponds to all voxels of the same $b_L$, while the points on the line represent voxels in $(1, b_L)$, $(2, b_L)$, $(3, b_L)$, and $(4, b_L)$ respectively. Similarly, the plots in columns 2, 4, and 6 group lines by $b_C$, and the x-axis steps correspond to $b_L \in \{1,2,3,4\}$ respectively. The y-axis in each plot shows the predictivity, averaged over participants and LMs.

In all ROIs and paradigms, predictivity rises monotonically across semantic consistency quartiles $b_{C}$ while $b_L$ is held fixed (red lines in columns 1, 3, 5; mean $r=0.40 \pm 0.01$). The correlation with the language quartile $b_L$ when controlling for $b_C$ is less clear: while there is some increase in ROI 1 and 2 for text-based paradigms (blue lines in rows 1, 2, columns 2, 6; mean $r = 0.26 \pm 0.04$), ROI 3 (row 3, columns 2, 4, 6; mean $r = 0.01 \pm 0.02$) and the picture paradigm overall (column 4; mean $r = -0.01 \pm 0.04$) display no such dependency. 
ROI 3, involved in visual but not language processing, demonstrates that semantic consistency drives predictivity even decoupled from language: $r_C = 0.33 \pm 0.03, r_L = 0.01 \pm 0.02$.

\subsection{LMs and VLMs share representational geometry with semantic brain regions
}
\label{sec:exp2}

\begin{figure}[t]
\begin{center}
\includegraphics[width=0.9\linewidth]{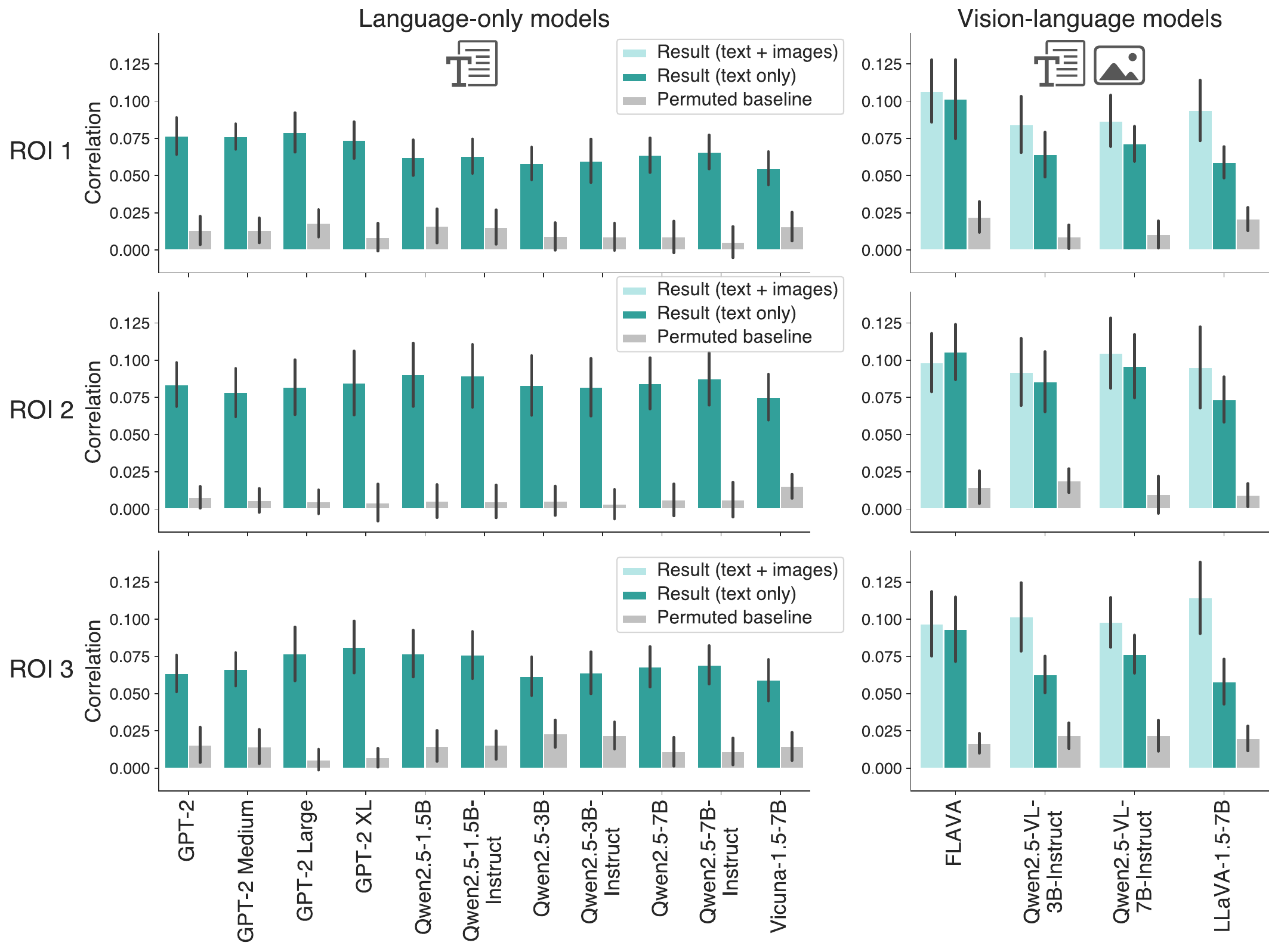}
\caption{\textbf{Concept-level RSA for each LM and brain ROI.} RSA quantifies correlations between pairwise concept distance matrices. Each concept's representations are averaged over stimuli (sentences and word clouds for text-only condition, all paradigms for text + image). 
Shuffled baseline included for comparison. Error bars show SEM across participants.
}
\label{fig:rsa}
\end{center}
\end{figure}

\Figref{fig:rsa} shows the RSA correlation between each model and each ROI, reported for the most aligned layer in each model. 
We additionally include a baseline in which we shuffle the brain's concept representations $\{\Vec{b}^i\}_{i=1}^{180}$, so that the pairwise concept distances are not matched between the brain and the model; each baseline is reported for its own best layer.

In the three ROIs associated with high semantic consistency (rows in \Figref{fig:rsa}), the alignment in all models is significantly higher than the baseline. We do not see a clear trend for model size (within the GPT-2 or Qwen2.5 families; cf. \citealp{schrimpf2021neural}) or a noticeable effect of instruction tuning (between Qwen2.5 and Qwen2.5-Instruct models of the same size; cf. \citealp{aw2024instruction}). 
While language-only models (left set of bars) only represent the textual stimuli (\textsc{s} and \textsc{wc}), for vision-language models (right set of bars) we compare the same setting with an all-paradigm average. The text-only performance is comparable in VLMs and their base LM counterparts (LLaVA vs. Vicuna, Qwen2.5-VL vs. Qwen2.5). Interestingly, the addition of multimodal stimuli increases alignment, most notably in the ventrotemporal ROI 3---a region adjacent to areas associated with high-level vision \citep[e.g.,][]{kanwisher1997fusiform} as well as the visual word form area and the basal language areas \citep{li2024demystifying}.

\section{Discussion and conclusion}
\label{sec:discussion}

We evaluated model--brain alignment for 15 transformer language and vision-language models in a brain encoding experiment. To do so, we introduced a new metric that identifies brain voxels with consistent responses to conceptual content across different paradigms, based on fMRI data from multimodal stimuli. We show that the more concept-consistent the voxels are, the better they are predicted by LM representations.
In line with prior work \citep{ayeshlanguage}, we also find that LM predictivity is correlated with language selectivity in the regions overlapping with the canonical language areas of \citeauthor{fedorenko2024language} (superior temporal ROI 1) or adjacent to them (inferior frontal ROI 2).

Aiming to extract modality-independent conceptual representations of the stimuli from the participant's brain (rather than purely linguistic/visual ones), we target a novel set of ROIs in our alignment experiments. We focus on three brain regions that show the most consistent preferences for certain concepts, regardless of presentation paradigm (as measured by our proposed semantic consistency metric). These regions are distinct from the established brain networks typically used to evaluate LM--brain alignment, such as the language network (\citealt{fedorenko2010new}; evidenced by two of the ROIs showing little to no response to the language localizer; see \sref{sec:appendix-lang-response}). The temporal ROI 1 overlaps both with the language network and with the areas where evidence of amodal semantic processing was found previously \citep[Ch. 5]{wurm2019distinct,popham2021visual,ivanova2022role}---we hypothesize that ROI 1 may serve as a gateway between the language system and the more abstract semantic areas. For consistency with prior work, we include a comparison of the brain encoding performance in our ROIs and in the language network parcels (\sref{sec:appendix-encoding-lang-rois}).

We do not see strong differences in brain encoding performance between individual models. We attribute that to the flexibility of our brain encoding pipeline, based on that of \cite{tuckute2024driving}: it not only chooses the most predictive layer for each model, ROI, and paradigm, but also tunes the regularization hyperparameter individually for each cross-validation fold at inference time. While it yields the best performance for each model, it obscures the differences between them, so we perform an additional comparison using RSA. We find significant alignment between the representational spaces of all models and the semantically consistent brain regions, but do not observe the trends noted in prior work: in our experiment, RSA alignment does not increase from smaller to larger models in the same architecture \citep{schrimpf2021neural} or with additional instruction tuning \citep{aw2024instruction}.

Past work has found that responses to images in high-level visual cortical areas---which overlap with the ventral ROI 3---are successfully predicted from LM embeddings of their descriptions \citep{doerig2022visual}. \cite{conwell2023unreasonable} show that much of this alignment is explained by the concepts (objects and agents) present in the image. Together with our consistency evaluation, these results suggest that certain conceptual information is retained in these regions---and stronger LM alignment with semantically consistent brain areas can be viewed as evidence for these models' ability to capture cross-modal conceptual meaning.

\section*{Acknowledgments}
We would like to acknowledge the Athinoula A. Martinos Imaging Center at the McGovern Institute for Brain Research at MIT and its support team (Steve Shannon and Atsushi Takahashi).
We thank Francisco Pereira and Juniper Pritchett for developing the experimental materials and collecting, preprocessing, and analyzing the fMRI data. EF was supported by NIH award NS121471 from NINDS and research funds from the McGovern Institute for Brain Research, MIT School of Science, the Simons Center for the Social Brain, and MIT’s Quest for Intelligence.
We thank EvLab members for help with data processing and visualization and the anonymous reviewers for their valuable feedback.

\bibliography{colm2025_conference}

\begin{thebibliography}{88}
\providecommand{\natexlab}[1]{#1}
\providecommand{\url}[1]{\texttt{#1}}
\expandafter\ifx\csname urlstyle\endcsname\relax
  \providecommand{\doi}[1]{doi: #1}\else
  \providecommand{\doi}{doi: \begingroup \urlstyle{rm}\Url}\fi

\bibitem[Abdou et~al.(2021)Abdou, Kulmizev, Hershcovich, Frank, Pavlick, and S{\o}gaard]{abdou-etal-2021-language}
Mostafa Abdou, Artur Kulmizev, Daniel Hershcovich, Stella Frank, Ellie Pavlick, and Anders S{\o}gaard.
\newblock Can language models encode perceptual structure without grounding? {A} case study in color.
\newblock In \emph{Proceedings of the 25th Conference on Computational Natural Language Learning}, pp.\  109--132. Association for Computational Linguistics, 2021.
\newblock URL \url{https://aclanthology.org/2021.conll-1.9/}.

\bibitem[Antonello et~al.(2023)Antonello, Vaidya, and Huth]{antonello2023scaling}
Richard Antonello, Aditya Vaidya, and Alexander Huth.
\newblock Scaling laws for language encoding models in {fMRI}.
\newblock In \emph{Advances in Neural Information Processing Systems}, volume~36. Curran Associates, Inc., 2023.
\newblock URL \url{https://proceedings.neurips.cc/paper_files/paper/2023/file/4533e4a352440a32558c1c227602c323-Paper-Conference.pdf}.

\bibitem[Aw \& Toneva(2023)Aw and Toneva]{aw2023training}
Khai~Loong Aw and Mariya Toneva.
\newblock Training language models to summarize narratives improves brain alignment.
\newblock In \emph{The Eleventh International Conference on Learning Representations}, 2023.
\newblock URL \url{https://openreview.net/pdf?id=KzkLAE49H9b}.

\bibitem[Aw et~al.(2024)Aw, Montariol, AlKhamissi, Schrimpf, and Bosselut]{aw2024instruction}
Khai~Loong Aw, Syrielle Montariol, Badr AlKhamissi, Martin Schrimpf, and Antoine Bosselut.
\newblock Instruction-tuning aligns {LLMs} to the human brain.
\newblock In \emph{First Conference on Language Modeling}, 2024.
\newblock URL \url{https://openreview.net/pdf?id=nXNN0x4wbl}.

\bibitem[Ayesh et~al.(2024)Ayesh, Jain, St~Luce, Huth, and Ivanova]{ayeshlanguage}
Eyas Ayesh, Shailee Jain, Josleen St~Luce, Alexander Huth, and Anna~A. Ivanova.
\newblock The language network occupies a privileged position among all brain voxels predicted by a language-based encoding model.
\newblock In \emph{Conference on Computational Cognitive Neuroscience}, 2024.
\newblock URL \url{https://2024.ccneuro.org/pdf/450_Paper_authored_Authored---Language-network-occupies-a-privileged-position.pdf}.

\bibitem[Bai et~al.(2023{\natexlab{a}})Bai, Bai, Chu, Cui, Dang, Deng, Fan, Ge, Han, Huang, Hui, Ji, Li, Lin, Lin, Liu, Liu, Lu, Lu, Ma, Men, Ren, Ren, Tan, Tan, Tu, Wang, Wang, Wang, Wu, Xu, Xu, Yang, Yang, Yang, Yang, Yao, Yu, Yuan, Yuan, Zhang, Zhang, Zhang, Zhang, Zhou, Zhou, Zhou, and Zhu]{qwen}
Jinze Bai, Shuai Bai, Yunfei Chu, Zeyu Cui, Kai Dang, Xiaodong Deng, Yang Fan, Wenbin Ge, Yu~Han, Fei Huang, Binyuan Hui, Luo Ji, Mei Li, Junyang Lin, Runji Lin, Dayiheng Liu, Gao Liu, Chengqiang Lu, Keming Lu, Jianxin Ma, Rui Men, Xingzhang Ren, Xuancheng Ren, Chuanqi Tan, Sinan Tan, Jianhong Tu, Peng Wang, Shijie Wang, Wei Wang, Shengguang Wu, Benfeng Xu, Jin Xu, An~Yang, Hao Yang, Jian Yang, Shusheng Yang, Yang Yao, Bowen Yu, Hongyi Yuan, Zheng Yuan, Jianwei Zhang, Xingxuan Zhang, Yichang Zhang, Zhenru Zhang, Chang Zhou, Jingren Zhou, Xiaohuan Zhou, and Tianhang Zhu.
\newblock Qwen technical report.
\newblock \emph{arXiv preprint arXiv:2309.16609}, 2023{\natexlab{a}}.
\newblock URL \url{https://arxiv.org/abs/2309.16609}.

\bibitem[Bai et~al.(2023{\natexlab{b}})Bai, Bai, Yang, Wang, Tan, Wang, Lin, Zhou, and Zhou]{Qwen-VL}
Jinze Bai, Shuai Bai, Shusheng Yang, Shijie Wang, Sinan Tan, Peng Wang, Junyang Lin, Chang Zhou, and Jingren Zhou.
\newblock Qwen-{VL}: A versatile vision-language model for understanding, localization, text reading, and beyond.
\newblock \emph{arXiv preprint arXiv:2308.12966}, 2023{\natexlab{b}}.
\newblock URL \url{https://arxiv.org/abs/2308.12966}.

\bibitem[Bai et~al.(2025)Bai, Chen, Liu, Wang, Ge, Song, Dang, Wang, Wang, Tang, Zhong, Zhu, Yang, Li, Wan, Wang, Ding, Fu, Xu, Ye, Zhang, Xie, Cheng, Zhang, Yang, Xu, and Lin]{Qwen2.5-VL}
Shuai Bai, Keqin Chen, Xuejing Liu, Jialin Wang, Wenbin Ge, Sibo Song, Kai Dang, Peng Wang, Shijie Wang, Jun Tang, Humen Zhong, Yuanzhi Zhu, Mingkun Yang, Zhaohai Li, Jianqiang Wan, Pengfei Wang, Wei Ding, Zheren Fu, Yiheng Xu, Jiabo Ye, Xi~Zhang, Tianbao Xie, Zesen Cheng, Hang Zhang, Zhibo Yang, Haiyang Xu, and Junyang Lin.
\newblock Qwen2.5-{VL} technical report.
\newblock \emph{arXiv preprint arXiv:2502.13923}, 2025.
\newblock URL \url{https://arxiv.org/abs/2502.13923}.

\bibitem[Bashivan et~al.(2019)Bashivan, Kar, and DiCarlo]{bashivan2019neural}
Pouya Bashivan, Kohitij Kar, and James~J. DiCarlo.
\newblock Neural population control via deep image synthesis.
\newblock \emph{Science}, 364\penalty0 (6439):\penalty0 eaav9436, 2019.
\newblock URL \url{https://www.science.org/doi/10.1126/science.aav9436}.

\bibitem[Bavaresco \& Fern{\'a}ndez(2025)Bavaresco and Fern{\'a}ndez]{bavaresco-fernandez-2025-experiential}
Anna Bavaresco and Raquel Fern{\'a}ndez.
\newblock Experiential semantic information and brain alignment: Are multimodal models better than language models?
\newblock In \emph{Proceedings of the 29th Conference on Computational Natural Language Learning}, pp.\  141--155. Association for Computational Linguistics, 2025.
\newblock URL \url{https://aclanthology.org/2025.conll-1.10/}.

\bibitem[Bavaresco et~al.(2024)Bavaresco, de~Heer~Kloots, Pezzelle, and Fern{\'a}ndez]{bavaresco2024modelling}
Anna Bavaresco, Marianne de~Heer~Kloots, Sandro Pezzelle, and Raquel Fern{\'a}ndez.
\newblock Modelling multimodal integration in human concept processing with vision-and-language models.
\newblock \emph{arXiv preprint arXiv:2407.17914}, 2024.
\newblock URL \url{https://arxiv.org/abs/2407.17914}.

\bibitem[Bender \& Koller(2020)Bender and Koller]{bender-koller-2020-climbing}
Emily~M. Bender and Alexander Koller.
\newblock Climbing towards {NLU}: {On} meaning, form, and understanding in the age of data.
\newblock In \emph{Proceedings of the 58th Annual Meeting of the Association for Computational Linguistics}, pp.\  5185--5198. Association for Computational Linguistics, 2020.
\newblock URL \url{https://aclanthology.org/2020.acl-main.463/}.

\bibitem[Benn et~al.(2023)Benn, Ivanova, Clark, Mineroff, Seikus, Silva, Varley, and Fedorenko]{benn2023language}
Yael Benn, Anna~A. Ivanova, Oliver Clark, Zachary Mineroff, Chloe Seikus, Jack~Santos Silva, Rosemary Varley, and Evelina Fedorenko.
\newblock The language network is not engaged in object categorization.
\newblock \emph{Cerebral Cortex}, 33\penalty0 (19):\penalty0 10380--10400, 2023.
\newblock URL \url{https://academic.oup.com/cercor/article-pdf/33/19/10380/51765576/bhad289.pdf}.

\bibitem[Caucheteux \& King(2022)Caucheteux and King]{caucheteux2022brains}
Charlotte Caucheteux and Jean-R{\'e}mi King.
\newblock Brains and algorithms partially converge in natural language processing.
\newblock \emph{Communications Biology}, 5\penalty0 (1):\penalty0 134, 2022.
\newblock URL \url{https://www.nature.com/articles/s42003-022-03036-1}.

\bibitem[Chen et~al.(2023)Chen, Affourtit, Ryskin, Regev, Norman-Haignere, Jouravlev, Malik-Moraleda, Kean, Varley, and Fedorenko]{chen2023human}
Xuanyi Chen, Josef Affourtit, Rachel Ryskin, Tamar~I. Regev, Samuel Norman-Haignere, Olessia Jouravlev, Saima Malik-Moraleda, Hope Kean, Rosemary Varley, and Evelina Fedorenko.
\newblock The human language system, including its inferior frontal component in {``Broca’s area,''} does not support music perception.
\newblock \emph{Cerebral Cortex}, 33\penalty0 (12):\penalty0 7904--7929, 2023.
\newblock URL \url{https://academic.oup.com/cercor/article-pdf/33/12/7904/51643838/bhad087.pdf}.

\bibitem[Chen \& Bonner(2024)Chen and Bonner]{chen2024universal}
Zirui Chen and Michael~F. Bonner.
\newblock Universal dimensions of visual representation.
\newblock \emph{arXiv preprint arXiv:2408.12804}, 2024.
\newblock URL \url{https://arxiv.org/abs/2408.12804}.

\bibitem[Chiang et~al.(2023)Chiang, Li, Lin, Sheng, Wu, Zhang, Zheng, Zhuang, Zhuang, Gonzalez, Stoica, and Xing]{vicuna2023}
Wei-Lin Chiang, Zhuohan Li, Zi~Lin, Ying Sheng, Zhanghao Wu, Hao Zhang, Lianmin Zheng, Siyuan Zhuang, Yonghao Zhuang, Joseph~E. Gonzalez, Ion Stoica, and Eric~P. Xing.
\newblock Vicuna: An open-source chatbot impressing {GPT-4} with 90\%* {ChatGPT} quality, March 2023.
\newblock URL \url{https://lmsys.org/blog/2023-03-30-vicuna/}.

\bibitem[Conwell et~al.(2023)Conwell, Prince, Alvarez, and Konkle]{conwell2023unreasonable}
Colin Conwell, Jacob Prince, George Alvarez, and Talia Konkle.
\newblock The unreasonable effectiveness of word models in predicting high-level visual cortex responses to natural images.
\newblock In \emph{Conference on Computational Cognitive Neuroscience}, 2023.
\newblock URL \url{https://2023.ccneuro.org/proceedings/0000564.pdf?s=W&pn=1642}.

\bibitem[Conwell et~al.(2024)Conwell, Prince, Kay, Alvarez, and Konkle]{conwell2024large}
Colin Conwell, Jacob~S. Prince, Kendrick~N. Kay, George~A. Alvarez, and Talia Konkle.
\newblock A large-scale examination of inductive biases shaping high-level visual representation in brains and machines.
\newblock \emph{Nature Communications}, 15\penalty0 (1):\penalty0 9383, 2024.
\newblock URL \url{https://www.nature.com/articles/s41467-024-53147-y}.

\bibitem[Doerig et~al.(2022)Doerig, Kietzmann, Allen, Wu, Naselaris, Kay, and Charest]{doerig2022visual}
Adrien Doerig, Tim~C Kietzmann, Emily Allen, Yihan Wu, Thomas Naselaris, Kendrick Kay, and Ian Charest.
\newblock Semantic scene descriptions as an objective of human vision.
\newblock \emph{arXiv preprint arXiv:2209.11737}, 2022.
\newblock URL \url{https://arxiv.org/abs/2209.11737}.

\bibitem[Dosovitskiy et~al.(2021)Dosovitskiy, Beyer, Kolesnikov, Weissenborn, Zhai, Unterthiner, Dehghani, Minderer, Heigold, Gelly, Uszkoreit, and Houlsby]{dosovitskiy2021image}
Alexey Dosovitskiy, Lucas Beyer, Alexander Kolesnikov, Dirk Weissenborn, Xiaohua Zhai, Thomas Unterthiner, Mostafa Dehghani, Matthias Minderer, Georg Heigold, Sylvain Gelly, Jakob Uszkoreit, and Neil Houlsby.
\newblock An image is worth 16 $\times$ 16 words: {T}ransformers for image recognition at scale.
\newblock In \emph{International Conference on Learning Representations}, 2021.
\newblock URL \url{https://openreview.net/pdf?id=YicbFdNTTy}.

\bibitem[Du et~al.(2025)Du, Fu, Wen, Sun, Peng, Wei, Gao, Wang, Zhang, Li, et~al.]{du2025human}
Changde Du, Kaicheng Fu, Bincheng Wen, Yi~Sun, Jie Peng, Wei Wei, Ying Gao, Shengpei Wang, Chuncheng Zhang, Jinpeng Li, et~al.
\newblock Human-like object concept representations emerge naturally in multimodal large language models.
\newblock \emph{Nature Machine Intelligence}, pp.\  1--16, 2025.
\newblock URL \url{https://www.nature.com/articles/s42256-025-01049-z}.

\bibitem[Fedorenko et~al.(2010)Fedorenko, Hsieh, Nieto-Casta{\~n}{\'o}n, Whitfield-Gabrieli, and Kanwisher]{fedorenko2010new}
Evelina Fedorenko, Po-Jang Hsieh, Alfonso Nieto-Casta{\~n}{\'o}n, Susan Whitfield-Gabrieli, and Nancy Kanwisher.
\newblock New method for {fMRI} investigations of language: {D}efining {ROIs} functionally in individual subjects.
\newblock \emph{Journal of Neurophysiology}, 104\penalty0 (2):\penalty0 1177--1194, 2010.
\newblock URL \url{https://journals.physiology.org/doi/prev/20100421-aop/pdf/10.1152/jn.00032.2010}.

\bibitem[Fedorenko et~al.(2011)Fedorenko, Behr, and Kanwisher]{fedorenko2011functional}
Evelina Fedorenko, Michael~K. Behr, and Nancy Kanwisher.
\newblock Functional specificity for high-level linguistic processing in the human brain.
\newblock \emph{Proceedings of the National Academy of Sciences}, 108\penalty0 (39):\penalty0 16428--16433, 2011.
\newblock URL \url{https://www.pnas.org/doi/abs/10.1073/pnas.1112937108}.

\bibitem[Fedorenko et~al.(2024)Fedorenko, Ivanova, and Regev]{fedorenko2024language}
Evelina Fedorenko, Anna~A. Ivanova, and Tamar~I. Regev.
\newblock The language network as a natural kind within the broader landscape of the human brain.
\newblock \emph{Nature Reviews Neuroscience}, 25\penalty0 (5):\penalty0 289--312, 2024.
\newblock URL \url{https://www.nature.com/articles/s41583-024-00802-4}.

\bibitem[Frisby et~al.(2023)Frisby, Halai, Cox, Ralph, and Rogers]{frisby2023decoding}
Saskia~L. Frisby, Ajay~D. Halai, Christopher~R. Cox, Matthew A.~Lambon Ralph, and Timothy~T. Rogers.
\newblock Decoding semantic representations in mind and brain.
\newblock \emph{Trends in cognitive sciences}, 27\penalty0 (3):\penalty0 258--281, 2023.
\newblock URL \url{https://www.cell.com/trends/cognitive-sciences/fulltext/S1364-6613(22)00323-0}.

\bibitem[Glasser et~al.(2016)Glasser, Coalson, Robinson, Hacker, Harwell, Yacoub, Ugurbil, Andersson, Beckmann, Jenkinson, Smith, and Van~Essen]{glasser2016multi}
Matthew~F. Glasser, Timothy~S. Coalson, Emma~C. Robinson, Carl~D. Hacker, John Harwell, Essa Yacoub, Kamil Ugurbil, Jesper Andersson, Christian~F. Beckmann, Mark Jenkinson, Stephen~M. Smith, and David~C. Van~Essen.
\newblock A multi-modal parcellation of human cerebral cortex.
\newblock \emph{Nature}, 536\penalty0 (7615):\penalty0 171--178, 2016.
\newblock URL \url{https://www.nature.com/articles/nature18933}.

\bibitem[Goldstein et~al.(2022)Goldstein, Zada, Buchnik, Schain, Price, Aubrey, Nastase, Feder, Emanuel, Cohen, Jansen, Gazula, Choe, Rao, Kim, Casto, Fanda, Doyle, Friedman, Dugan, Melloni, Reichart, Devore, Flinker, Hasenfratz, Levy, Hassidim, Brenner, Matias, Norman, Devinsky, and Hasson]{goldstein2022shared}
Ariel Goldstein, Zaid Zada, Eliav Buchnik, Mariano Schain, Amy Price, Bobbi Aubrey, Samuel~A. Nastase, Amir Feder, Dotan Emanuel, Alon Cohen, Aren Jansen, Harshvardhan Gazula, Gina Choe, Aditi Rao, Catherine Kim, Colton Casto, Lora Fanda, Werner Doyle, Daniel Friedman, Patricia Dugan, Lucia Melloni, Roi Reichart, Sasha Devore, Adeen Flinker, Liat Hasenfratz, Omer Levy, Avinatan Hassidim, Michael Brenner, Yossi Matias, Kenneth~A. Norman, Orrin Devinsky, and Uri Hasson.
\newblock Shared computational principles for language processing in humans and deep language models.
\newblock \emph{Nature Neuroscience}, 25\penalty0 (3):\penalty0 369--380, 2022.
\newblock URL \url{https://www.nature.com/articles/s41593-022-01026-4}.

\bibitem[Harris et~al.(2020)Harris, Millman, van~der Walt, Gommers, Virtanen, Cournapeau, Wieser, Taylor, Berg, Smith, Kern, Picus, Hoyer, van Kerkwijk, Brett, Haldane, del Río, Wiebe, Peterson, Gérard-Marchant, Sheppard, Reddy, Weckesser, Abbasi, Gohlke, and Oliphant]{numpy}
Charles~R. Harris, K.~Jarrod Millman, Stéfan~J. van~der Walt, Ralf Gommers, Pauli Virtanen, David Cournapeau, Eric Wieser, Julian Taylor, Sebastian Berg, Nathaniel~J. Smith, Robert Kern, Matti Picus, Stephan Hoyer, Marten~H. van Kerkwijk, Matthew Brett, Allan Haldane, Jaime~Fernández del Río, Mark Wiebe, Pearu Peterson, Pierre Gérard-Marchant, Kevin Sheppard, Tyler Reddy, Warren Weckesser, Hameer Abbasi, Christoph Gohlke, and Travis~E. Oliphant.
\newblock Array programming with {NumPy}.
\newblock \emph{Nature}, 585\penalty0 (7825):\penalty0 357--362, 2020.
\newblock URL \url{https://doi.org/10.1038/s41586-020-2649-2}.

\bibitem[Horikawa \& Kamitani(2017)Horikawa and Kamitani]{horikawa2017generic}
Tomoyasu Horikawa and Yukiyasu Kamitani.
\newblock Generic decoding of seen and imagined objects using hierarchical visual features.
\newblock \emph{Nature Communications}, 8\penalty0 (1):\penalty0 15037, 2017.
\newblock URL \url{https://www.nature.com/articles/ncomms15037}.

\bibitem[Horn(2016)]{Horn2016}
Andreas Horn.
\newblock {HCP-MMP1.0 projected on MNI2009a GM (volumetric) in NIfTI format}, August 2016.
\newblock URL \url{https://figshare.com/articles/dataset/HCP-MMP1_0_projected_on_MNI2009a_GM_volumetric_in_NIfTI_format/3501911}.

\bibitem[Hosseini et~al.(2024)Hosseini, Casto, Zaslavsky, Conwell, Richardson, and Fedorenko]{hosseini2024universality}
Eghbal Hosseini, Colton Casto, Noga Zaslavsky, Colin Conwell, Mark Richardson, and Evelina Fedorenko.
\newblock Universality of representation in biological and artificial neural networks.
\newblock \emph{bioRxiv}, 2024.
\newblock URL \url{https://www.biorxiv.org/content/10.1101/2024.12.26.629294v1.abstract}.

\bibitem[Huh et~al.(2024)Huh, Cheung, Wang, and Isola]{huh2024position}
Minyoung Huh, Brian Cheung, Tongzhou Wang, and Phillip Isola.
\newblock Position: The platonic representation hypothesis.
\newblock In \emph{Proceedings of the 41st International Conference on Machine Learning}, volume 235 of \emph{Proceedings of Machine Learning Research}. PMLR, 2024.
\newblock URL \url{https://proceedings.mlr.press/v235/huh24a.html}.

\bibitem[Ivanova(2022)]{ivanova2022role}
Anna~Alexandrovna Ivanova.
\newblock \emph{The role of language in broader human cognition: evidence from neuroscience}.
\newblock PhD thesis, Massachusetts Institute of Technology, 2022.
\newblock URL \url{https://dspace.mit.edu/handle/1721.1/147484}.

\bibitem[Kaniuth \& Hebart(2022)Kaniuth and Hebart]{kaniuth2022feature}
Philipp Kaniuth and Martin~N. Hebart.
\newblock Feature-reweighted representational similarity analysis: A method for improving the fit between computational models, brains, and behavior.
\newblock \emph{NeuroImage}, 257:\penalty0 119294, 2022.
\newblock URL \url{https://doi.org/10.1016/j.neuroimage.2022.119294}.

\bibitem[Kanwisher et~al.(1997)Kanwisher, McDermott, and Chun]{kanwisher1997fusiform}
Nancy Kanwisher, Josh McDermott, and Marvin~M. Chun.
\newblock The fusiform face area: a module in human extrastriate cortex specialized for face perception.
\newblock \emph{Journal of Neuroscience}, 17\penalty0 (11):\penalty0 4302--4311, 1997.
\newblock URL \url{https://www.jneurosci.org/content/17/11/4302}.

\bibitem[Karamolegkou et~al.(2023)Karamolegkou, Abdou, and S{\o}gaard]{karamolegkou-etal-2023-mapping}
Antonia Karamolegkou, Mostafa Abdou, and Anders S{\o}gaard.
\newblock Mapping brains with language models: A survey.
\newblock In \emph{Findings of the Association for Computational Linguistics: ACL 2023}, pp.\  9748--9762. Association for Computational Linguistics, 2023.
\newblock URL \url{https://aclanthology.org/2023.findings-acl.618/}.

\bibitem[Kiefer \& Pulverm{\"u}ller(2012)Kiefer and Pulverm{\"u}ller]{kiefer2012conceptual}
Markus Kiefer and Friedemann Pulverm{\"u}ller.
\newblock Conceptual representations in mind and brain: Theoretical developments, current evidence and future directions.
\newblock \emph{Cortex}, 48\penalty0 (7):\penalty0 805--825, 2012.
\newblock URL \url{https://www.sciencedirect.com/science/article/abs/pii/S0010945211001018}.

\bibitem[Kriegeskorte et~al.(2008)Kriegeskorte, Mur, and Bandettini]{kriegeskorte2008representational}
Nikolaus Kriegeskorte, Marieke Mur, and Peter Bandettini.
\newblock Representational similarity analysis-connecting the branches of systems neuroscience.
\newblock \emph{Frontiers in Systems Neuroscience}, 2:\penalty0 249, 2008.
\newblock URL \url{https://www.frontiersin.org/journals/systems-neuroscience/articles/10.3389/neuro.06.004.2008/full}.

\bibitem[Kumar et~al.(2024)Kumar, Sumers, Yamakoshi, Goldstein, Hasson, Norman, Griffiths, Hawkins, and Nastase]{kumar2024shared}
Sreejan Kumar, Theodore~R. Sumers, Takateru Yamakoshi, Ariel Goldstein, Uri Hasson, Kenneth~A. Norman, Thomas~L. Griffiths, Robert~D. Hawkins, and Samuel~A. Nastase.
\newblock Shared functional specialization in transformer-based language models and the human brain.
\newblock \emph{Nature Communications}, 15\penalty0 (1):\penalty0 5523, 2024.
\newblock URL \url{https://www.nature.com/articles/s41467-024-49173-5}.

\bibitem[Li et~al.(2024{\natexlab{a}})Li, Karamolegkou, Kementchedjhieva, Abdou, and S{\o}gaard]{pmlr-v228-li24a}
Jiaang Li, Antonia Karamolegkou, Yova Kementchedjhieva, Mostafa Abdou, and Anders S{\o}gaard.
\newblock Structural similarities between language models and neural response measurements.
\newblock In \emph{Proceedings of the 2nd NeurIPS Workshop on Symmetry and Geometry in Neural Representations}, volume 228 of \emph{Proceedings of Machine Learning Research}, pp.\  346--365. PMLR, 16 Dec 2024{\natexlab{a}}.
\newblock URL \url{https://proceedings.mlr.press/v228/li24a.html}.

\bibitem[Li et~al.(2024{\natexlab{b}})Li, Kementchedjhieva, Fierro, and S{\o}gaard]{li2024vision}
Jiaang Li, Yova Kementchedjhieva, Constanza Fierro, and Anders S{\o}gaard.
\newblock Do vision and language models share concepts? {A} vector space alignment study.
\newblock \emph{Transactions of the Association for Computational Linguistics}, 12:\penalty0 1232--1249, 2024{\natexlab{b}}.
\newblock URL \url{https://doi.org/10.1162/tacl_a_00698}.

\bibitem[Li et~al.(2024{\natexlab{c}})Li, Hiersche, and Saygin]{li2024demystifying}
Jin Li, Kelly~J. Hiersche, and Zeynep~M. Saygin.
\newblock Demystifying visual word form area visual and nonvisual response properties with precision {fMRI}.
\newblock \emph{iScience}, 27\penalty0 (12), 2024{\natexlab{c}}.
\newblock URL \url{https://doi.org/10.1016/j.isci.2024.111481}.

\bibitem[Lipkin et~al.(2022)Lipkin, Tuckute, Affourtit, Small, Mineroff, Kean, Jouravlev, Rakocevic, Pritchett, Siegelman, et~al.]{lipkin2022probabilistic}
Benjamin Lipkin, Greta Tuckute, Josef Affourtit, Hannah Small, Zachary Mineroff, Hope Kean, Olessia Jouravlev, Lara Rakocevic, Brianna Pritchett, Matthew Siegelman, et~al.
\newblock Probabilistic atlas for the language network based on precision fmri data from> 800 individuals.
\newblock \emph{Scientific data}, 9\penalty0 (1):\penalty0 529, 2022.

\bibitem[Liu et~al.(2023)Liu, Li, Wu, and Lee]{liu2024visual}
Haotian Liu, Chunyuan Li, Qingyang Wu, and Yong~Jae Lee.
\newblock Visual instruction tuning.
\newblock In \emph{Advances in Neural Information Processing Systems}, volume~36. Curran Associates, Inc., 2023.
\newblock URL \url{https://proceedings.neurips.cc/paper_files/paper/2023/file/6dcf277ea32ce3288914faf369fe6de0-Paper-Conference.pdf}.

\bibitem[Liu et~al.(2024)Liu, Li, Li, and Lee]{liu2024improved}
Haotian Liu, Chunyuan Li, Yuheng Li, and Yong~Jae Lee.
\newblock Improved baselines with visual instruction tuning.
\newblock In \emph{Proceedings of the IEEE/CVF Conference on Computer Vision and Pattern Recognition (CVPR)}, 2024.
\newblock URL \url{https://openaccess.thecvf.com/content/CVPR2024/papers/Liu_Improved_Baselines_with_Visual_Instruction_Tuning_CVPR_2024_paper.pdf}.

\bibitem[Mahowald \& Fedorenko(2016)Mahowald and Fedorenko]{mahowald2016reliable}
Kyle Mahowald and Evelina Fedorenko.
\newblock Reliable individual-level neural markers of high-level language processing: A necessary precursor for relating neural variability to behavioral and genetic variability.
\newblock \emph{Neuroimage}, 139:\penalty0 74--93, 2016.
\newblock URL \url{https://doi.org/10.1016/j.neuroimage.2016.05.073}.

\bibitem[Maniparambil et~al.(2024)Maniparambil, Akshulakov, Djilali, El~Amine~Seddik, Narayan, Mangalam, and O'Connor]{maniparambil2024vision}
Mayug Maniparambil, Raiymbek Akshulakov, Yasser Abdelaziz~Dahou Djilali, Mohamed El~Amine~Seddik, Sanath Narayan, Karttikeya Mangalam, and Noel~E. O'Connor.
\newblock Do vision and language encoders represent the world similarly?
\newblock In \emph{Proceedings of the IEEE/CVF Conference on Computer Vision and Pattern Recognition (CVPR)}, 2024.
\newblock URL \url{https://openaccess.thecvf.com/content/CVPR2024/papers/Maniparambil_Do_Vision_and_Language_Encoders_Represent_the_World_Similarly_CVPR_2024_paper.pdf}.

\bibitem[McKinney(2010)]{pandas}
Wes McKinney.
\newblock {D}ata {S}tructures for {S}tatistical {C}omputing in {P}ython.
\newblock In \emph{{P}roceedings of the 9th {P}ython in {S}cience {C}onference}, pp.\  56--61, 2010.
\newblock URL \url{https://proceedings.scipy.org/articles/Majora-92bf1922-00a}.

\bibitem[Merlin \& Toneva(2024)Merlin and Toneva]{merlin-toneva-2024-language}
Gabriele Merlin and Mariya Toneva.
\newblock Language models and brains align due to more than next-word prediction and word-level information.
\newblock In \emph{Proceedings of the 2024 Conference on Empirical Methods in Natural Language Processing}, pp.\  18431--18454. Association for Computational Linguistics, 2024.
\newblock URL \url{https://aclanthology.org/2024.emnlp-main.1024/}.

\bibitem[Merullo et~al.(2023)Merullo, Castricato, Eickhoff, and Pavlick]{merullo2023linearly}
Jack Merullo, Louis Castricato, Carsten Eickhoff, and Ellie Pavlick.
\newblock Linearly mapping from image to text space.
\newblock In \emph{The Eleventh International Conference on Learning Representations}, 2023.
\newblock URL \url{https://openreview.net/pdf?id=8tYRqb05pVn}.

\bibitem[Oota et~al.(2022{\natexlab{a}})Oota, Arora, Agarwal, Marreddy, Gupta, and Surampudi]{oota-etal-2022-neural}
Subba~Reddy Oota, Jashn Arora, Veeral Agarwal, Mounika Marreddy, Manish Gupta, and Bapi~R. Surampudi.
\newblock Neural language taskonomy: Which {NLP} tasks are the most predictive of f{MRI} brain activity?
\newblock In \emph{Proceedings of the 2022 Conference of the North American Chapter of the Association for Computational Linguistics: Human Language Technologies}, pp.\  3220--3237. Association for Computational Linguistics, 2022{\natexlab{a}}.
\newblock URL \url{https://aclanthology.org/2022.naacl-main.235/}.

\bibitem[Oota et~al.(2022{\natexlab{b}})Oota, Arora, Gupta, and Bapi]{oota-etal-2022-multi}
Subba~Reddy Oota, Jashn Arora, Manish Gupta, and Raju~S. Bapi.
\newblock Multi-view and cross-view brain decoding.
\newblock In \emph{Proceedings of the 29th International Conference on Computational Linguistics}, pp.\  105--115. International Committee on Computational Linguistics, 2022{\natexlab{b}}.
\newblock URL \url{https://aclanthology.org/2022.coling-1.10/}.

\bibitem[Oota et~al.(2022{\natexlab{c}})Oota, Arora, Rowtula, Gupta, and Bapi]{oota-etal-2022-visio}
Subba~Reddy Oota, Jashn Arora, Vijay Rowtula, Manish Gupta, and Raju~S. Bapi.
\newblock Visio-linguistic brain encoding.
\newblock In \emph{Proceedings of the 29th International Conference on Computational Linguistics}, pp.\  116--133. International Committee on Computational Linguistics, 2022{\natexlab{c}}.
\newblock URL \url{https://aclanthology.org/2022.coling-1.11/}.

\bibitem[Oota et~al.(2023)Oota, Gupta, and Toneva]{oota2023joint}
Subba~Reddy Oota, Manish Gupta, and Mariya Toneva.
\newblock Joint processing of linguistic properties in brains and language models.
\newblock In \emph{Advances in Neural Information Processing Systems}, volume~36, pp.\  18001--18014, 2023.
\newblock URL \url{https://proceedings.neurips.cc/paper_files/paper/2023/file/3a0e2de215bd17c39ad08ba1d16c1b12-Paper-Conference.pdf}.

\bibitem[Oota et~al.(2024{\natexlab{a}})Oota, {\c{C}}elik, Deniz, and Toneva]{oota-etal-2024-speech}
Subba~Reddy Oota, Emin {\c{C}}elik, Fatma Deniz, and Mariya Toneva.
\newblock Speech language models lack important brain-relevant semantics.
\newblock In \emph{Proceedings of the 62nd Annual Meeting of the Association for Computational Linguistics (Volume 1: Long Papers)}, pp.\  8503--8528. Association for Computational Linguistics, 2024{\natexlab{a}}.
\newblock URL \url{https://aclanthology.org/2024.acl-long.462/}.

\bibitem[Oota et~al.(2024{\natexlab{b}})Oota, Chen, Gupta, Surampudi, Jobard, Alexandre, and Hinaut]{oota2024deep}
Subba~Reddy Oota, Zijiao Chen, Manish Gupta, Bapi~Raju Surampudi, Ga{\"e}l Jobard, Fr{\'e}d{\'e}ric Alexandre, and Xavier Hinaut.
\newblock Deep neural networks and brain alignment: Brain encoding and decoding (survey).
\newblock \emph{Transactions on Machine Learning Research}, 2024{\natexlab{b}}.
\newblock URL \url{https://openreview.net/pdf?id=YxKJihRcby}.

\bibitem[Paszke et~al.(2019)Paszke, Gross, Massa, Lerer, Bradbury, Chanan, Killeen, Lin, Gimelshein, Antiga, Desmaison, Kopf, Yang, DeVito, Raison, Tejani, Chilamkurthy, Steiner, Fang, Bai, and Chintala]{NEURIPS2019_bdbca288}
Adam Paszke, Sam Gross, Francisco Massa, Adam Lerer, James Bradbury, Gregory Chanan, Trevor Killeen, Zeming Lin, Natalia Gimelshein, Luca Antiga, Alban Desmaison, Andreas Kopf, Edward Yang, Zachary DeVito, Martin Raison, Alykhan Tejani, Sasank Chilamkurthy, Benoit Steiner, Lu~Fang, Junjie Bai, and Soumith Chintala.
\newblock {PyTorch}: An imperative style, high-performance deep learning library.
\newblock In \emph{Advances in Neural Information Processing Systems}, volume~32. Curran Associates, Inc., 2019.
\newblock URL \url{https://proceedings.neurips.cc/paper_files/paper/2019/file/bdbca288fee7f92f2bfa9f7012727740-Paper.pdf}.

\bibitem[Patterson \& Ralph(2016)Patterson and Ralph]{patterson2016hub}
Karalyn Patterson and Matthew A.~Lambon Ralph.
\newblock The hub-and-spoke hypothesis of semantic memory.
\newblock In \emph{Neurobiology of Language}, pp.\  765--775. Elsevier, 2016.
\newblock URL \url{https://www.sciencedirect.com/science/article/abs/pii/B9780124077942000614}.

\bibitem[Pedregosa et~al.(2011)Pedregosa, Varoquaux, Gramfort, Michel, Thirion, Grisel, Blondel, Prettenhofer, Weiss, Dubourg, Vanderplas, Passos, Cournapeau, Brucher, Perrot, and {{\'E}}douard Duchesnay]{sklearn}
Fabian Pedregosa, Ga{{\"e}}l Varoquaux, Alexandre Gramfort, Vincent Michel, Bertrand Thirion, Olivier Grisel, Mathieu Blondel, Peter Prettenhofer, Ron Weiss, Vincent Dubourg, Jake Vanderplas, Alexandre Passos, David Cournapeau, Matthieu Brucher, Matthieu Perrot, and {{\'E}}douard Duchesnay.
\newblock {Scikit-learn: Machine Learning in Python}.
\newblock \emph{{J}ournal of {M}achine {L}earning {R}esearch}, 12\penalty0 (85):\penalty0 2825--2830, 2011.
\newblock URL \url{http://jmlr.org/papers/v12/pedregosa11a.html}.

\bibitem[Pennington et~al.(2014)Pennington, Socher, and Manning]{pennington-etal-2014-glove}
Jeffrey Pennington, Richard Socher, and Christopher Manning.
\newblock {G}lo{V}e: Global vectors for word representation.
\newblock In \emph{Proceedings of the 2014 Conference on Empirical Methods in Natural Language Processing ({EMNLP})}, pp.\  1532--1543. Association for Computational Linguistics, 2014.
\newblock URL \url{https://aclanthology.org/D14-1162/}.

\bibitem[Pereira et~al.(2018)Pereira, Lou, Pritchett, Ritter, Gershman, Kanwisher, Botvinick, and Fedorenko]{pereira2018toward}
Francisco Pereira, Bin Lou, Brianna Pritchett, Samuel Ritter, Samuel~J. Gershman, Nancy Kanwisher, Matthew Botvinick, and Evelina Fedorenko.
\newblock Toward a universal decoder of linguistic meaning from brain activation.
\newblock \emph{Nature Communications}, 9\penalty0 (1):\penalty0 963, 2018.
\newblock URL \url{https://www.nature.com/articles/s41467-018-03068-4}.

\bibitem[Piantadosi \& Hill(2022)Piantadosi and Hill]{piantadosi2022meaning}
Steven~T. Piantadosi and Felix Hill.
\newblock Meaning without reference in large language models.
\newblock In \emph{NeurIPS 2022 Workshop on Neuro Causal and Symbolic AI (nCSI)}, 2022.
\newblock URL \url{https://arxiv.org/abs/2208.02957}.

\bibitem[Popham et~al.(2021)Popham, Huth, Bilenko, Deniz, Gao, Nunez-Elizalde, and Gallant]{popham2021visual}
Sara~F. Popham, Alexander~G. Huth, Natalia~Y. Bilenko, Fatma Deniz, James~S. Gao, Anwar~O. Nunez-Elizalde, and Jack~L. Gallant.
\newblock Visual and linguistic semantic representations are aligned at the border of human visual cortex.
\newblock \emph{Nature Neuroscience}, 24\penalty0 (11):\penalty0 1628--1636, 2021.
\newblock URL \url{https://www.nature.com/articles/s41593-021-00921-6}.

\bibitem[Prince et~al.(2022)Prince, Charest, Kurzawski, Pyles, Tarr, and Kay]{glmsingle}
Jacob~S Prince, Ian Charest, Jan~W Kurzawski, John~A Pyles, Michael~J Tarr, and Kendrick~N Kay.
\newblock Improving the accuracy of single-trial {fMRI} response estimates using {GLMsingle}.
\newblock \emph{eLife}, 11:\penalty0 e77599, 2022.
\newblock URL \url{https://elifesciences.org/articles/77599}.

\bibitem[Radford et~al.(2018)Radford, Narasimhan, Salimans, and Sutskever]{gpt}
Alec Radford, Karthik Narasimhan, Tim Salimans, and Ilya Sutskever.
\newblock Improving language understanding by generative pre-training.
\newblock Technical report, OpenAI, 2018.
\newblock URL \url{https://cdn.openai.com/research-covers/language-unsupervised/language_understanding_paper.pdf}.

\bibitem[Radford et~al.(2019)Radford, Wu, Child, Luan, Amodei, and Sutskever]{radford2019language}
Alec Radford, Jeff Wu, Rewon Child, David Luan, Dario Amodei, and Ilya Sutskever.
\newblock Language models are unsupervised multitask learners.
\newblock Technical report, OpenAI, 2019.
\newblock URL \url{https://cdn.openai.com/better-language-models/language_models_are_unsupervised_multitask_learners.pdf}.

\bibitem[Radford et~al.(2021)Radford, Kim, Hallacy, Ramesh, Goh, Agarwal, Sastry, Askell, Mishkin, Clark, Krueger, and Sutskever]{pmlr-v139-radford21a}
Alec Radford, Jong~Wook Kim, Chris Hallacy, Aditya Ramesh, Gabriel Goh, Sandhini Agarwal, Girish Sastry, Amanda Askell, Pamela Mishkin, Jack Clark, Gretchen Krueger, and Ilya Sutskever.
\newblock Learning transferable visual models from natural language supervision.
\newblock In \emph{Proceedings of the 38th International Conference on Machine Learning}, volume 139 of \emph{Proceedings of Machine Learning Research}. PMLR, 2021.
\newblock URL \url{https://proceedings.mlr.press/v139/radford21a.html}.

\bibitem[Reilly et~al.(2025)Reilly, Shain, Borghesani, Kuhnke, Vigliocco, Peelle, Mahon, Buxbaum, Majid, Brysbaert, Borghi, De~Deyne, Dove, Papeo, Pexman, Poeppel, Lupyan, Boggio, Hickok, Gwilliams, Fernandino, Mirman, Chrysikou, Sandberg, Crutch, Pylkkänen, Yee, Jackson, Rodd, Bedny, Connell, Kiefer, Kemmerer, de~Zubicaray, Jefferies, Lynott, Siew, Desai, McRae, Diaz, Bolognesi, Fedorenko, Kiran, Montefinese, Binder, Yap, Hartwigsen, Cantlon, Bi, Hoffman, Garcea, and Vinson]{reilly2025we}
Jamie Reilly, Cory Shain, Valentina Borghesani, Philipp Kuhnke, Gabriella Vigliocco, Jonathan~E. Peelle, Bradford~Z. Mahon, Laurel~J. Buxbaum, Asifa Majid, Marc Brysbaert, Anna~M. Borghi, Simon De~Deyne, Guy Dove, Liuba Papeo, Penny~M. Pexman, David Poeppel, Gary Lupyan, Paulo Boggio, Gregory Hickok, Laura Gwilliams, Leonardo Fernandino, Daniel Mirman, Evangelia~G. Chrysikou, Chaleece~W. Sandberg, Sebastian~J. Crutch, Liina Pylkkänen, Eiling Yee, Rebecca~L. Jackson, Jennifer~M. Rodd, Marina Bedny, Louise Connell, Markus Kiefer, David Kemmerer, Greig de~Zubicaray, Elizabeth Jefferies, Dermot Lynott, Cynthia S.~Q. Siew, Rutvik~H. Desai, Ken McRae, Michele~T. Diaz, Marianna Bolognesi, Evelina Fedorenko, Swathi Kiran, Maria Montefinese, Jeffrey~R. Binder, Melvin~J. Yap, Gesa Hartwigsen, Jessica Cantlon, Yanchao Bi, Paul Hoffman, Frank~E Garcea, and David Vinson.
\newblock What we mean when we say semantic: Toward a multidisciplinary semantic glossary.
\newblock \emph{Psychonomic Bulletin \& Review}, 32\penalty0 (1):\penalty0 243--280, 2025.
\newblock URL \url{https://link.springer.com/article/10.3758/s13423-024-02556-7}.

\bibitem[Rolls(2023)]{10.1093/oso/9780198887911.003.0002}
Edmund~T. Rolls.
\newblock The ventral visual system.
\newblock In \emph{Brain Computations and Connectivity}. Oxford University Press, July 2023.
\newblock ISBN 9780198887911.
\newblock URL \url{https://doi.org/10.1093/oso/9780198887911.003.0002}.

\bibitem[Schrimpf et~al.(2021)Schrimpf, Blank, Tuckute, Kauf, Hosseini, Kanwisher, Tenenbaum, and Fedorenko]{schrimpf2021neural}
Martin Schrimpf, Idan~Asher Blank, Greta Tuckute, Carina Kauf, Eghbal~A. Hosseini, Nancy Kanwisher, Joshua~B. Tenenbaum, and Evelina Fedorenko.
\newblock The neural architecture of language: {I}ntegrative modeling converges on predictive processing.
\newblock \emph{Proceedings of the National Academy of Sciences}, 118\penalty0 (45):\penalty0 e2105646118, 2021.
\newblock URL \url{https://www.pnas.org/doi/full/10.1073/pnas.2105646118}.

\bibitem[Singh et~al.(2022)Singh, Hu, Goswami, Couairon, Galuba, Rohrbach, and Kiela]{singh2022flava}
Amanpreet Singh, Ronghang Hu, Vedanuj Goswami, Guillaume Couairon, Wojciech Galuba, Marcus Rohrbach, and Douwe Kiela.
\newblock {FLAVA}: A foundational language and vision alignment model.
\newblock In \emph{Proceedings of the IEEE/CVF Conference on Computer Vision and Pattern Recognition (CVPR)}, 2022.
\newblock URL \url{https://openaccess.thecvf.com/content/CVPR2022/papers/Singh_FLAVA_A_Foundational_Language_and_Vision_Alignment_Model_CVPR_2022_paper.pdf}.

\bibitem[Sucholutsky et~al.(2024)Sucholutsky, Muttenthaler, Weller, Peng, Bobu, Kim, Love, Cueva, Grant, Groen, Achterberg, Tenenbaum, Collins, Hermann, Oktar, Greff, Hebart, Cloos, Kriegeskorte, Jacoby, Zhang, Marjieh, Geirhos, Chen, Kornblith, Rane, Konkle, O'Connell, Unterthiner, Lampinen, Müller, Toneva, and Griffiths]{sucholutsky2024gettingalignedrepresentationalalignment}
Ilia Sucholutsky, Lukas Muttenthaler, Adrian Weller, Andi Peng, Andreea Bobu, Been Kim, Bradley~C. Love, Christopher~J. Cueva, Erin Grant, Iris Groen, Jascha Achterberg, Joshua~B. Tenenbaum, Katherine~M. Collins, Katherine~L. Hermann, Kerem Oktar, Klaus Greff, Martin~N. Hebart, Nathan Cloos, Nikolaus Kriegeskorte, Nori Jacoby, Qiuyi Zhang, Raja Marjieh, Robert Geirhos, Sherol Chen, Simon Kornblith, Sunayana Rane, Talia Konkle, Thomas~P. O'Connell, Thomas Unterthiner, Andrew~K. Lampinen, Klaus-Robert Müller, Mariya Toneva, and Thomas~L. Griffiths.
\newblock Getting aligned on representational alignment.
\newblock \emph{arXiv preprint arXiv 2310.13018}, 2024.
\newblock URL \url{https://arxiv.org/abs/2310.13018}.

\bibitem[Toneva \& Wehbe(2019)Toneva and Wehbe]{toneva2019interpreting}
Mariya Toneva and Leila Wehbe.
\newblock Interpreting and improving natural-language processing (in machines) with natural language-processing (in the brain).
\newblock In \emph{Advances in Neural Information Processing Systems}, volume~32. Curran Associates, Inc., 2019.
\newblock URL \url{https://papers.nips.cc/paper_files/paper/2019/hash/749a8e6c231831ef7756db230b4359c8-Abstract.html}.

\bibitem[Touvron et~al.(2023)Touvron, Martin, Stone, Albert, Almahairi, Babaei, Bashlykov, Batra, Bhargava, Bhosale, Bikel, Blecher, Ferrer, Chen, Cucurull, Esiobu, Fernandes, Fu, Fu, Fuller, Gao, Goswami, Goyal, Hartshorn, Hosseini, Hou, Inan, Kardas, Kerkez, Khabsa, Kloumann, Korenev, Koura, Lachaux, Lavril, Lee, Liskovich, Lu, Mao, Martinet, Mihaylov, Mishra, Molybog, Nie, Poulton, Reizenstein, Rungta, Saladi, Schelten, Silva, Smith, Subramanian, Tan, Tang, Taylor, Williams, Kuan, Xu, Yan, Zarov, Zhang, Fan, Kambadur, Narang, Rodriguez, Stojnic, Edunov, and Scialom]{touvron2023llama2openfoundation}
Hugo Touvron, Louis Martin, Kevin Stone, Peter Albert, Amjad Almahairi, Yasmine Babaei, Nikolay Bashlykov, Soumya Batra, Prajjwal Bhargava, Shruti Bhosale, Dan Bikel, Lukas Blecher, Cristian~Canton Ferrer, Moya Chen, Guillem Cucurull, David Esiobu, Jude Fernandes, Jeremy Fu, Wenyin Fu, Brian Fuller, Cynthia Gao, Vedanuj Goswami, Naman Goyal, Anthony Hartshorn, Saghar Hosseini, Rui Hou, Hakan Inan, Marcin Kardas, Viktor Kerkez, Madian Khabsa, Isabel Kloumann, Artem Korenev, Punit~Singh Koura, Marie-Anne Lachaux, Thibaut Lavril, Jenya Lee, Diana Liskovich, Yinghai Lu, Yuning Mao, Xavier Martinet, Todor Mihaylov, Pushkar Mishra, Igor Molybog, Yixin Nie, Andrew Poulton, Jeremy Reizenstein, Rashi Rungta, Kalyan Saladi, Alan Schelten, Ruan Silva, Eric~Michael Smith, Ranjan Subramanian, Xiaoqing~Ellen Tan, Binh Tang, Ross Taylor, Adina Williams, Jian~Xiang Kuan, Puxin Xu, Zheng Yan, Iliyan Zarov, Yuchen Zhang, Angela Fan, Melanie Kambadur, Sharan Narang, Aurelien Rodriguez, Robert Stojnic, Sergey Edunov, and Thomas
  Scialom.
\newblock Llama 2: Open foundation and fine-tuned chat models.
\newblock \emph{arXiv preprint arXiv:2307.09288}, 2023.
\newblock URL \url{https://arxiv.org/abs/2307.09288}.

\bibitem[Tuckute et~al.(2024{\natexlab{a}})Tuckute, Kanwisher, and Fedorenko]{tuckute2024language}
Greta Tuckute, Nancy Kanwisher, and Evelina Fedorenko.
\newblock Language in brains, minds, and machines.
\newblock \emph{Annual Review of Neuroscience}, 47, 2024{\natexlab{a}}.
\newblock URL \url{https://www.annualreviews.org/content/journals/10.1146/annurev-neuro-120623-101142?TRACK=RSS}.

\bibitem[Tuckute et~al.(2024{\natexlab{b}})Tuckute, Sathe, Srikant, Taliaferro, Wang, Schrimpf, Kay, and Fedorenko]{tuckute2024driving}
Greta Tuckute, Aalok Sathe, Shashank Srikant, Maya Taliaferro, Mingye Wang, Martin Schrimpf, Kendrick Kay, and Evelina Fedorenko.
\newblock Driving and suppressing the human language network using large language models.
\newblock \emph{Nature Human Behaviour}, 8\penalty0 (3):\penalty0 544--561, 2024{\natexlab{b}}.
\newblock URL \url{https://www.nature.com/articles/s41562-023-01783-7}.

\bibitem[Vaswani et~al.(2017)Vaswani, Shazeer, Parmar, Uszkoreit, Jones, Gomez, Kaiser, and Polosukhin]{Vaswani+2017}
Ashish Vaswani, Noam Shazeer, Niki Parmar, Jakob Uszkoreit, Llion Jones, Aidan~N. Gomez, {\L}ukasz Kaiser, and Illia Polosukhin.
\newblock Attention is all you need.
\newblock In \emph{Advances in Neural Information Processing Systems}, volume~30. Curran Associates, Inc., 2017.
\newblock URL \url{https://proceedings.neurips.cc/paper_files/paper/2017/file/3f5ee243547dee91fbd053c1c4a845aa-Paper.pdf}.

\bibitem[{Virtanen} et~al.(2020){Virtanen}, {Gommers}, {Oliphant}, {Haberland}, {Reddy}, {Cournapeau}, {Burovski}, {Peterson}, {Weckesser}, {Bright}, {van der Walt}, {Brett}, {Wilson}, {Jarrod Millman}, {Mayorov}, {Nelson}, {Jones}, {Kern}, {Larson}, {Carey}, {Polat}, {Feng}, {Moore}, {Vand erPlas}, {Laxalde}, {Perktold}, {Cimrman}, {Henriksen}, {Quintero}, {Harris}, {Archibald}, {Ribeiro}, {Pedregosa}, {van Mulbregt}, and {Contributors}]{scipy}
Pauli {Virtanen}, Ralf {Gommers}, Travis~E. {Oliphant}, Matt {Haberland}, Tyler {Reddy}, David {Cournapeau}, Evgeni {Burovski}, Pearu {Peterson}, Warren {Weckesser}, Jonathan {Bright}, St{\'e}fan~J. {van der Walt}, Matthew {Brett}, Joshua {Wilson}, K.~{Jarrod Millman}, Nikolay {Mayorov}, Andrew R.~J. {Nelson}, Eric {Jones}, Robert {Kern}, Eric {Larson}, CJ~{Carey}, {\.I}lhan {Polat}, Yu~{Feng}, Eric~W. {Moore}, Jake {Vand erPlas}, Denis {Laxalde}, Josef {Perktold}, Robert {Cimrman}, Ian {Henriksen}, E.~A. {Quintero}, Charles~R {Harris}, Anne~M. {Archibald}, Ant{\^o}nio~H. {Ribeiro}, Fabian {Pedregosa}, Paul {van Mulbregt}, and SciPy 1.~0 {Contributors}.
\newblock {SciPy 1.0: Fundamental Algorithms for Scientific Computing in Python}.
\newblock \emph{Nature Methods}, 17:\penalty0 261--272, 2020.
\newblock URL \url{https://doi.org/10.1038/s41592-019-0686-2}.

\bibitem[Wang et~al.(2024)Wang, Bai, Tan, Wang, Fan, Bai, Chen, Liu, Wang, Ge, Fan, Dang, Du, Ren, Men, Liu, Zhou, Zhou, and Lin]{Qwen2-VL}
Peng Wang, Shuai Bai, Sinan Tan, Shijie Wang, Zhihao Fan, Jinze Bai, Keqin Chen, Xuejing Liu, Jialin Wang, Wenbin Ge, Yang Fan, Kai Dang, Mengfei Du, Xuancheng Ren, Rui Men, Dayiheng Liu, Chang Zhou, Jingren Zhou, and Junyang Lin.
\newblock Qwen2-{VL}: Enhancing vision-language model's perception of the world at any resolution.
\newblock \emph{arXiv preprint arXiv:2409.12191}, 2024.
\newblock URL \url{https://arxiv.org/abs/2409.12191}.

\bibitem[Wolf et~al.(2020)Wolf, Debut, Sanh, Chaumond, Delangue, Moi, Cistac, Rault, Louf, Funtowicz, Davison, Shleifer, von Platen, Ma, Jernite, Plu, Xu, Scao, Gugger, Drame, Lhoest, and Rush]{wolf-etal-2020-transformers}
Thomas Wolf, Lysandre Debut, Victor Sanh, Julien Chaumond, Clement Delangue, Anthony Moi, Pierric Cistac, Tim Rault, Rémi Louf, Morgan Funtowicz, Joe Davison, Sam Shleifer, Patrick von Platen, Clara Ma, Yacine Jernite, Julien Plu, Canwen Xu, Teven~Le Scao, Sylvain Gugger, Mariama Drame, Quentin Lhoest, and Alexander~M. Rush.
\newblock Transformers: State-of-the-art natural language processing.
\newblock In \emph{Proceedings of the 2020 Conference on Empirical Methods in Natural Language Processing: System Demonstrations}, pp.\  38--45. Association for Computational Linguistics, 2020.
\newblock URL \url{https://www.aclweb.org/anthology/2020.emnlp-demos.6}.

\bibitem[Wu et~al.(2025)Wu, Yu, Yogatama, Lu, and Kim]{wu2024semantic}
Zhaofeng Wu, Xinyan~Velocity Yu, Dani Yogatama, Jiasen Lu, and Yoon Kim.
\newblock The semantic hub hypothesis: Language models share semantic representations across languages and modalities.
\newblock In \emph{International Conference on Learning Representations}, 2025.
\newblock URL \url{https://openreview.net/pdf?id=FrFQpAgnGE}.

\bibitem[Wurm \& Caramazza(2019)Wurm and Caramazza]{wurm2019distinct}
Moritz~F. Wurm and Alfonso Caramazza.
\newblock Distinct roles of temporal and frontoparietal cortex in representing actions across vision and language.
\newblock \emph{Nature Communications}, 10\penalty0 (1):\penalty0 289, 2019.
\newblock URL \url{https://www.nature.com/articles/s41467-018-08084-y}.

\bibitem[Yamins et~al.(2014)Yamins, Hong, Cadieu, Solomon, Seibert, and DiCarlo]{yamins2014performance}
Daniel L.~K. Yamins, Ha~Hong, Charles~F. Cadieu, Ethan~A. Solomon, Darren Seibert, and James~J. DiCarlo.
\newblock Performance-optimized hierarchical models predict neural responses in higher visual cortex.
\newblock \emph{Proceedings of the National Academy of Sciences}, 111\penalty0 (23):\penalty0 8619--8624, 2014.
\newblock URL \url{https://www.pnas.org/doi/abs/10.1073/pnas.1403112111}.

\bibitem[Yang et~al.(2024{\natexlab{a}})Yang, Yang, Hui, Zheng, Yu, Zhou, Li, Li, Liu, Huang, Dong, Wei, Lin, Tang, Wang, Yang, Tu, Zhang, Ma, Xu, Zhou, Bai, He, Lin, Dang, Lu, Chen, Yang, Li, Xue, Ni, Zhang, Wang, Peng, Men, Gao, Lin, Wang, Bai, Tan, Zhu, Li, Liu, Ge, Deng, Zhou, Ren, Zhang, Wei, Ren, Fan, Yao, Zhang, Wan, Chu, Liu, Cui, Zhang, and Fan]{qwen2}
An~Yang, Baosong Yang, Binyuan Hui, Bo~Zheng, Bowen Yu, Chang Zhou, Chengpeng Li, Chengyuan Li, Dayiheng Liu, Fei Huang, Guanting Dong, Haoran Wei, Huan Lin, Jialong Tang, Jialin Wang, Jian Yang, Jianhong Tu, Jianwei Zhang, Jianxin Ma, Jin Xu, Jingren Zhou, Jinze Bai, Jinzheng He, Junyang Lin, Kai Dang, Keming Lu, Keqin Chen, Kexin Yang, Mei Li, Mingfeng Xue, Na~Ni, Pei Zhang, Peng Wang, Ru~Peng, Rui Men, Ruize Gao, Runji Lin, Shijie Wang, Shuai Bai, Sinan Tan, Tianhang Zhu, Tianhao Li, Tianyu Liu, Wenbin Ge, Xiaodong Deng, Xiaohuan Zhou, Xingzhang Ren, Xinyu Zhang, Xipin Wei, Xuancheng Ren, Yang Fan, Yang Yao, Yichang Zhang, Yu~Wan, Yunfei Chu, Yuqiong Liu, Zeyu Cui, Zhenru Zhang, and Zhihao Fan.
\newblock Qwen2 technical report.
\newblock \emph{arXiv preprint arXiv:2407.10671}, 2024{\natexlab{a}}.
\newblock URL \url{https://arxiv.org/abs/2407.10671}.

\bibitem[Yang et~al.(2024{\natexlab{b}})Yang, Yang, Zhang, Hui, Zheng, Yu, Li, Liu, Huang, Wei, Lin, Yang, Tu, Zhang, Yang, Yang, Zhou, Lin, Dang, Lu, Bao, Yang, Yu, Li, Xue, Zhang, Zhu, Men, Lin, Li, Xia, Ren, Ren, Fan, Su, Zhang, Wan, Liu, Cui, Zhang, and Qiu]{qwen2.5}
An~Yang, Baosong Yang, Beichen Zhang, Binyuan Hui, Bo~Zheng, Bowen Yu, Chengyuan Li, Dayiheng Liu, Fei Huang, Haoran Wei, Huan Lin, Jian Yang, Jianhong Tu, Jianwei Zhang, Jianxin Yang, Jiaxi Yang, Jingren Zhou, Junyang Lin, Kai Dang, Keming Lu, Keqin Bao, Kexin Yang, Le~Yu, Mei Li, Mingfeng Xue, Pei Zhang, Qin Zhu, Rui Men, Runji Lin, Tianhao Li, Tingyu Xia, Xingzhang Ren, Xuancheng Ren, Yang Fan, Yang Su, Yichang Zhang, Yu~Wan, Yuqiong Liu, Zeyu Cui, Zhenru Zhang, and Zihan Qiu.
\newblock Qwen2.5 technical report.
\newblock \emph{arXiv preprint arXiv:2412.15115}, 2024{\natexlab{b}}.
\newblock URL \url{https://arxiv.org/abs/2412.15115}.

\bibitem[Yu et~al.(2024)Yu, Gu, Huang, and Li]{yu2024predicting}
Shaoyun Yu, Chanyuan Gu, Kexin Huang, and Ping Li.
\newblock Predicting the next sentence (not word) in large language models: What model-brain alignment tells us about discourse comprehension.
\newblock \emph{Science advances}, 10\penalty0 (21):\penalty0 eadn7744, 2024.
\newblock URL \url{https://doi.org/10.1126/sciadv.adn7744}.

\bibitem[Zheng et~al.(2024)Zheng, Chiang, Sheng, Zhuang, Wu, Zhuang, Lin, Li, Li, Xing, Zhang, Gonzalez, and Stoica]{zheng2024judging}
Lianmin Zheng, Wei-Lin Chiang, Ying Sheng, Siyuan Zhuang, Zhanghao Wu, Yonghao Zhuang, Zi~Lin, Zhuohan Li, Dacheng Li, Eric~P. Xing, Hao Zhang, Joseph~E. Gonzalez, and Ion Stoica.
\newblock Judging {LLM}-as-a-judge with {MT}-bench and {C}hatbot {A}rena.
\newblock In \emph{Advances in Neural Information Processing Systems}, volume~36. Curran Associates, Inc., 2024.
\newblock URL \url{https://proceedings.neurips.cc/paper_files/paper/2023/file/91f18a1287b398d378ef22505bf41832-Paper-Datasets_and_Benchmarks.pdf}.

\end{thebibliography}
\bibliographystyle{colm2025_conference}

\appendix
\section{fMRI data}
\label{sec:appendix-data}

This section contains additional details on the fMRI dataset used in this study \citep[Experiment 1]{pereira2018toward}, summarized from the original publication, as well as a description of our alternative processing choices. The stimuli and the fMRI data processed by \citeauthor{pereira2018toward}'s original processing pipeline are published or linked at \url{https://osf.io/crwz7/}.

\begin{table}[htb]
\begin{center}
\begin{tabular}{ccccc}
\toprule
\texttt{Ability} & \texttt{Cook} & \texttt{Food} & \texttt{Music} & \texttt{Sin}\\
\texttt{Accomplished} & \texttt{Counting} & \texttt{Garbage} & \texttt{Nation} & \texttt{Skin} \\ 
\texttt{Angry} & \texttt{Crazy} & \texttt{Gold} & \texttt{News} & \texttt{Smart}\\
\texttt{Apartment} & \texttt{Damage} & \texttt{Great} & \texttt{Noise} & \texttt{Smiling}\\
\texttt{Applause} & \texttt{Dance} & \texttt{Gun} & \texttt{Obligation} & \texttt{Solution}\\
\texttt{Argument} & \texttt{Dangerous} & \texttt{Hair} & \texttt{Pain} & \texttt{Soul} \\
\texttt{Argumentatively} & \texttt{Deceive} & \texttt{Help} & \texttt{Personality} & \texttt{Sound}\\
\texttt{Art} & \texttt{Dedication} & \texttt{Hurting} & \texttt{Philosophy} & \texttt{Spoke}\\
\texttt{Attitude} & \texttt{Deliberately} & \texttt{Ignorance} & \texttt{Picture} & \texttt{Star} \\
\texttt{Bag} & \texttt{Delivery} & \texttt{Illness} & \texttt{Pig} & \texttt{Student}\\
\texttt{Ball} & \texttt{Dessert} & \texttt{Impress} & \texttt{Plan} & \texttt{Stupid}\\
\texttt{Bar} & \texttt{Device} & \texttt{Invention} & \texttt{Plant} & \texttt{Successful} \\
\texttt{Bear} & \texttt{Dig} & \texttt{Investigation} & \texttt{Play} & \texttt{Sugar}\\
\texttt{Beat} & \texttt{Dinner} & \texttt{Invisible} & \texttt{Pleasure} & \texttt{Suspect}\\
\texttt{Bed} & \texttt{Disease} & \texttt{Job} & \texttt{Poor} & \texttt{Table}\\
\texttt{Beer} & \texttt{Dissolve} & \texttt{Jungle} & \texttt{Prison} & \texttt{Taste}\\
\texttt{Big} & \texttt{Disturb} & \texttt{Kindness} & \texttt{Professional} & \texttt{Team} \\
\texttt{Bird} & \texttt{Do} & \texttt{King} & \texttt{Protection} & \texttt{Texture}\\
\texttt{Blood} & \texttt{Doctor} & \texttt{Lady} & \texttt{Quality} & \texttt{Time} \\
\texttt{Body} & \texttt{Dog} & \texttt{Land} & \texttt{Reaction} & \texttt{Tool} \\
\texttt{Brain} & \texttt{Dressing} & \texttt{Laugh} & \texttt{Read} & \texttt{Toy}\\
\texttt{Broken} &  \texttt{Driver} & \texttt{Law} & \texttt{Relationship} & \texttt{Tree}\\
\texttt{Building} & \texttt{Economy} & \texttt{Left} & \texttt{Religious} & \texttt{Trial}\\
\texttt{Burn} & \texttt{Election} & \texttt{Level} & \texttt{Residence} & \texttt{Tried} \\
\texttt{Business} & \texttt{Electron} & \texttt{Liar} & \texttt{Road} & \texttt{Typical}\\
\texttt{Camera} & \texttt{Elegance} & \texttt{Light} & \texttt{Sad} & \texttt{Unaware}\\
\texttt{Carefully} & \texttt{Emotion} & \texttt{Magic} & \texttt{Science} & \texttt{Usable} \\ 
\texttt{Challenge} & \texttt{Emotionally} & \texttt{Marriage} & \texttt{Seafood} & \texttt{Useless} \\
\texttt{Charity} & \texttt{Engine} & \texttt{Material} & \texttt{Sell} & \texttt{Vacation}\\
\texttt{Charming} &\texttt{Event}  & \texttt{Mathematical} & \texttt{Sew} & \texttt{War}\\
\texttt{Clothes} & \texttt{Experiment} & \texttt{Mechanism} & \texttt{Sexy} & \texttt{Wash}\\
\texttt{Cockroach} & \texttt{Extremely} & \texttt{Medication} & \texttt{Shape} & \texttt{Weak}\\
\texttt{Code} & \texttt{Feeling} & \texttt{Money} & \texttt{Ship} & \texttt{Wear}\\
\texttt{Collection} & \texttt{Fight} & \texttt{Mountain} & \texttt{Show} & \texttt{Weather}\\
\texttt{Computer} & \texttt{Fish} & \texttt{Movement} & \texttt{Sign} & \texttt{Willingly}\\
\texttt{Construction} & \texttt{Flow} & \texttt{Movie} & \texttt{Silly} & \texttt{Word}\\
\bottomrule
\end{tabular}
\caption{{\bf The 180 concept words of \cite{pereira2018toward}}. Each word is a label of a cluster of words obtained by performing spectral clustering over a space of GloVe embeddings.
\label{tab:concepts}}
\end{center}
\end{table}

\subsection{Concepts} 

The full list of 180 concepts used in the study is provided in \Tref{tab:concepts}. \cite{pereira2018toward} perform spectral clustering over a pre-trained English GloVe embedding space \citep{pennington-etal-2014-glove} and manually label the obtained clusters. The list includes 128 nouns, 22 verbs, 29 adjectives and adverbs, and 1 function word.

\subsection{Stimuli}

\begin{figure}[htb]
    \centering
    \includegraphics[width=\linewidth]{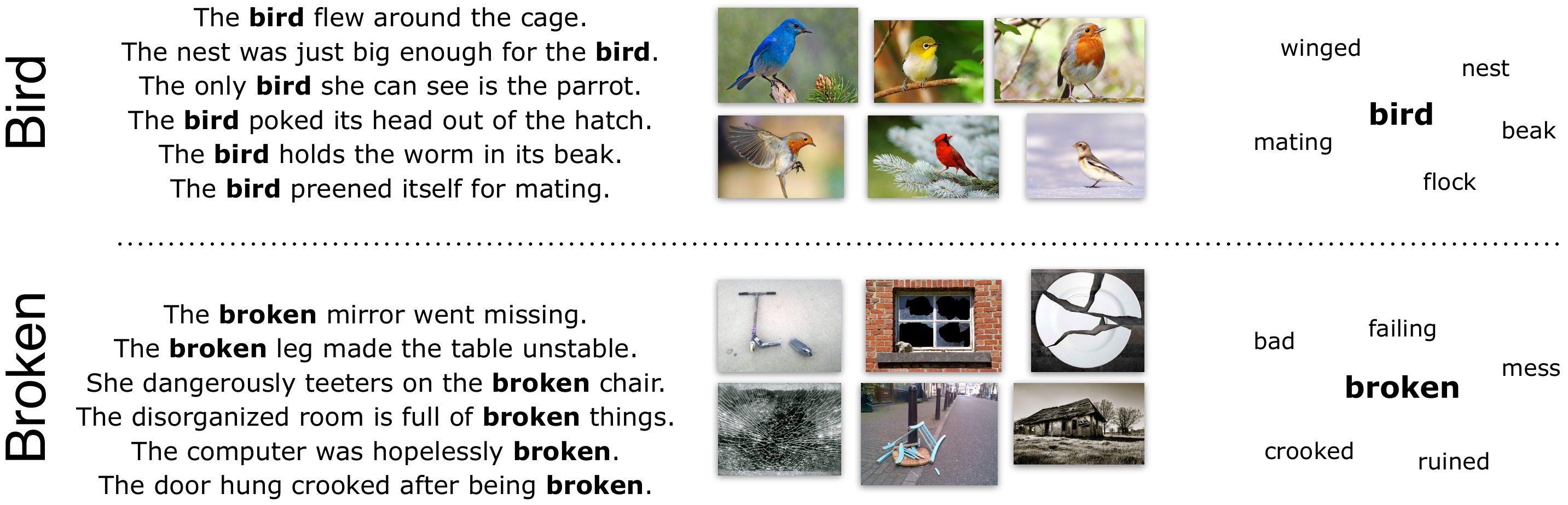}
    \caption{\textbf{Example stimuli for two of the concepts in the \cite{pereira2018toward} Experiment 1 dataset.} The dataset includes six sentences, six images, and six spatial configurations of the same word cloud. The concept word is always bolded in the sentences and word clouds, and it is also added to every image in the picture paradigm. The participants were asked to think about the concept in relation to the accompanying context or image.}
    \label{fig:stimuli}
\end{figure}

\Figref{fig:stimuli} shows example stimuli for two of the 180 concepts under each paradigm.

\subsection{Participants} 

It should be noted that the set of participants considered in this study (M01--M17) is not identical to that of \cite{pereira2018toward}: we exclude one participant scanned at Princeton (P01) but include the two novice subjects (M11, M12) that were excluded from the original analyses. The 17 participants (mean age 26.1, range 20–48; 10 men, 7 women; all native speakers of English; 14 right-handed, two left-handed, one ambidextrous) received payment for their participation, and gave informed consent in accordance with the requirements of the Committee on the Use of Humans as Experimental Subjects. 

\subsection{fMRI protocol} 

fMRI scanning was performed using a whole-body 3-Tesla Siemens Trio scanner with a 32-channel head coil. 
Each two-hour scan session included 4--6 groups of 180 stimuli (one per concept), with each group randomly split into two runs (90+90 concepts). Each stimulus was presented for 3 seconds followed by a 2-second fixation period, with additional 10-second fixation periods at the beginning, middle, and end of each run. The scan repetition time (TR) was set to 2 seconds.

\subsection{fMRI data processing} 
The responses to each stimulus were estimated using a general linear model (GLM) with additional denoising and regularization, implemented using the GLMsingle \citep{glmsingle} Python toolkit (version 0.0.1).\footnote{\url{https://github.com/cvnlab/GLMsingle}} Each stimulus presentation was modeled with a boxcar function convolved with the canonical haemodynamic response (HRF). The time-series data is upsampled using PCHIP interpolation to TR=1s (from TR=2s in the original data from \citeauthor{pereira2018toward}) in order to align the duration of the stimulus presentations (3s) with the TR boundaries. We set the following GLMsingle hyperparameters: number of GLMdenoise regressors = 5; fractional regularization level = 0.05; default values for the rest.

\section{Semantic consistency ROIs}
\label{sec:appendix-b}

\subsection{Statistically significant voxels}
\label{sec:appendix-prob-map}

We determine the statistical significance of a voxel's semantic consistency using independent permutation tests performed on two non-overlapping halves of the data.

First, we partition the stimuli set into two halves: for example, for a given concept (e.g., \texttt{Ability}) and paradigm (e.g., sentences), sentences 1, 2, and 5 are allocated to the first half and sentences 3, 4, and 6 to the second half. Since for certain concept--paradigm pairs there are occasional participants who have only been presented 4 stimuli out of the possible 6, we partition the stimuli in a way that would result in the most even data split between the two halves: specifically, we choose a split that minimizes the number of cases where a subject would have seen three stimuli from one half but only one from the other half.

We then perform a permutation test on the brain activations for each half of the stimuli. For each paradigm $\Omega \in \{\text{\s, \p, \wc} \}$ we consider a voxel's response vector $\Vec{\beta}_{\Omega} \in \mathbb{R}^{180}$, in which every element represents the strength of a response to a particular concept (computed based on the appropriate half of the stimuli only), always in the same order. We shuffle the elements of $\Vec{\beta}_{\Omega}$ for each paradigm $\Omega$ independently 1,000 times. Let $\Tilde{\Vec{\beta}}^{(k)}_{\Omega}$ denote the vector resulting from the $k-$th shuffling ($1 \leq k \leq 1000$); we can compute the ``shuffled'' correlation value $\Tilde{C}^{(k)}$ by substituting $\Tilde{\Vec{\beta}}^{(k)}_{\Omega}$ for $\Vec{\beta}_{\Omega}$ for each paradigm $\Omega$ in \eqref{eq:semcons}. We can then compute the one-sided $p-$value of the permutation test as $p = \sum_{k=1}^{1000} \mathbb{I}[C>\Tilde{C}^{(k)}]$, where $\mathbb{I}$ is an indicator function. 

\begin{figure}[htb]
\begin{center}
\includegraphics[width=0.8\linewidth]{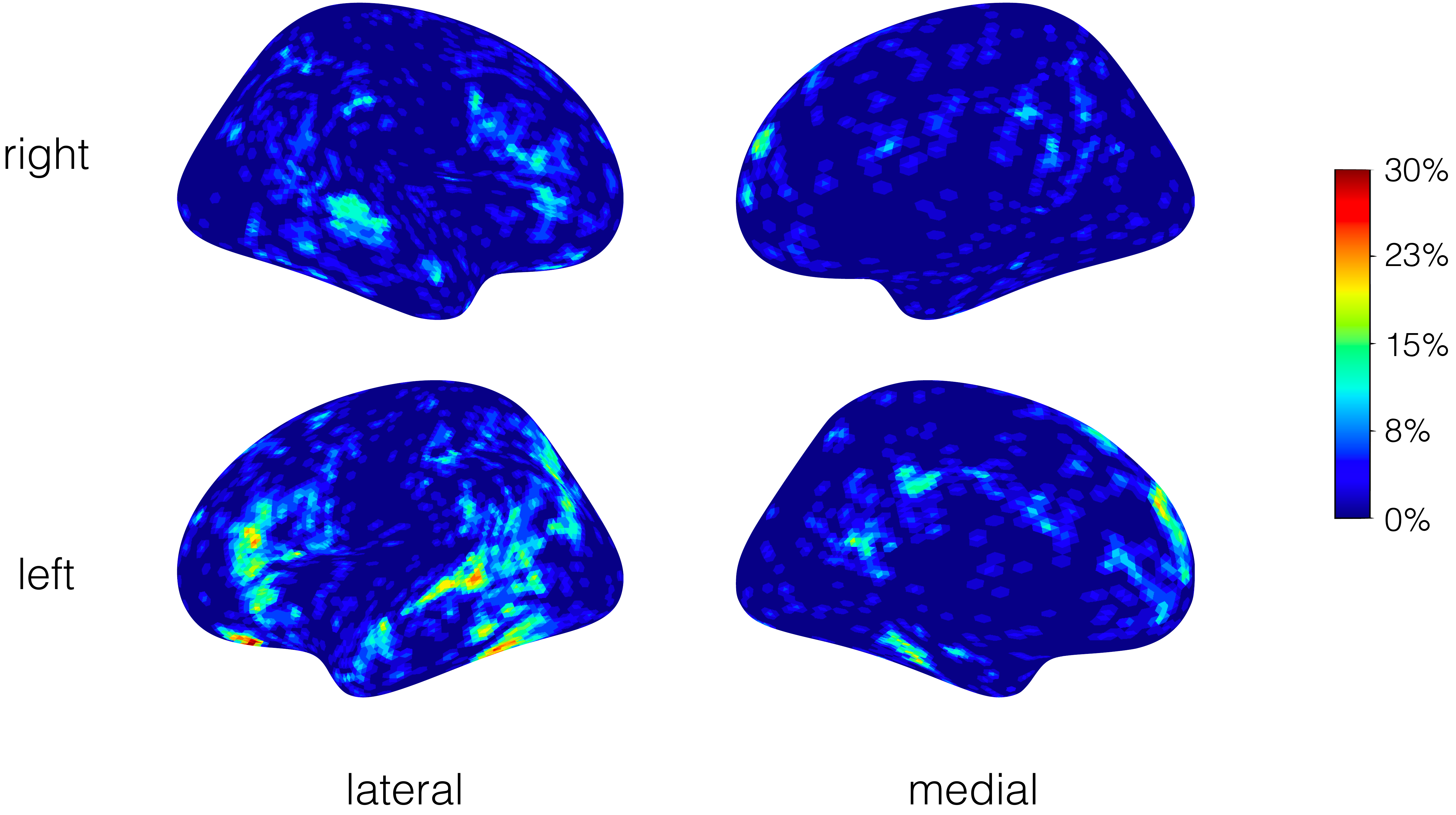}
\end{center}
\caption{{\bf Probabilistic map of voxels with statistically significant semantic consistency across all participants.} Each voxel's value shows the \% of the participants in whose brain this voxel has $p < 0.05$ in both permutation tests.}
\label{fig:prob_map}
\end{figure}

Doing so for every voxel in every participant's brain yields two $p$-values for voxel. We select for each participant the set of voxels that were statistically significant on both halves of the stimuli (i.e., both $p$-values were below 0.05) and convert it to a binarized 3D map (1 in statistically significant voxels, 0 otherwise). Finally, we average the obtained binary map over participants to obtain a probabilistic map of semantically consistent voxels, where a value corresponding to each voxel represents the percentage of participants in whose brain this voxel had statistically significant semantic consistency (Fig.~\ref{fig:prob_map}).

\subsection{Defining ROIs}
\label{sec:appendix-rois}

\begin{figure}[htb]
\begin{center}
\includegraphics[width=0.8\linewidth]{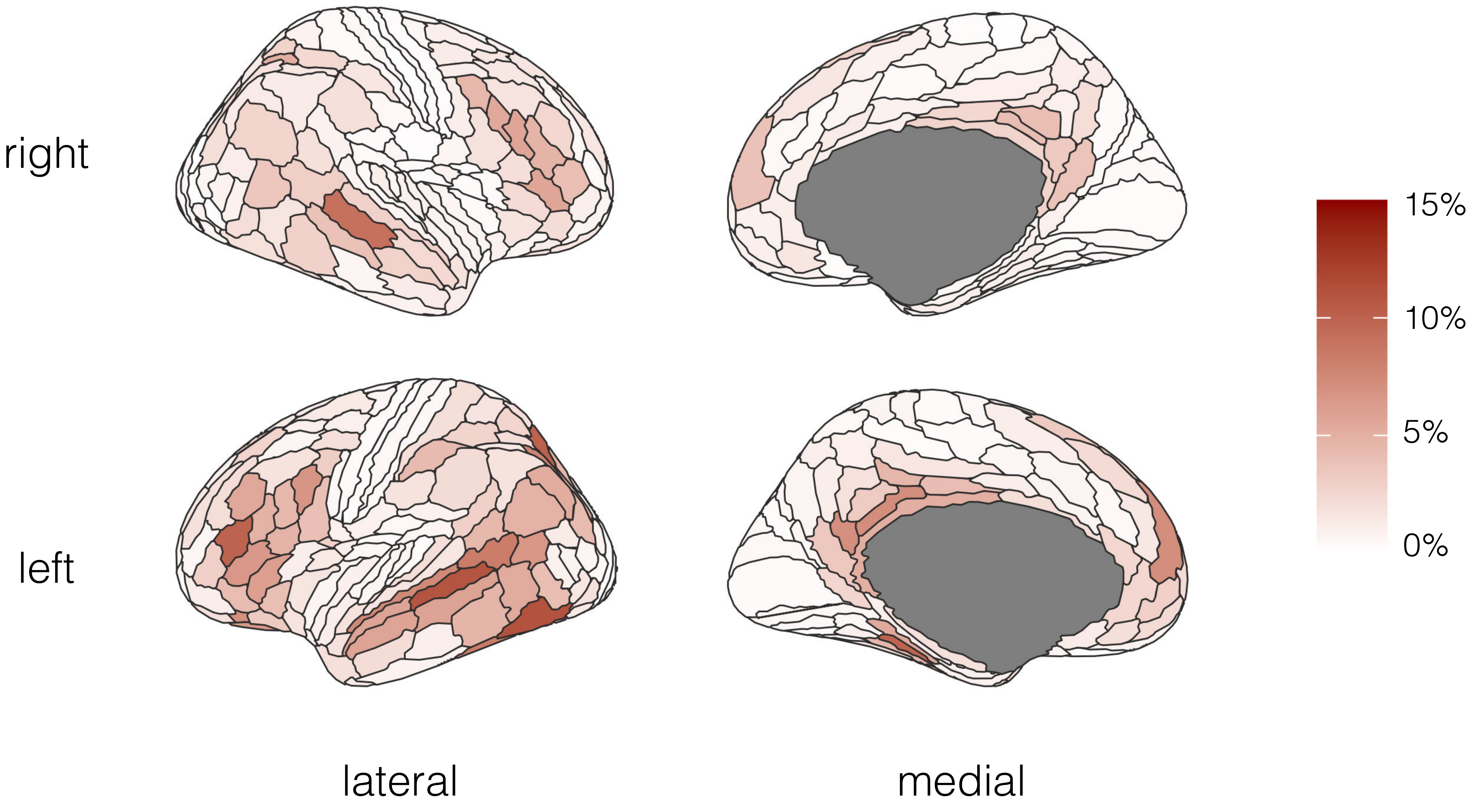}
\end{center}
\caption{{\bf The probabilistic map in \Figref{fig:prob_map}, averaged by anatomical area as defined by \cite{glasser2016multi}.} Thresholding these areas by probability, dividing the remaining ones into contiguous clusters, and filtering by size results in the three ROIs shown in \Figref{fig:2}c. }
\label{fig:glasser}
\end{figure}

We use the probabilistic map in \Figref{fig:prob_map} to define the boundaries of our regions of interest (ROIs). We use the anatomical parcellation of \citealp{glasser2016multi} \citep[in volumetric projection by][]{Horn2016} and average the values of the probabilistic map over all voxels in each anatomical area. The result is shown in \Figref{fig:glasser}. After discarding all Glasser areas in which the value is below 5.9\% (i.e., 1/17, where 17 is the number of participants), we are left with 19 anatomical areas forming 10 contiguous regions of the brain cortex;  of these regions lie in the left hemisphere. Finally, we filter these 10 regions by size ($>$600 voxels), which leaves the three left-hemisphere ROIs shown in \Figref{fig:2}c. The size and anatomical makeup of each ROI is listed in \Tref{tab:glasser-parcels}.

\begin{table}[t]
\begin{center}
\begin{tabular}{lp{3cm}cp{7cm}}
\toprule
\bf{ROI} & \bf{Location} & \bf{\# voxels} & {\bf Areas, named per \cite{glasser2016multi}}  \\
\midrule
\multirow{4}{*}{ROI 1} & \multirow{4}{*}{Superior temporal} &  \multirow{4}{*}{975} & Auditory 5 Complex (A5) \\
& & & Area STSd posterior (STSdp)\\
& & & Area Temporo-Parieto-Occipital Junction 1 (TPOJ1) \\
& & & Area Temporo-Parieto-Occipital Junction 2 (TPOJ2)\\
\midrule
\multirow{3}{*}{ROI 2} & \multirow{3}{*}{Inferior frontal} & \multirow{3}{*}{675} & Area IFSa (IFSa) \\
& & & Area 45 (45) \\
& & & Area Frontal Opercular 5 (FOP5)\\
\midrule
\multirow{2}{*}{ROI 3} & \multirow{ 2}{*}{Ventral temporal} & \multirow{2}{*}{646}& Area TE2 posterior (TE2p)\\
& & & Area PH (PH)\\
\bottomrule
\end{tabular}
\caption{\textbf{The breakdown of the three identified left-hemisphere regions of interest (ROIs), shown in \Figref{fig:2}c.} The individual area definitions follow \cite{glasser2016multi}.
\label{tab:glasser-parcels}}
\end{center}
\end{table}

\subsection{Language response in semantic consistency ROIs}
\label{sec:appendix-lang-response}

Mean language selectivity (measured per \sref{sec:factors}) for each ROI is reported in \Tref{tab:roi-lang-response}.

\begin{table}[ht]
\begin{center}
\begin{tabular}{l wr{2cm}}
\toprule
ROI & \multicolumn{1}{c}{$\Delta \beta_{\text{Sentences,Non-words}}$}\\
\midrule
ROI 1 & $0.59 \pm 0.06$\\
ROI 2 & $0.13 \pm 0.07$\\
ROI 3 & $-0.16 \pm 0.04$\\
\bottomrule
\end{tabular}
\caption{\label{tab:roi-lang-response} \textbf{Mean (over voxels and participants) language selectivity in the semantic consistency ROIs.} The value is measured as the effect size for the sentences vs. non-words contrast of \citeauthor{fedorenko2010new}'s (\citeyear{fedorenko2010new}) language localizer. Standard error is shown over participants.}
\end{center}
\end{table}

\section{Models and methods}
\subsection{Sources and implementation}
\label{sec:appendix-sources}

All alignment experiments are implemented using the \texttt{numpy} \citep{numpy}, \texttt{scipy} \citep{scipy}, \texttt{scikit-learn}  \citep{sklearn}, and \texttt{pandas} \citep{pandas} libraries. For brain encoding, we use the \texttt{RidgeCV} class in \texttt{scikit-learn} to automatically tune the regularization hyperparameter $\alpha$ via leave-one-out cross-validation.

We download all pretrained models from the HuggingFace Hub.\footnote{\url{https://huggingface.co/}} \Tref{tab:models} includes the links to the model repositories on HuggingFace. All experiments involving LMs are performed using PyTorch \citep{NEURIPS2019_bdbca288} and the \texttt{transformers} Python library \citep{wolf-etal-2020-transformers}.

\newcommand\hflink[1]{\href{https://huggingface.co/#1}{\texttt{#1}}}

\begin{table}[htb]
\begin{center}
\begin{tabular}{llr}
\toprule
\textbf{Model} & \textbf{HuggingFace ID} & \textbf{Parameters} \\
\midrule
GPT-2 & \hflink{openai-community/gpt2} & 124M\\
GPT-2 Medium & \hflink{openai-community/gpt2-medium} & 355M\\
GPT-2 Large & \hflink{openai-community/gpt2-large} & 774M\\
GPT-2 XL & \hflink{openai-community/gpt2-xl} & 1.6B\\
FLAVA & \hflink{facebook/flava-full} & 241M\\
Vicuna-1.5-7B & \hflink{lmsys/vicuna-7b-v1.5} & 7B \\
LLaVA-1.5-7B & \hflink{llava-hf/llava-1.5-7b-hf} & 7B \\
Qwen2.5-1.5B & \hflink{Qwen/Qwen2.5-1.5B} & 1.5B \\
Qwen2.5-1.5B-Instruct & \hflink{Qwen/Qwen2.5-1.5B-Instruct} & 1.5B \\
Qwen2.5-3B & \hflink{Qwen/Qwen2.5-3B} & 3B \\
Qwen2.5-3B-Instruct & \hflink{Qwen/Qwen2.5-3B-Instruct} & 3B \\
Qwen2.5-VL-3B-Instruct & \hflink{Qwen/Qwen2.5-VL-3B-Instruct} & 3B \\
Qwen2.5-7B & \hflink{Qwen/Qwen2.5-7B} & 7B \\
Qwen2.5-7B-Instruct & \hflink{Qwen/Qwen2.5-7B-Instruct} & 7B \\
Qwen2.5-VL-7B-Instruct & \hflink{Qwen/Qwen2.5-VL-7B-Instruct} & 7B \\
\bottomrule
\end{tabular}
\caption{\textbf{Pretrained language models used in this study}. We provide the HuggingFace identifier and a hyperlink for downloading each model's weights.}
\label{tab:models}
\end{center}
\end{table}

\subsection{Stimuli input format}
\label{sec:appendix-embedding}

To input the stimuli from each paradigm (Fig.~\ref{fig:stimuli}) into the models, we format them as follows:

\begin{itemize}
\item Sentences are inputted as-is: \texttt{The bird flew around the cage.} 
\item Word clouds are presented as a sequence of space-separated words, with the concept word given first: \texttt{bird nest flock mating beak winged.}\\
Since all word clouds for the same concept contain the same words, we use a single sequence to represent them all. 
\item For picture + concept word inputs we add special VLM image tokens where needed:\\
\texttt{<image> Bird} (LLaVA format) or \texttt{<|vision\_start|><|image\_pad|><|vision\_end|> Bird} (Qwen-VL format).
\end{itemize}

\section{Brain encoding performance}

\subsection{Whole-brain correlation with semantic consistency}
\label{sec:appendix-raw-c}

\Figref{fig:predict_glasser} in \Sref{sec:exp1} shows how mean LM predictivity and semantic consistency correlate across anatomical areas. The semantic consistency of an area is measured by (1) obtaining the probabilistic map of reliably consistent voxels in that area across participants (\Sref{sec:appendix-prob-map}, Fig.~\ref{fig:prob_map}) and (2) averaging it over all voxels in an area (Fig.~\ref{fig:glasser}).

\begin{figure}[htb]
\begin{center}
\includegraphics[width=\linewidth]{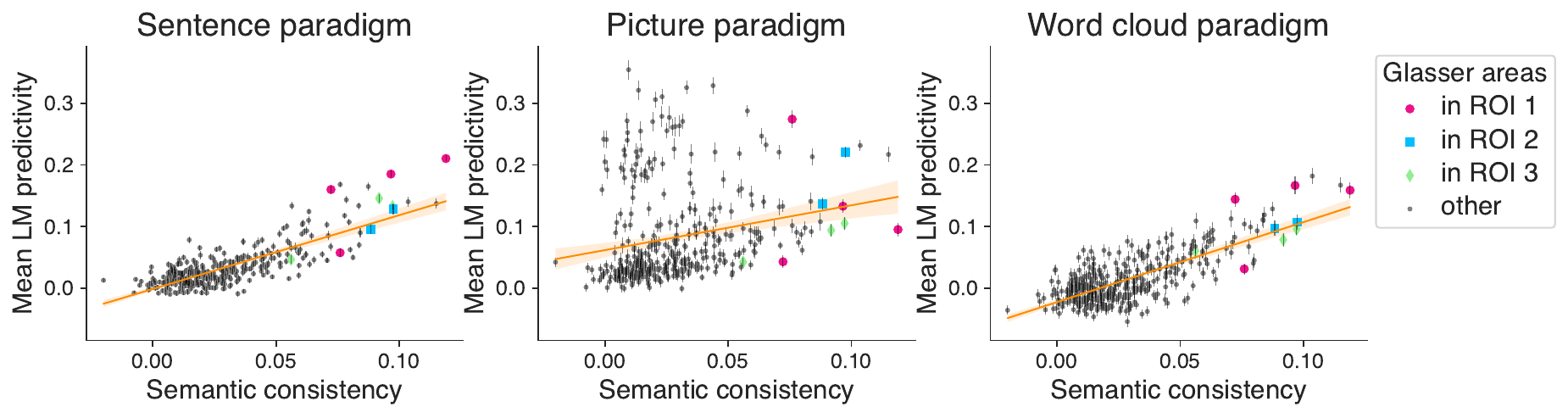}
\end{center}
\caption{\textbf{LM predictivity per \cite{glasser2016multi} area, with semantic consistency (x-axis) showing the mean value of the metric} $C$. The Pearson correlations between $C$ and predictivity for each paradigm are: $r[\text{\textsc{s}}] = 0.80, r[\text{\textsc{p}}] = 0.22, r[\text{\textsc{wc}}] = 0.75$.}
\label{fig:predict_glasser_c}
\end{figure}

However, since our main brain encoding experiment (Fig.~\ref{fig:quartiles}) uses the raw value of $C$ rather than the probabilistic map to group voxels, we rerun this analysis with $C$ as the measure of semantic consistency. As can be expected, these two consistency measures are highly correlated ($r=0.83$). The result is shown in \Figref{fig:predict_glasser_c}.

\subsection{Inter-participant noise ceiling}
\label{sec:appendix-noise-ceiling}

In the whole-brain encoding experiment (Fig.~\ref{fig:predict_glasser}), we correlate LM predictivity in each anatomical area \citep[defined per][]{glasser2016multi} with its level of semantic consistency. However, the  higher predictivity might be not only due to increased brain--LM alignment: brain activations in some areas might be better predicted than in others because the signal there is simply less noisy. One standard approach to estimate a noise ceiling is to quantify the variability between the responses to the same stimulus in a given area of the same participant's brain (repeated trials). In our case, no participant sees the same stimulus more than once, i.e., trials are never repeated; instead, we compute the \emph{inter-participant} noise ceiling, estimating the variability in responses to the same stimulus across all participants. We follow the procedure described by \citet[][section SI 5]{tuckute2024driving} to obtain an across-participant ``noise ceiling''.

When we divide the mean LM predictivity in each area by its estimated noise ceiling, we find that the correlations with the area's semantic consistency (mean probability of the area's voxels having statistically significant consistency) remain positive: $r[\text{\textsc{s}}] = 0.46, r[\text{\textsc{p}}] = 0.02, r[\text{\textsc{wc}}] = 0.63$. Although other noise ceiling estimation approaches could offer additional insights (though they may not be feasible given the experimental design), these results confirm that our key findings hold even after accounting for cross-participant reliability.

\subsection{Brain encoding across model layers}
\label{sec:appendix-layers}

\begin{figure}[htb]
\begin{center}
\includegraphics[width=\linewidth]{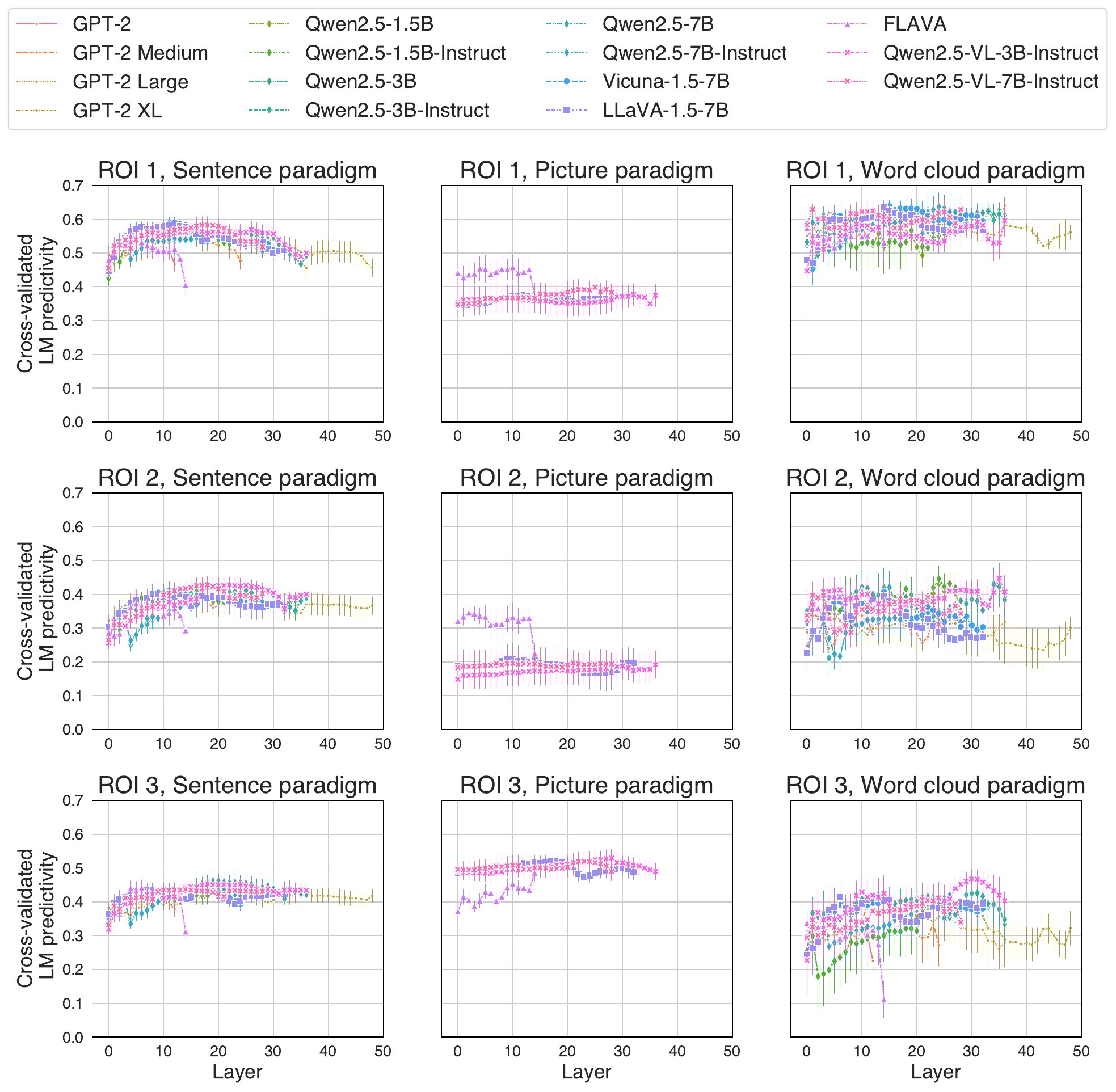}
\end{center}
\caption{\textbf{Brain encoding performance by LM layer (with mean pooling over tokens).} The target brain region (ROI) activations are averaged over all participants (\sref{sec:brain-enc}). The error bars show standard error of the mean over the five cross-validation folds.}
\label{fig:best-layer-mean}
\end{figure}

\begin{figure}[htb]
\begin{center}
\includegraphics[width=\linewidth]{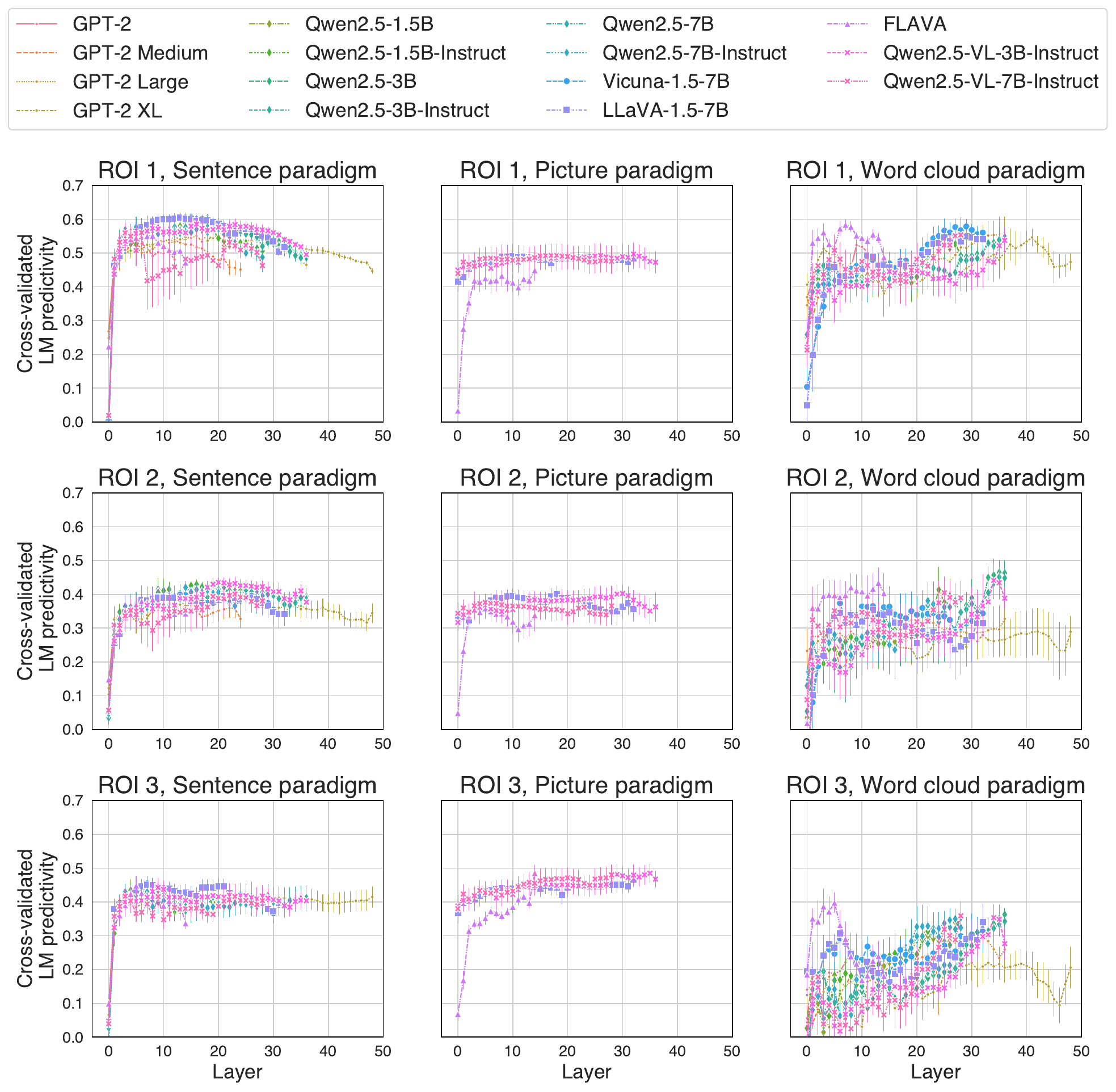}
\end{center}
\caption{\textbf{Brain encoding performance by LM layer (with last token pooling).} The target brain region (ROI) activations are averaged over all participants (\sref{sec:brain-enc}). The error bars show standard error of the mean over the five cross-validation folds.}
\label{fig:best-layer-last-tok}
\end{figure}

Figures~\ref{fig:best-layer-mean} and \ref{fig:best-layer-last-tok} show how well the the participant-average brain activations in the chosen ROIs can be predicted from the representations extracted from different model layers (with mean and last-token pooling respectively). For each model, we choose the layer and the token pooling method that together yield the best performance, and use this setting for all other experiments in this paper. The middle layers are typically the most predictive, consistent with the observations of \cite{caucheteux2022brains} and \cite{tuckute2024driving}. 

The mean-pooled embedding layer (layer 0 in Fig.~\ref{fig:best-layer-mean}) serves as a baseline: it shows how well brain activations can be predicted from the non-contextual embeddings of the individual tokens. As expected, the predictivity at layer 0 is lower for sentences (where the syntactic information is important for reconstructing the meaning), but less so for word clouds or pictures (since layer 0 in VLMs includes the projected features from the vision encoder).

\subsection{Brain encoding by voxel quartile}
\label{appendix-indmodels}
\begin{figure}[thbp]
\begin{center}
\includegraphics[width=\linewidth]{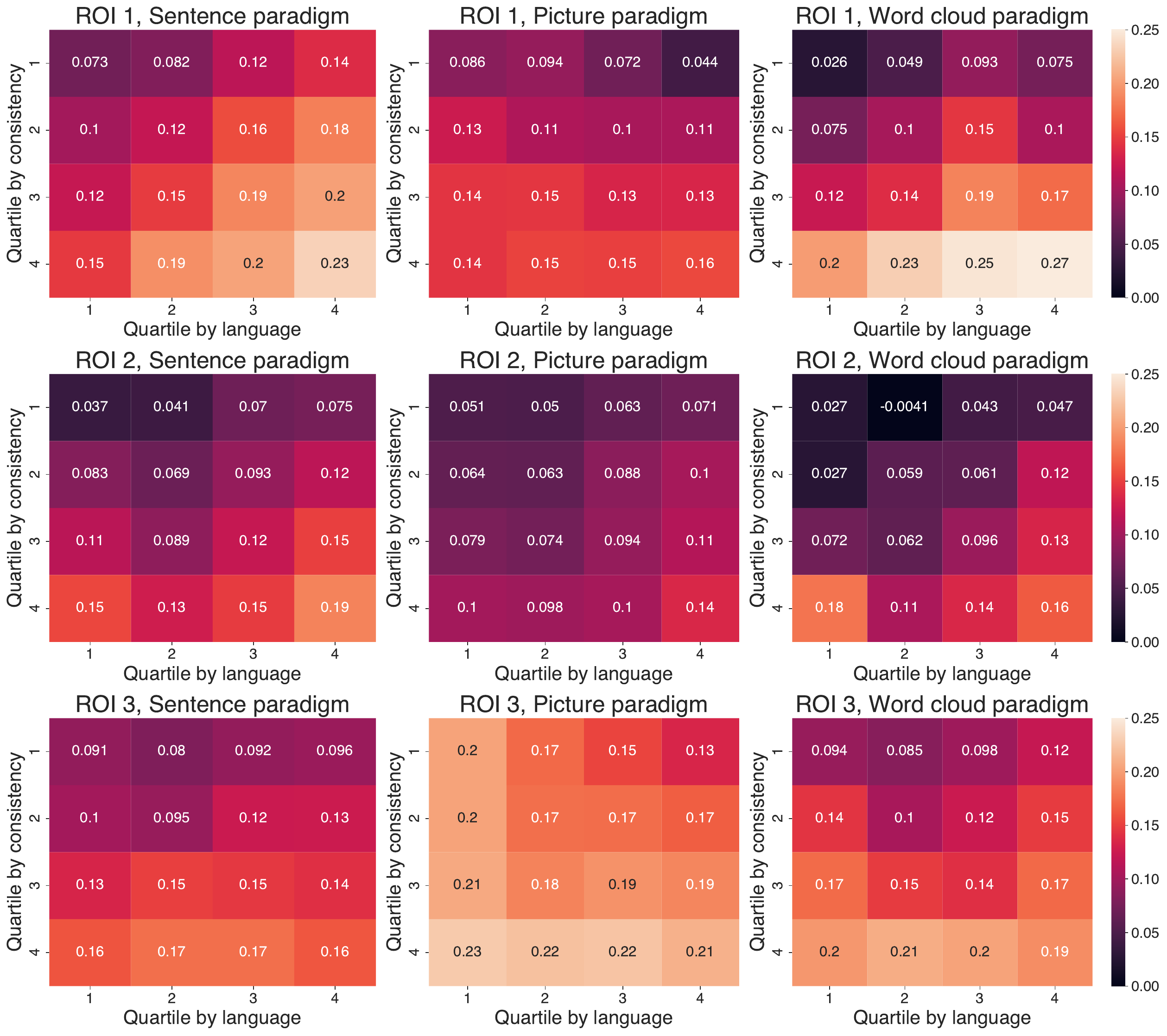}
\end{center}
\caption{\textbf{LM predictivity by quartile for each ROI and paradigm, averaged over all models.} The values shown reflect the mean correlation the ground-truth brain activations and the ones predicted from LM representations. 
In each heatmap, the x and y axes correspond to quartiles by language selectivity (sentences vs. non-words contrast; see \Sref{sec:factors}) and by semantic consistency $C$ respectively. Each cell of the grid shows the predictivity level on all voxels in a ROI that fall at the intersection of the given language selectivity and consistency quartiles. In ROI 1 (top row) and ROI 2 (middle row), the predictivity correlates with both language selectivity and semantic consistency, although the former is weaker for the word cloud (left column) and picture (middle column) paradigms. In ROI 3 (bottom row), only semantic consistency correlates with predictivity.
}
\label{fig:heatmap}
\end{figure}

\Figref{fig:heatmap} shows the average (over models and participants) predictivity values per semantic consistency or language selectivity quartile  (also shown in \Figref{fig:quartiles}), visualized as a heatmap.

Figures \ref{fig:linecharts_gpt2}--\ref{fig:linecharts_qwen-7b-vl-instruct} show how each LM's predictivity varies by language selectivity and semantic consistency quartile. Odd-numbered columns show how predictivity in each ROI changes across voxel quartiles by language selectivity, with each line corresponding to one semantic consistency quartile. Even-numbered columns show how predictivity changes across voxel quartiles by semantic consistency, with each line corresponding to one language selectivity quartile. The thickness of the line corresponds to the quartile (thicker=higher), and the error intervals show standard error across participants. Each plot is averaged over participants; average over all models is shown in \Figref{fig:quartiles}. The correlations with both semantic consistency ($C$) and language selectivity ($L$) for each ROI are reported in the caption (averaged over paradigms).

\subsection{Brain encoding in the language network}
\label{sec:appendix-encoding-lang-rois}

\begin{figure}[htb]
\begin{center}
\includegraphics[width=\linewidth]{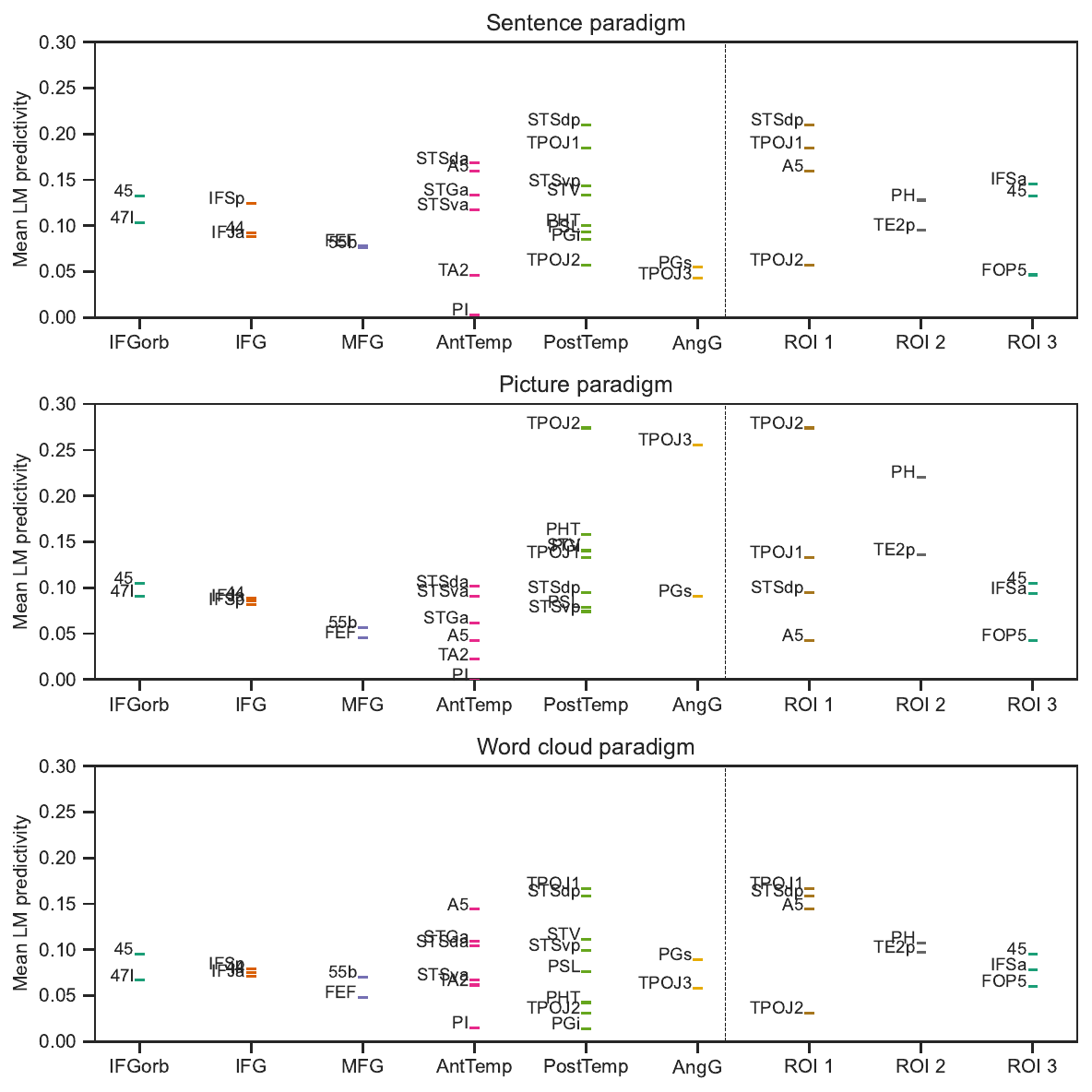}
\end{center}
\caption{\textbf{Brain encoding performance in the left-hemisphere language network parcels and the semantic consistency ROIs}. For details, see Appendix \ref{sec:appendix-encoding-lang-rois}. LM predictivity is averaged over all models and participants. Each data point displayed corresponds to an anatomical area of \cite{glasser2016multi}. The language parcels from prior work are shown on the left of the dashed line, and the semantic consistency ROIs are shown on the right. The \cite{glasser2016multi} areas on the left are chosen by overlap ($>25\%$) with language parcels.}
\label{fig:lang-glasser}
\end{figure}

For comparison with prior works on LM--brain alignment that target the brain's language network, we conduct an additional analysis comparing the brain encoding performance in the left-hemisphere regions often engaged by linguistic processing \citep{fedorenko2010new,mahowald2016reliable,lipkin2022probabilistic} with that in the semantic consistency ROIs identified in this paper. We use the six language parcels (located in the inferior frontal gyrus, orbital inferior frontal gyrus, middle frontal gyrus, anterior temporal lobe, posterior temporal lobe, and angular gyrus) created from a probabilistic overlap map from 220 participants.\footnote{Downloaded from \url{https://evlab.squarespace.com/s/allParcels-language-SN220.nii}}
Since the semantic consistency ROIs are defined as sets of \cite{glasser2016multi} anatomical areas, we also identify all \cite{glasser2016multi} areas that overlap substantially (by over 25\% of an area's voxels) with any of the language parcels. If an area overlaps with more than one language parcel, we assign it to the parcel with the highest overlap. The brain encoding results are presented in \Figref{fig:lang-glasser}, reported by \cite{glasser2016multi} area.

\section{RSA performance}
\label{sec:rsa-semcons}
\begin{figure}[htb]
\begin{center}
\includegraphics[width=\linewidth]{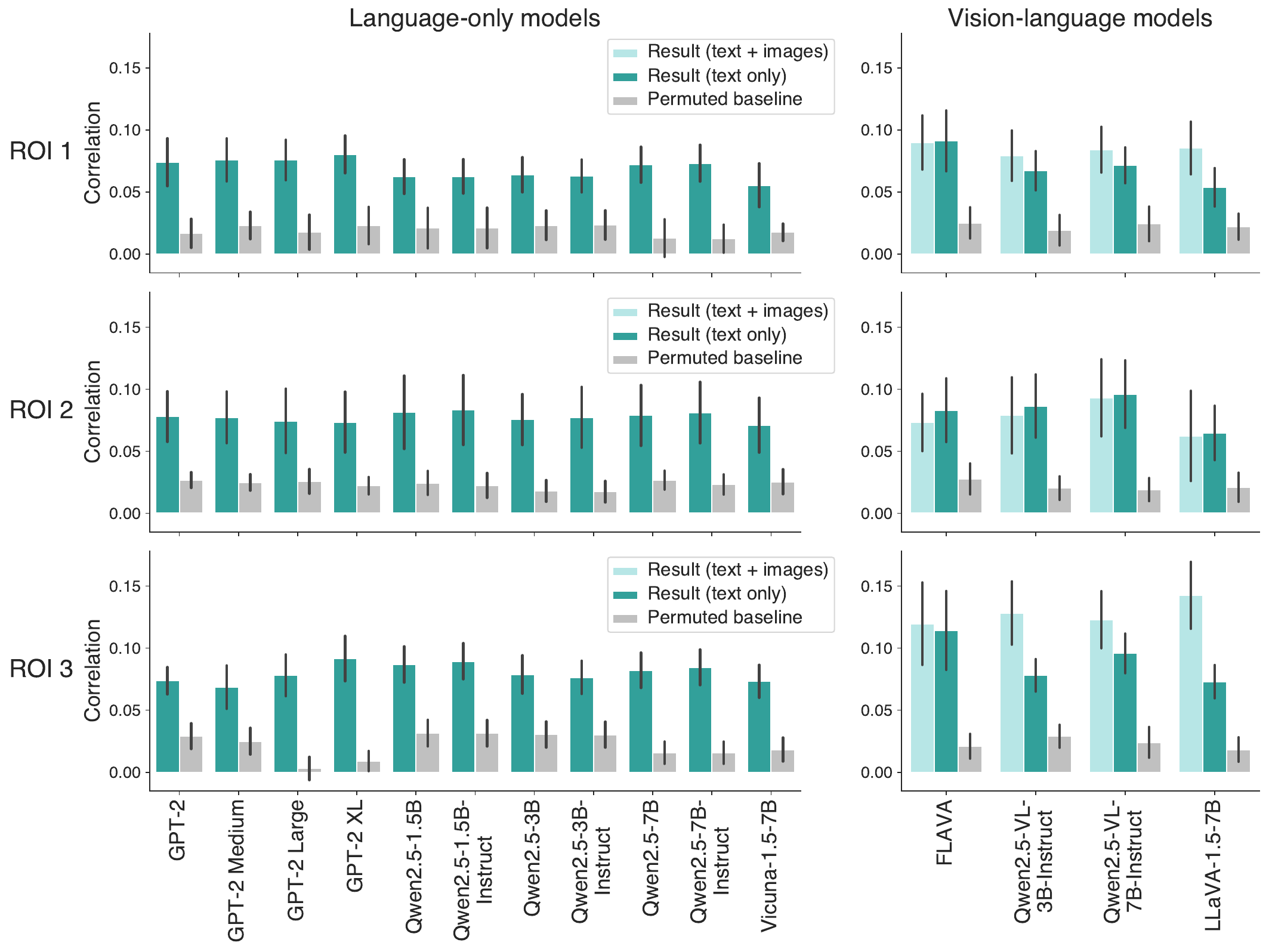}
\end{center}
\caption{
\textbf{RSA scores for each model when using only voxels with significant semantic consistency (\sref{sec:appendix-prob-map}).} For more details, see Appendix \ref{sec:rsa-semcons} and \Figref{fig:rsa}. The three conditions correspond to using all three paradigms (text + images), using only sentences and word clouds (text only), and the baseline where the concepts are shuffled on one of the sides before computing correlations (\Sref{sec:exp2}).
The results are consistent with those from using all voxels in each ROI (\Figref{fig:rsa}), although the gains from adding images are reduced for non-visual ROIs 1 and 2. Error bars show standard error over participants.
}
\label{fig:rsa_fair_semcons}
\end{figure}

To complement the analysis in \Sref{sec:exp2}, we conduct an additional RSA experiment using only the voxels with statistically significant semantic consistency (\Sref{sec:appendix-prob-map}). This dramatically reduces the number of voxels to $\sim$10--20 per ROI in most participants. The results (mean and SEM across participants) are shown in \Figref{fig:rsa_fair_semcons}. While the overall trends remain the same (see \Figref{fig:rsa}), the VLM--ROI alignment gain from adding the picture paradigm data has decreased in the non-visual ROIs (1 and 2).

\newcommand\redc{\mathcolor{indianred}{C}}
\newcommand\bluel{\mathcolor{royalblue}{L}}

\begin{figure}[htb]
\begin{center}
\includegraphics[width=0.7\linewidth]{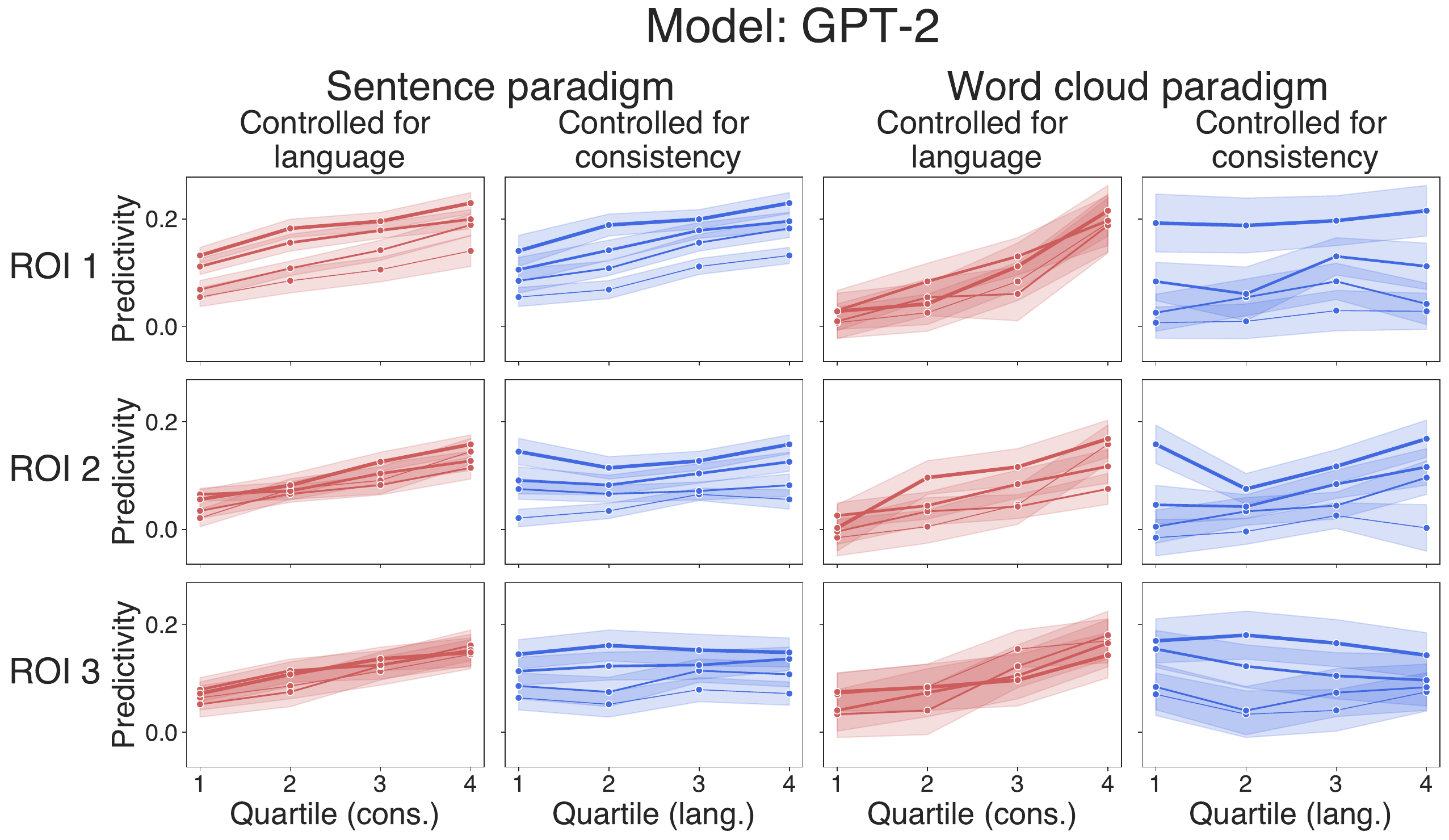}
\end{center}
\caption{\textbf{GPT-2 predictivity by voxel quartile}. 
ROI 1: $r_{\redc}=0.41 \pm 0.02, r_{\bluel}=0.24 \pm 0.07$.
ROI 2: $r_{\redc}=0.38 \pm 0.02, r_{\bluel}=0.15 \pm 0.03$.
ROI 3: $r_{\redc}=0.27 \pm 0.02, r_{\bluel}=0.01 \pm 0.03$.}
\label{fig:linecharts_gpt2}
\end{figure}

\begin{figure}[htb]
\begin{center}
\includegraphics[width=0.7\linewidth]{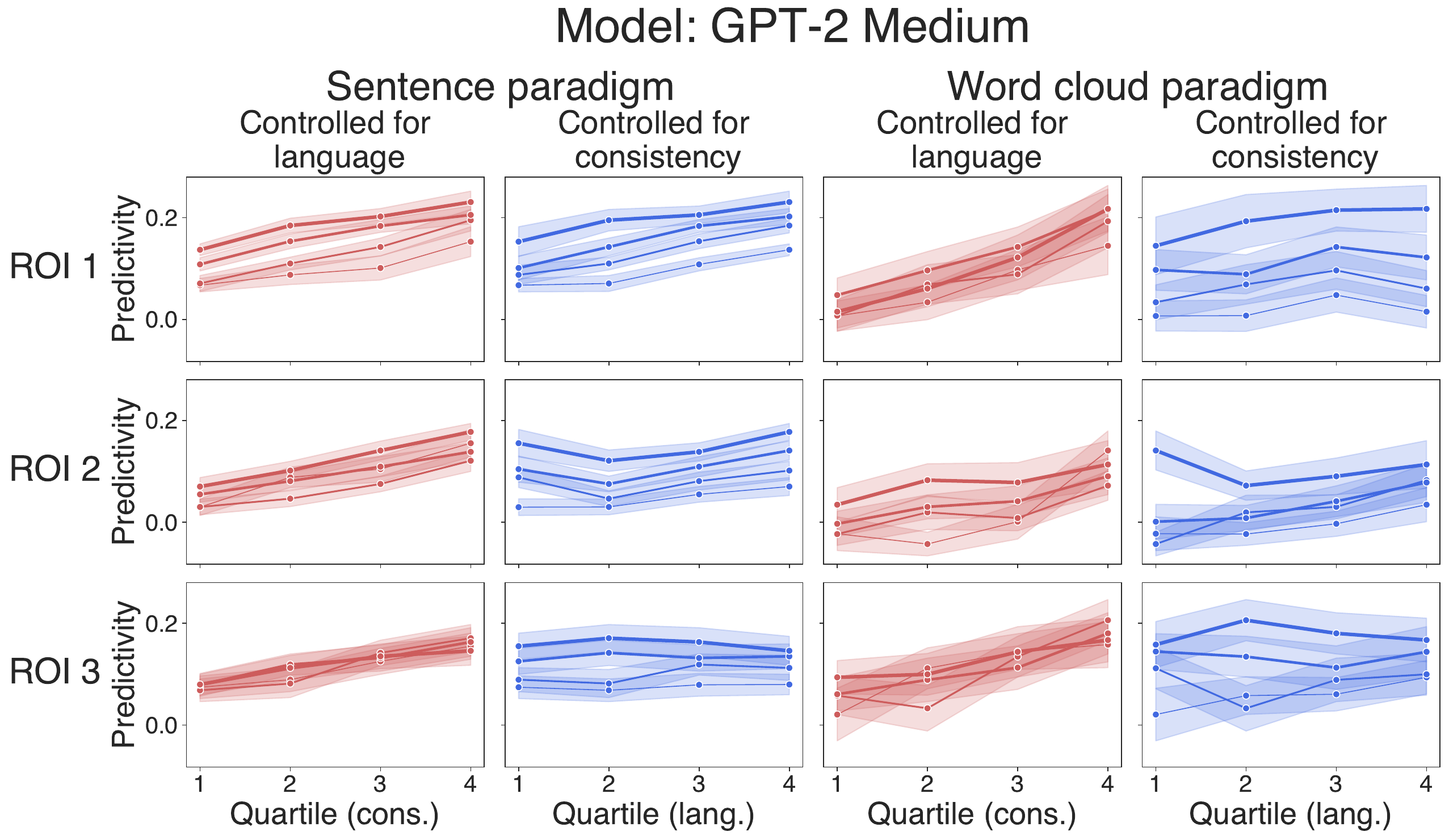}
\end{center}
\caption{\textbf{GPT-2 Medium predictivity by voxel quartile}. 
ROI 1: $r_{\redc}=0.42 \pm 0.03, r_{\bluel}=0.26 \pm 0.07$.
ROI 2: $r_{\redc}=0.36 \pm 0.04, r_{\bluel}=0.17 \pm 0.04$.
ROI 3: $r_{\redc}=0.28 \pm 0.02, r_{\bluel}=0.04 \pm 0.02$.
}
\label{fig:linecharts_gpt2-medium}
\end{figure}

\begin{figure}[htb]
\begin{center}
\includegraphics[width=0.7\linewidth]{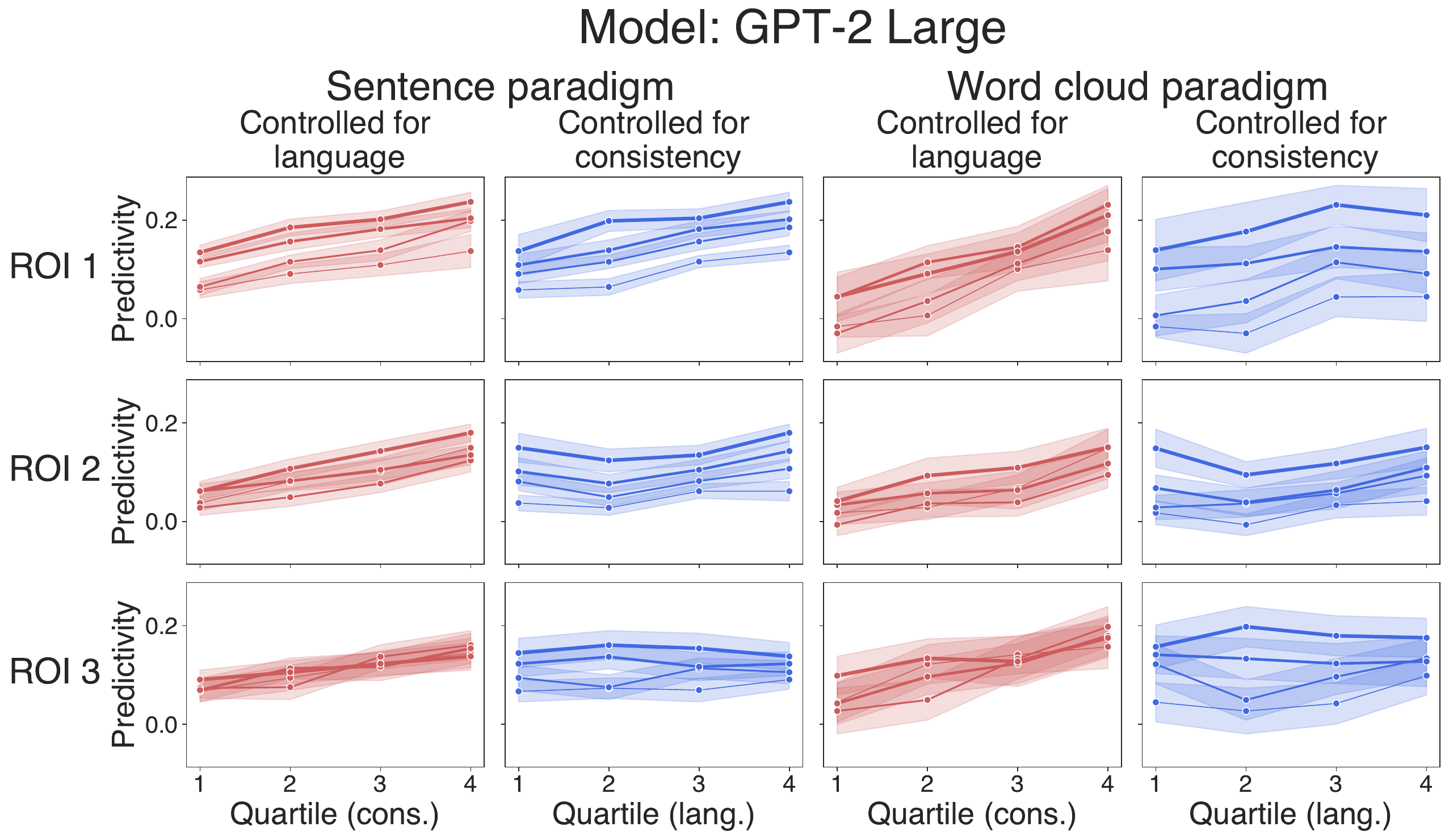}
\end{center}
\caption{\textbf{GPT-2 Large predictivity by voxel quartile}. 
ROI 1: $r_{\redc}=0.40 \pm 0.03, r_{\bluel}=0.29 \pm 0.05$.
ROI 2: $r_{\redc}=0.36 \pm 0.03, r_{\bluel}=0.14 \pm 0.02$.
ROI 3: $r_{\redc}=0.27 \pm 0.03, r_{\bluel}=0.04 \pm 0.02$.
}
\label{fig:linecharts_gpt2-large}
\end{figure}

\begin{figure}[htb]
\begin{center}
\includegraphics[width=0.7\linewidth]{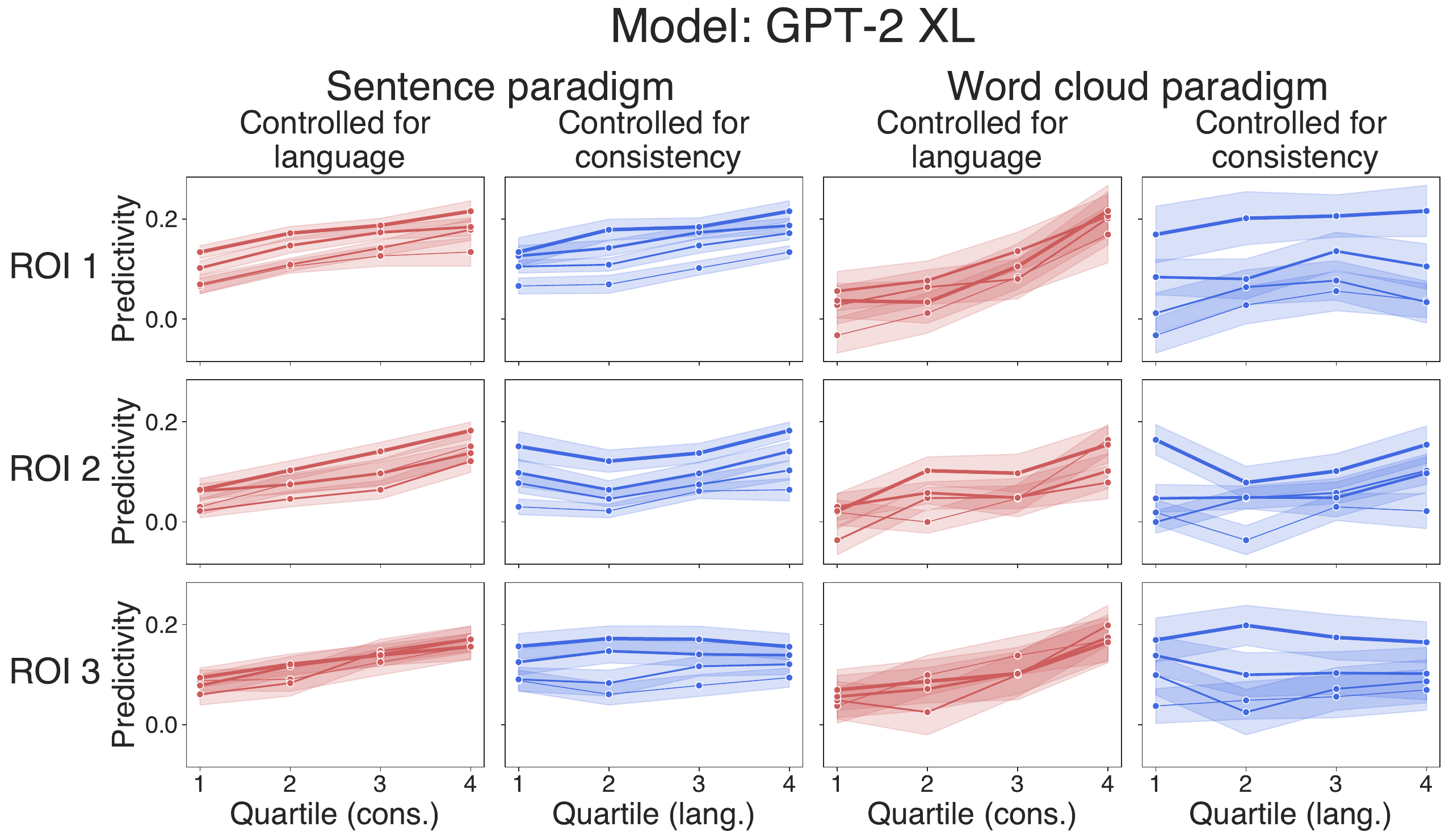}
\end{center}
\caption{\textbf{GPT-2 XL predictivity by voxel quartile}. 
ROI 1: $r_{\redc}=0.39 \pm 0.02, r_{\bluel}=0.23 \pm 0.05$.
ROI 2: $r_{\redc}=0.37 \pm 0.04, r_{\bluel}=0.16 \pm 0.04$.
ROI 3: $r_{\redc}=0.29 \pm 0.03, r_{\bluel}=0.03 \pm 0.02$.
}
\label{fig:linecharts_gpt2-xl}
\end{figure}

\begin{figure}[htb]
\begin{center}
\includegraphics[width=0.7\linewidth]{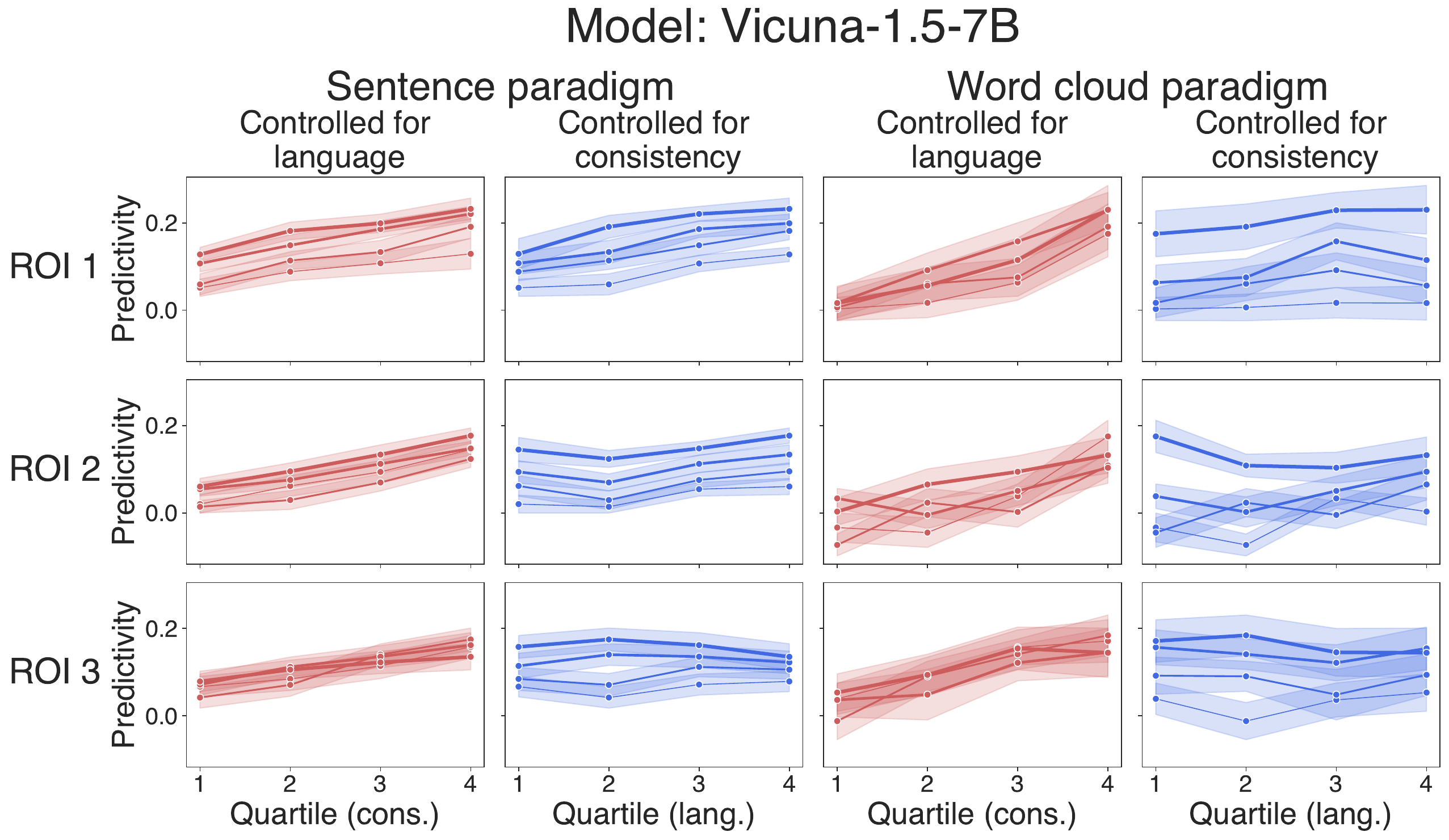}
\end{center}
\caption{\textbf{Vicuna-1.5-7B predictivity by voxel quartile}. 
ROI 1: $r_{\redc}=0.39 \pm 0.02, r_{\bluel}=0.23 \pm 0.05$.
ROI 2: $r_{\redc}=0.42 \pm 0.03, r_{\bluel}=0.16 \pm 0.04$.
ROI 3: $r_{\redc}=0.30 \pm 0.03, r_{\bluel}=0.01 \pm 0.03$.}
\label{fig:linecharts_vicuna-7b}
\end{figure}

\begin{figure}[htb]
\begin{center}
\includegraphics[width=0.7\linewidth]{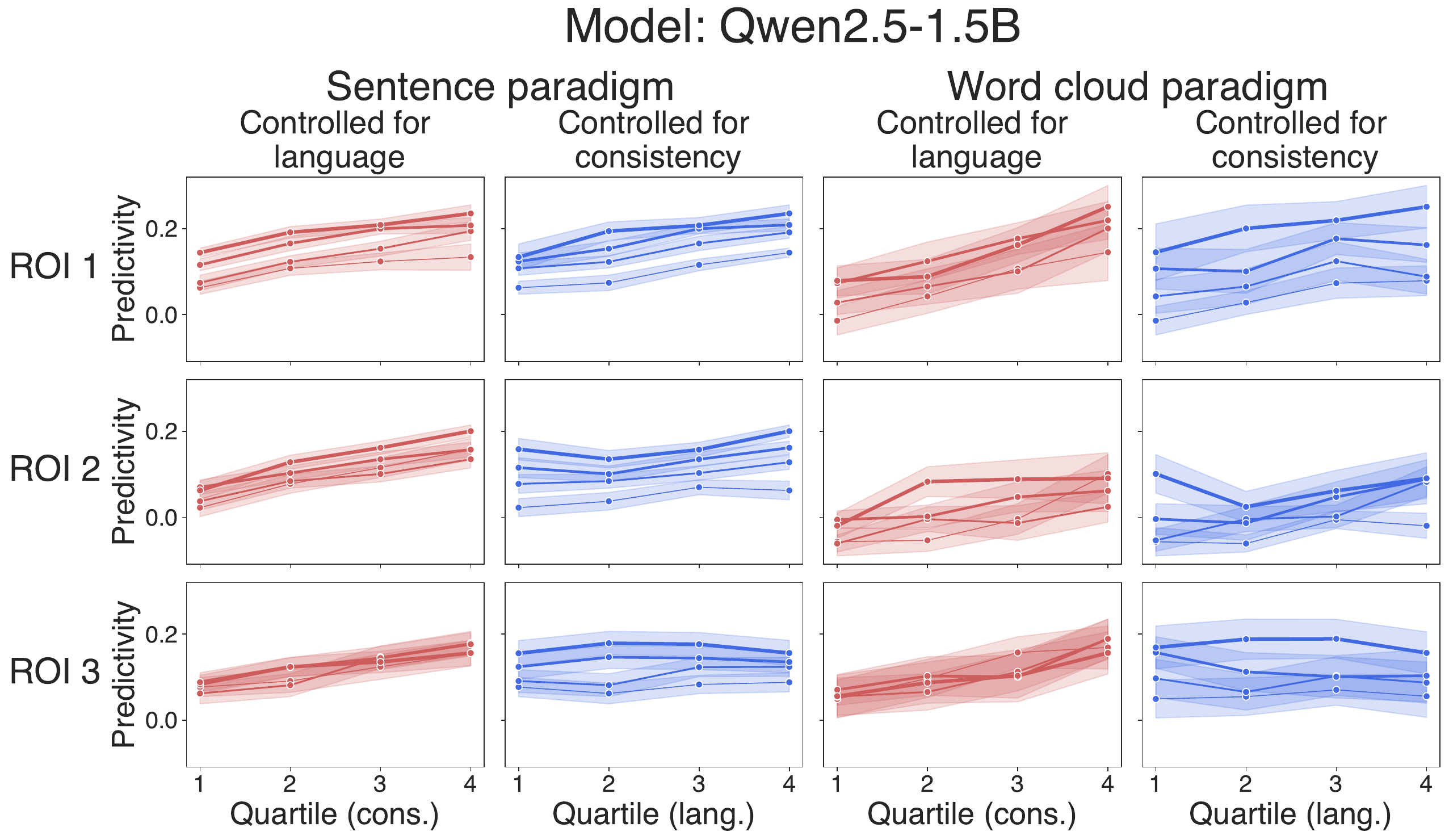}
\end{center}
\caption{\textbf{Qwen2.5-1.5B predictivity by voxel quartile.}
ROI 1: $r_{\redc}=0.39 \pm 0.03, r_{\bluel}=0.31 \pm 0.05$.
ROI 2: $r_{\redc}=0.36 \pm 0.05, r_{\bluel}=0.21 \pm 0.04$.
ROI 3: $r_{\redc}=0.27 \pm 0.03, r_{\bluel}=0.02 \pm 0.03$.
}
\label{fig:linecharts_qwen-1.5b}
\end{figure}

\begin{figure}[htb]
\begin{center}
\includegraphics[width=0.7\linewidth]{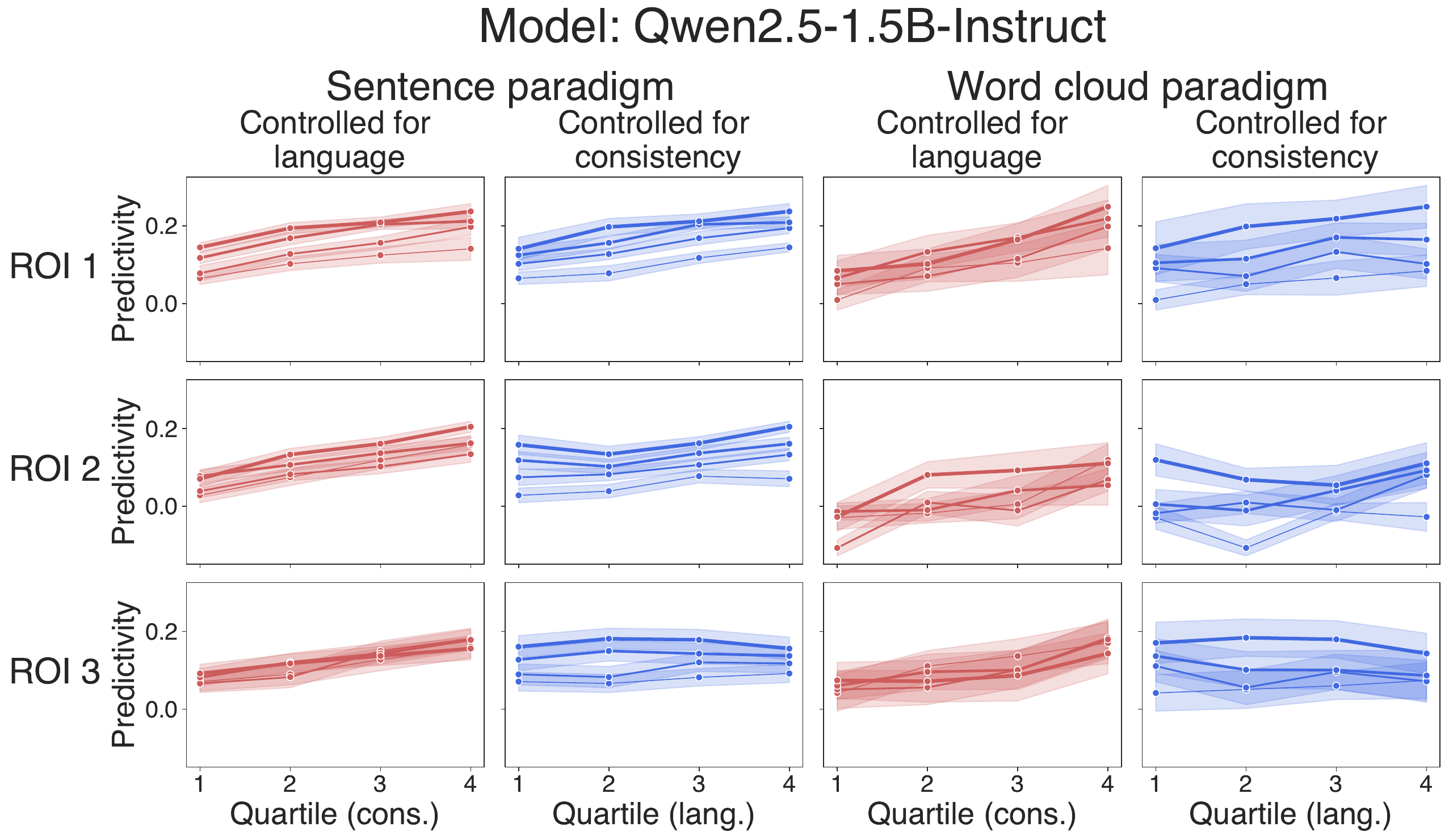}
\end{center}
\caption{\textbf{Qwen2.5-1.5B-Instruct predictivity by voxel quartile.}
ROI 1: $r_{\redc}=0.37 \pm 0.04, r_{\bluel}=0.29 \pm 0.06$.
ROI 2: $r_{\redc}=0.39 \pm 0.04, r_{\bluel}=0.19 \pm 0.04$.
ROI 3: $r_{\redc}=0.26 \pm 0.03, r_{\bluel}=0.02 \pm 0.03$.
}
\label{fig:linecharts_qwen-1.5b-instruct}
\end{figure}

\begin{figure}[htb]
\begin{center}
\includegraphics[width=0.7\linewidth]{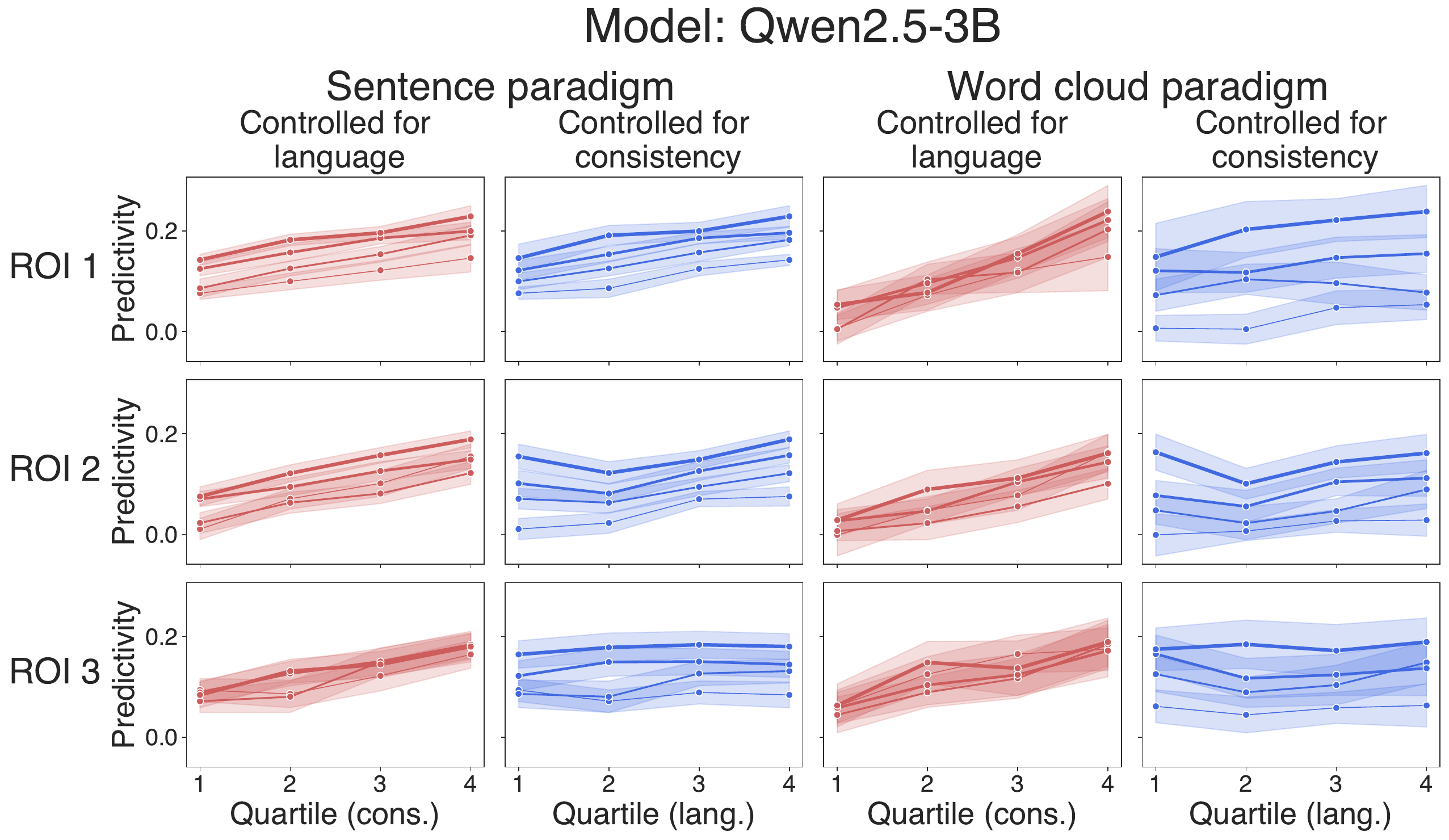}
\end{center}
\caption{\textbf{Qwen2.5-3B predictivity by voxel quartile.}
ROI 1: $r_{\redc}=0.39 \pm 0.02, r_{\bluel}=0.25 \pm 0.06$.
ROI 2: $r_{\redc}=0.40 \pm 0.03, r_{\bluel}=0.18 \pm 0.04$.
ROI 3: $r_{\redc}=0.29 \pm 0.02, r_{\bluel}=0.04 \pm 0.02$.}
\label{fig:linecharts_qwen-3b}
\end{figure}

\begin{figure}[htb]
\begin{center}
\includegraphics[width=0.7\linewidth]{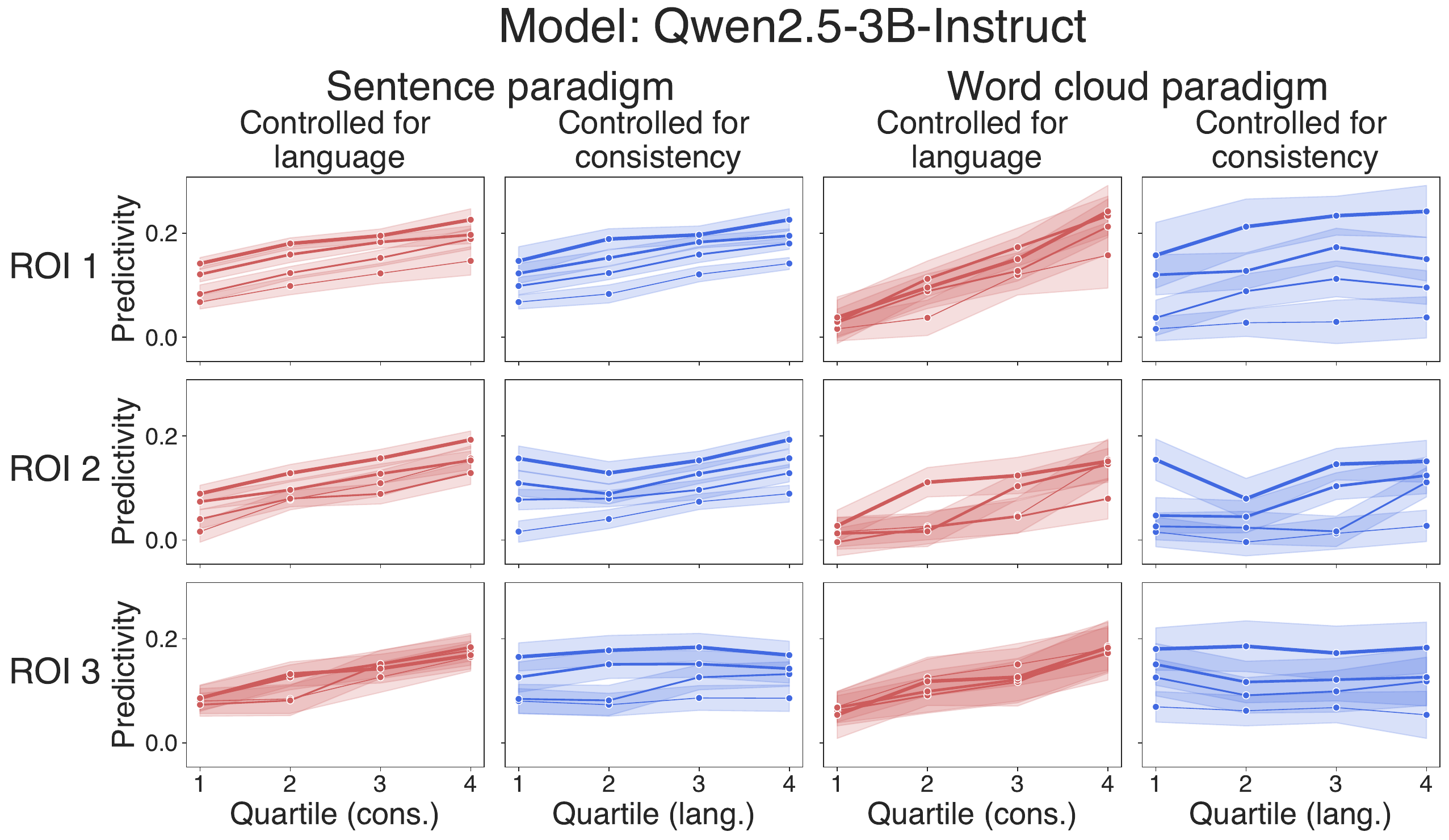}
\end{center}
\caption{\textbf{Qwen2.5-3B-Instruct predictivity by voxel quartile.}
ROI 1: $r_{\redc}=0.41 \pm 0.02, r_{\bluel}=0.27 \pm 0.06$.
ROI 2: $r_{\redc}=0.38 \pm 0.03, r_{\bluel}=0.20 \pm 0.04$.
ROI 3: $r_{\redc}=0.28 \pm 0.02, r_{\bluel}=0.03 \pm 0.03$.}
\label{fig:linecharts_qwen-3b-instruct}
\end{figure}

\begin{figure}[htb]
\begin{center}
\includegraphics[width=0.7\linewidth]{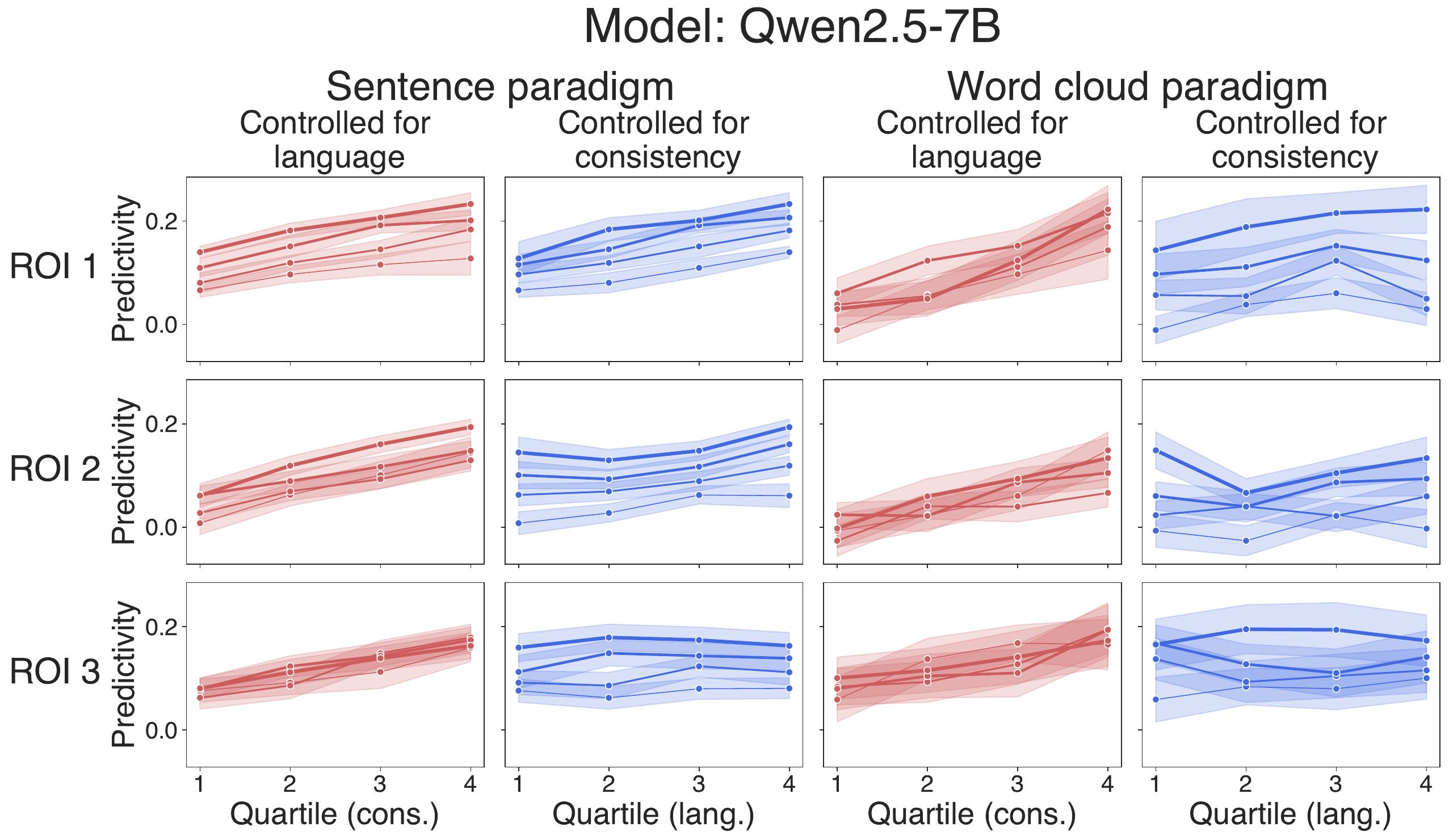}
\end{center}
\caption{\textbf{Qwen2.5-7B predictivity by voxel quartile.}
ROI 1: $r_{\redc}=0.39 \pm 0.03, r_{\bluel}=0.26 \pm 0.06$.
ROI 2: $r_{\redc}=0.39 \pm 0.03, r_{\bluel}=0.16 \pm 0.04$.
ROI 3: $r_{\redc}=0.27 \pm 0.03, r_{\bluel}=0.03 \pm 0.02$.
}
\label{fig:linecharts_qwen-7b}
\end{figure}

\begin{figure}[htb]
\begin{center}
\includegraphics[width=0.7\linewidth]{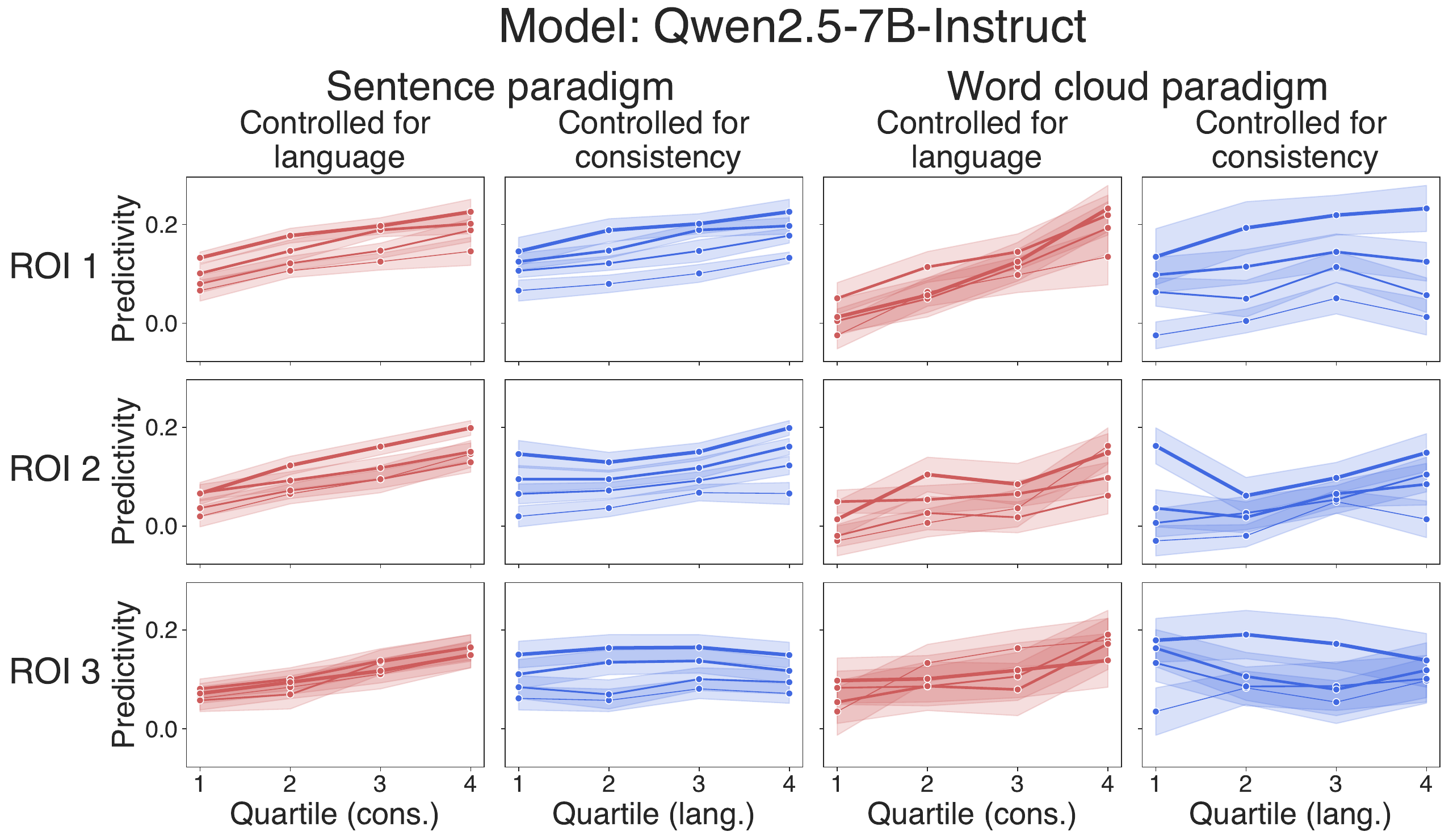}
\end{center}
\caption{
\textbf{Qwen2.5-7B-Instruct predictivity by voxel quartile.}
ROI 1: $r_{\redc}=0.41 \pm 0.02, r_{\bluel}=0.23 \pm 0.05$.
ROI 2: $r_{\redc}=0.36 \pm 0.05, r_{\bluel}=0.21 \pm 0.03$.
ROI 3: $r_{\redc}=0.26 \pm 0.03, r_{\bluel}=0.00 \pm 0.03$.
}
\label{fig:linecharts_qwen-7b-instruct}
\end{figure}

\begin{figure}[htb]
\begin{center}
\includegraphics[width=\linewidth]{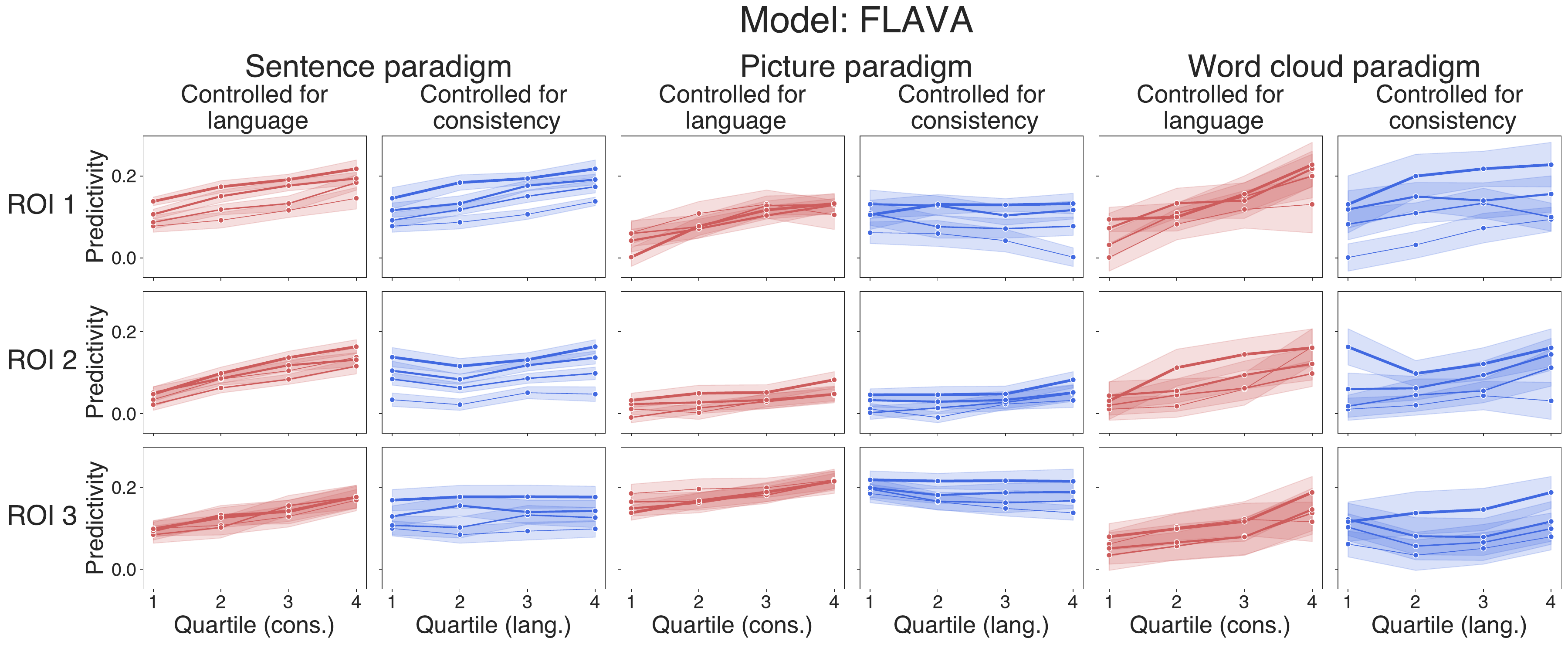}
\end{center}
\caption{\textbf{FLAVA predictivity by voxel quartile.}
ROI 1: $r_{\redc}=0.34 \pm 0.03, r_{\bluel}=0.15 \pm 0.06$.
ROI 2: $r_{\redc}=0.32 \pm 0.04, r_{\bluel}=0.15 \pm 0.02$.
ROI 3: $r_{\redc}=0.24 \pm 0.02, r_{\bluel}=-0.00 \pm 0.03$.
}
\label{fig:linecharts_flava}
\end{figure}

\begin{figure}[htb]
\begin{center}
\includegraphics[width=\linewidth]{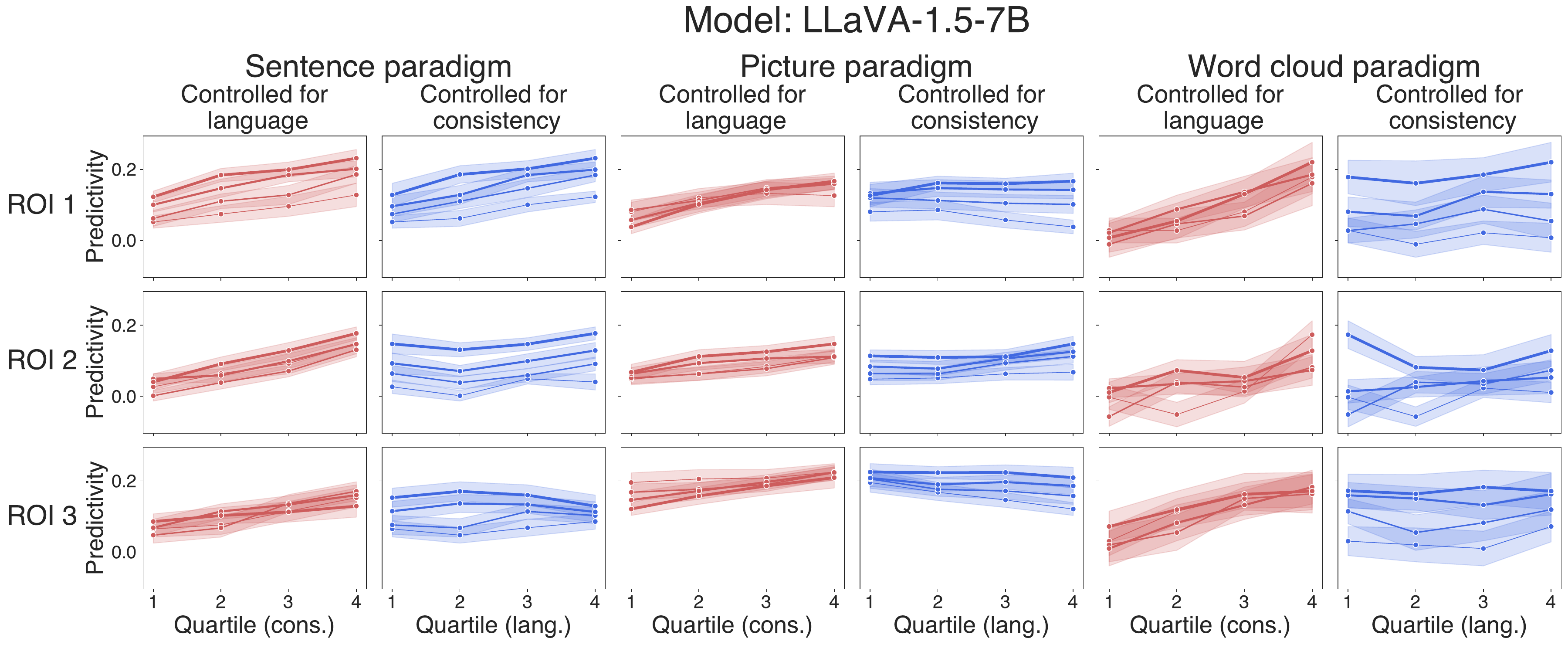}
\end{center}
\caption{\textbf{LLaVA-1.5-7B predictivity by voxel quartile.}
ROI 1: $r_{\redc}=0.35 \pm 0.03, r_{\bluel}=0.14 \pm 0.06$.
ROI 2: $r_{\redc}=0.35 \pm 0.04, r_{\bluel}=0.14 \pm 0.03$.
ROI 3: $r_{\redc}=0.28 \pm 0.03, r_{\bluel}=-0.03 \pm 0.03$.
}
\label{fig:linecharts_llava-7b}
\end{figure}

\begin{figure}[htb]
\begin{center}
\includegraphics[width=\linewidth]{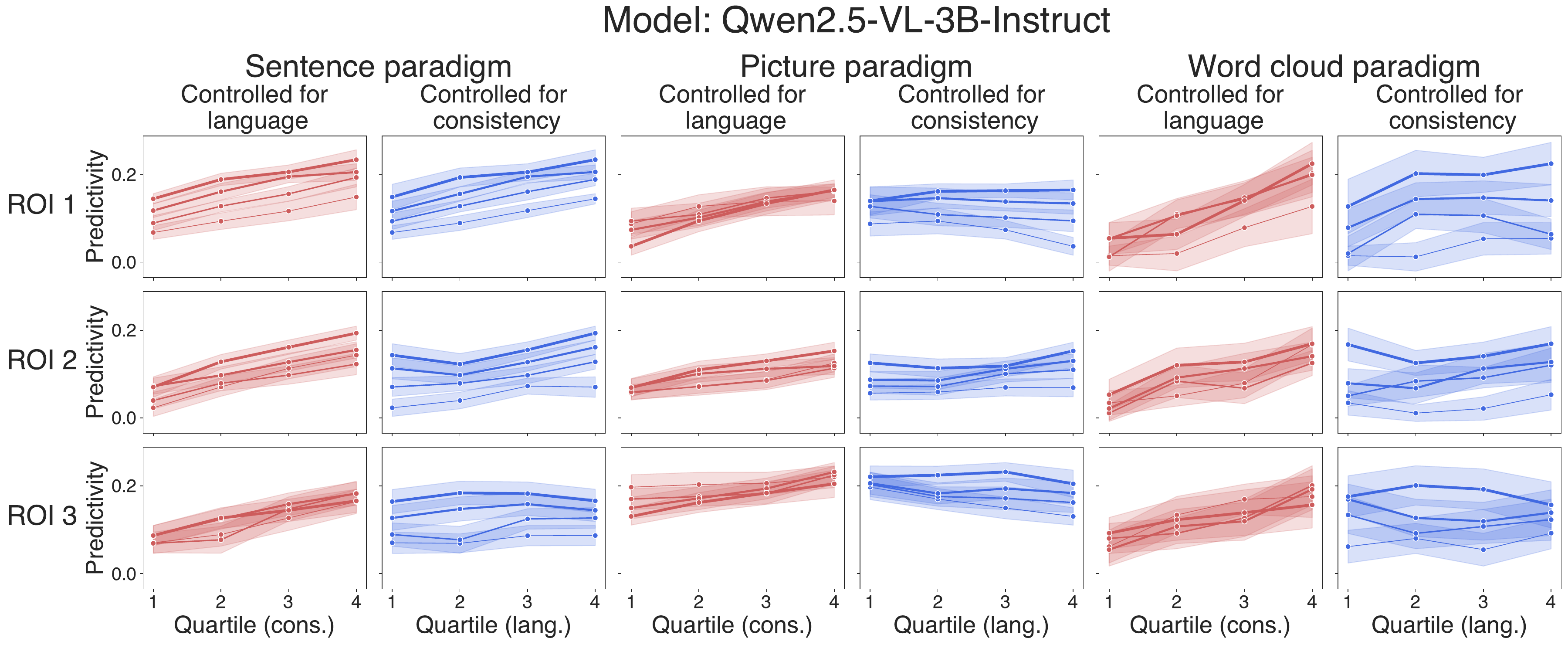}
\end{center}
\caption{\textbf{Qwen2.5-VL-3B-Instruct predictivity by voxel quartile.}
ROI 1: $r_{\redc}=0.36 \pm 0.03, r_{\bluel}=0.16 \pm 0.06$.
ROI 2: $r_{\redc}=0.34 \pm 0.02, r_{\bluel}=0.17 \pm 0.02$.
ROI 3: $r_{\redc}=0.26 \pm 0.03, r_{\bluel}=-0.02 \pm 0.03$.
}
\label{fig:linecharts_qwen-3b-vl-instruct}
\end{figure}

\begin{figure}[htb]
\begin{center}
\includegraphics[width=\linewidth]{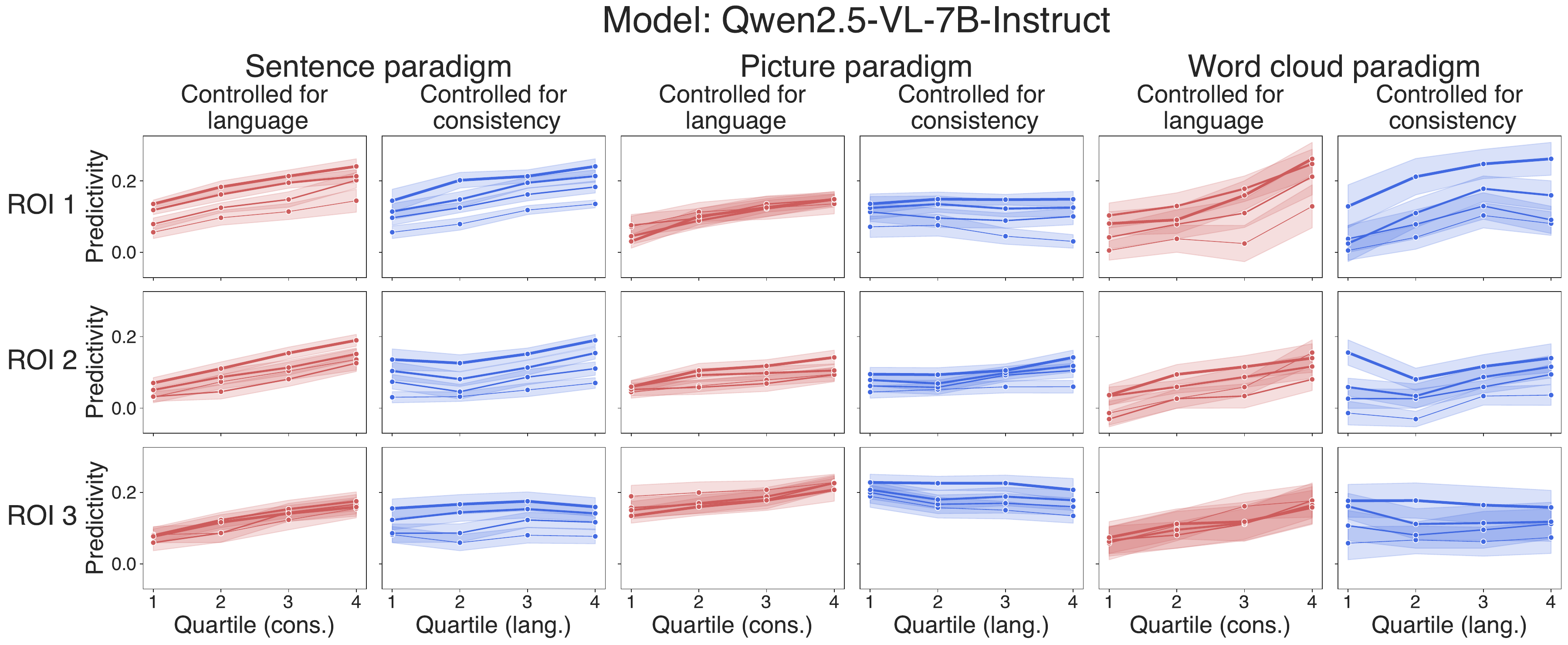}
\end{center}
\caption{
\textbf{Qwen2.5-VL-7B-Instruct predictivity by voxel quartile.}
ROI 1: $r_{\redc}=0.37 \pm 0.03, r_{\bluel}=0.20 \pm 0.06$.
ROI 2: $r_{\redc}=0.34 \pm 0.03, r_{\bluel}=0.19 \pm 0.02$.
ROI 3: $r_{\redc}=0.25 \pm 0.02, r_{\bluel}=-0.02 \pm 0.03$.
}
\label{fig:linecharts_qwen-7b-vl-instruct}
\end{figure}

\end{document}